%% file: main.tex
\documentclass[10pt,twocolumn,letterpaper]{article}

\usepackage[pagenumbers]{cvpr} 
\usepackage{graphicx}
\usepackage{float}
\usepackage{subcaption}
\usepackage{adjustbox} 
\input{preamble}

\definecolor{cvprblue}{rgb}{0.21,0.49,0.74}
\usepackage[pagebackref,breaklinks,colorlinks,allcolors=cvprblue]{hyperref}

\def\paperID{*****} 
\def\confName{CVPR}
\def\confYear{2026}

\usepackage{graphicx}
\graphicspath{{figures/}}

\newcommand{\symfootmark}{%
  \begingroup
    \renewcommand\thefootnote{\fnsymbol{footnote}}%
    \hyperlink{fn:rank}{\footnotemark[2]}%
  \endgroup
}

\newcommand{\symfoottext}{%
  \begingroup
    \renewcommand\thefootnote{\fnsymbol{footnote}}%
    \footnotetext[2]{\hypertarget{fn:rank}{}%
      \textbf{Bold} denotes the column best; \underline{underline} the second best.}%
  \endgroup
}

\title{Selective Token-Level Cryptographic Redaction for Privacy-Preserving Clinical Deployment of Large Language Models}

\author{
    Farhan Sheth$^{1,}$\thanks{The first two authors contributed equally to this work.}~, Ziyuan Yang$^{1,3,}$\footnotemark[1]~, Yongying Lan$^{1,2}$, Si Yong Yeo$^{1,3,}$\thanks{Corresponding author: Si Yong Yeo~(\texttt{siyong.yeo@ntu.edu.sg})} \vspace{5pt} \\
    $^1$ MedVisAI Lab, Singapore\\
    $^2$ Ruijin Hospital, Shanghai Jiao Tong University School of Medicine, China\\
    $^3$ Lee Kong Chian School of Medicine, Nanyang Technological University, Singapore \\
    }

\begin{document}
\maketitle

\begin{abstract}
While large language models (LLMs) are increasingly used for clinical applications, many existing pipelines require sending raw sensitive health information to remote servers for processing, which heightens the risk of privacy leakage. A natural approach to mitigate this risk is to encrypt the data before transmission. However, straightforward solutions such as encrypting the entire dataset introduce prohibitive computational, alignment, and communication overheads, rendering large-scale practical deployment infeasible. To preserve privacy while maintaining usability, we present \textbf{H}ealthcare \textbf{E}ncryption \& \textbf{R}edaction via \textbf{A}daptive \textbf{L}inguistic \textbf{D}ecomposition (\textbf{\texttt{HERALD}}), a token-level cryptographic redaction framework designed to achieve this balance by encrypting only sensitive tokens while preserving the surrounding context for downstream model utility. \texttt{HERALD} combines medical named-entity recognizer~(NER) with part-of-speech~(POS) driven policies to select candidate tokens, performs targeted lemmatization to stabilize surface forms, and substitutes each protected token with a deterministic ciphertext wrapped in explicit delimiters. Notably, \texttt{HERALD} is model-agnostic and operates entirely on the client side, ensuring that sensitive content remains encrypted throughout storage, transmission, and processing without requiring changes to downstream models. We evaluated \texttt{HERALD} on both classification and medical question answering~(MQA) tasks on public datasets. Across different tasks, experiments illustrate that fully secured baselines suffer significant utility loss, whereas \texttt{HERALD} consistently recovers performance close to plaintext. Overall, \texttt{HERALD} provides a novel utilization pipeline.
\end{abstract}

\section{Introduction}
\label{sec:intro}
Internet-scale services have broadened access to medical knowledge, but have also heightened long-standing concerns around data privacy. The rapid integration of large language models (LLMs) in clinical and operational workflows further intensifies these concerns, as LLMs are increasingly used
in the clinical prescribing, diagnosis, reporting, and patient communication pipelines~\cite{thirunavukarasu2023large,xiao2024comprehensive,nazi2024large,gaber2025evaluating,huang2024critical}.
Yet in many real-world deployments, private health information (PHI) is still transmitted to remote infrastructure in plaintext, creating risks of privacy leakage through inadvertent logging, adversarial probing, or network interception when interacting with cloud-based services such as ChatGPT~\cite{openai_chatgpt_2025} and Gemini~\cite{team2023gemini}.

\begin{figure}[!t]
    \centering
    \includegraphics[width=\linewidth]{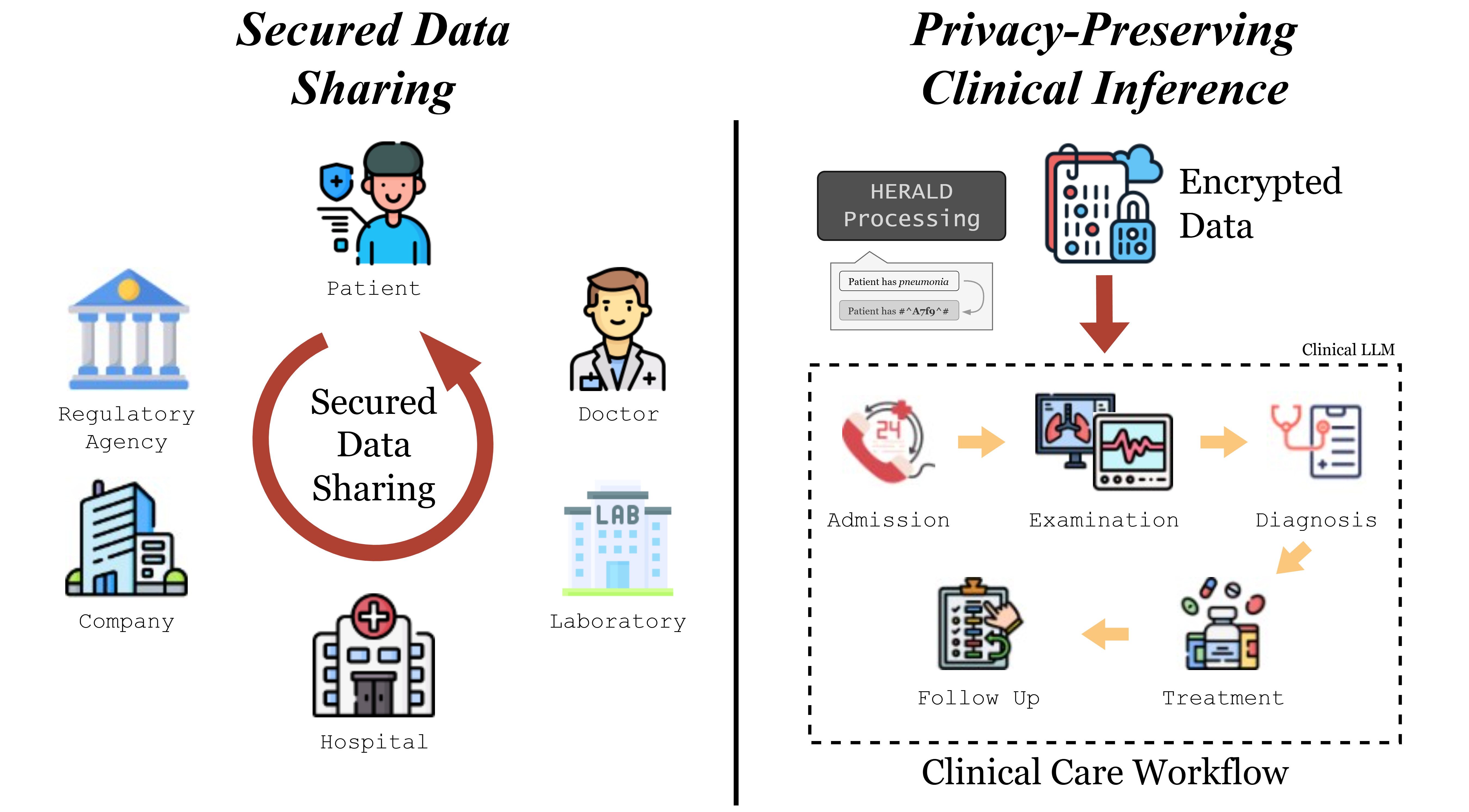}
    \caption{Overview of \texttt{HERALD} for secured clinical data sharing and privacy-preserving clinical deployment of LLMs.}
    \vspace{-15pt}
    \label{fig:first-overview}
\end{figure}

\begin{figure*}[!t]
    \centering
    \includegraphics[width=\linewidth]{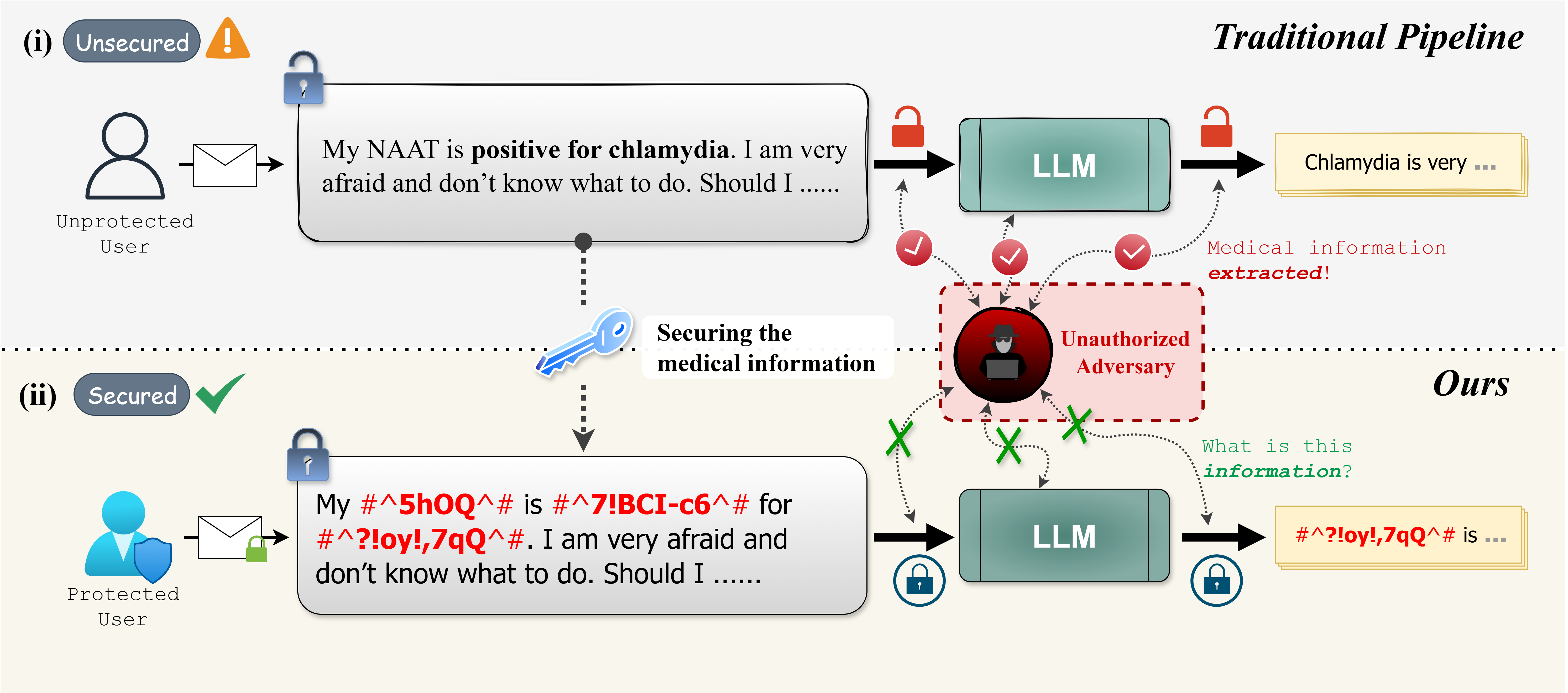}
    \caption{Comparison of \texttt{HERALD} with a traditional pipeline. \textbf{(i) Traditional:} plaintext clinical notes are sent to a remote model, exposing sensitive content. \textbf{(ii) \texttt{HERALD}:} sensitive spans are encrypted on-device into tokens (\textcolor{red}{shown in red}); the server processes only the transformed input and the client decodes outputs locally. Observers see only unreadable strings for protected tokens, while task utility is preserved.}
    \vspace{-15pt}
    \label{fig:overview}
\end{figure*}

In clinical applications such as ICU summarization, discharge instruction generation, and medication reconciliation, raw patient data is often exposed to service providers and intermediary systems, even when transport and storage protections are in place \cite{jonnagaddala2025privacy}.
Hence, transmitting plaintext-based pipeline introduces three classes of risk: (1) \textbf{Technical attacks:} LLMs are vulnerable to various threats, including jailbreaks, attribute inference attacks, and others~\cite{staab2023beyond,ishihara2023training,peng2024jailbreaking}.
(2) \textbf{Policy Limitations:} policies may restrict access to sensitive data, for instance U.S. defense limitations and industry bans motivated by data security concerns.\footnote{\tiny\url{https://www.cnbc.com/2025/01/28/us-navy-restricts-use-of-deepseek-ai-imperative-to-avoid-using.html}}
(3) \textbf{Regulatory constraints:} The health sector’s requirements complicate scaling, especially when sensitive data routinely crosses organizations \cite{tovino2025artificial, yanamala2024balancing}. Existing work spans heavyweight cryptography via encrypted computation that evaluates models on ciphertext using homomorphic encryption \cite{zhao2024privacy,gilad2016cryptonets,castro2025encryptedllm,rho2024encryption}; statistical privacy and secure execution, including secure computation or TEEs for collaborative training or protected inference \cite{majmudar2022differentially,li2023survey,mo2024machine}; federated learning as a privacy-preserving approach for biomedical NLP information extraction \cite{peng2024depth}; and de-identification–style input transformations that mask or restructure tokens or representations \cite{kan2023protecting,chowdhury2025pr,MishraSentinelLMs,linemojiprompt,norgeot2020protected, doi:10.47974/JDMSC-2146}. However, these defenses often impose substantial computational and communication costs, making them difficult to deploy in real-world clinical settings~\cite{MishraSentinelLMs,yang2025novel,linemojiprompt}. 
To address privacy concerns and ensure practicality, we rethink the pipeline shifting from \emph{sending plaintext to a model} toward privacy-preserving interactions. For medical applications, we observe that the content is largely structured text, which we partition inputs into \emph{sensitive} and \emph{non-sensitive} components. ``Sensitive text" includes PHI (names, dates, identifiers), quasi-identifiers, and detailed clinical entities (diagnoses, medications, procedures, labs, narrative findings) while ``non-sensitive text" comprises connective language (e.g., articles, prepositions, boilerplate) with minimal identity or clinical risk. Securing only sensitive spans preserves (i) the distributional and syntactic cues in the remaining plaintext and (ii) stable ciphertext token that behave like an auxiliary vocabulary the model can learn to condition on. This ensures that high-risk spans never appear in the clear during training or inference, and avoids the redundant computation and communication overhead of encrypting non-sensitive content.

Motivated by the above findings, in this paper, we propose a \textbf{H}ealthcare \textbf{E}ncryption \& \textbf{R}edaction via \textbf{A}daptive \textbf{L}inguistic \textbf{D}ecomposition (\textbf{\texttt{HERALD}}) method, which is a \emph{token-level} cryptographic redaction framework for client-side preprocessing that is model-agnostic and compatible with local, black-box, and cloud LLMs. Only tokens identified as sensitive are deterministically transformed into keyed ciphertext token identities, wrapped in explicit delimiters; non-sensitive context passes through unchanged. Each sensitive plaintext token maps to the same ciphertext identity, enabling models to learn and reuse representations over encrypted spans without architectural modification. \texttt{HERALD} thus keeps protected content in ciphertext at rest, in transit, and in use, supporting common compliance regimes~\cite{tovino2025artificial,yanamala2024balancing}.
Figure~\ref{fig:first-overview} summarizes the clinical deployment scenarios of \texttt{HERALD}, which reshapes how clinical data are shared and how LLMs are utilized in clinical practice. Specifically, clinical data are transmitted as ciphertext across different stakeholders, and the inputs to the LLM are also maintained in encrypted form, enabling end-to-end privacy-preserving inference without exposing raw sensitive information.


\begin{figure*}[!t]
\centering
\includegraphics[width=.9\linewidth]{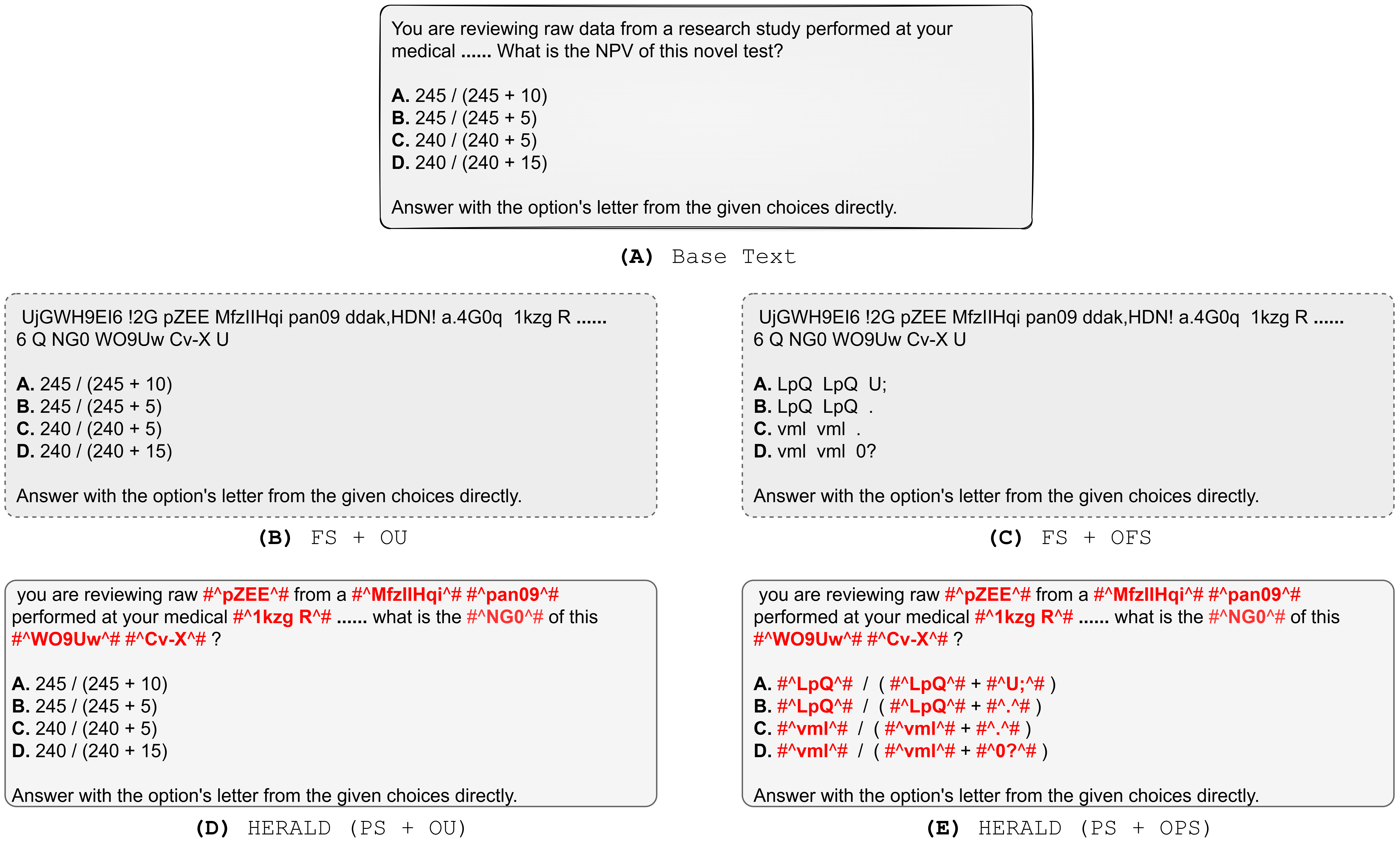}
\vspace{-10pt}
    \caption{\textbf{Text variants used in our MCQ experiments.} (A) Unsecured base multiple-choice question. (B) \textbf{FS+OU:} the question stem is fully secured, while the answer options remain in plaintext (“options unsecured”). (C) \textbf{FS+OFS:} both stem and options are fully secured. (D) \textbf{PS+OU:} the stem is partially secured using \texttt{HERALD}; options remain in plaintext. (E) \textbf{PS+OPS:} both stem and options are partially secured with \texttt{HERALD}. These controlled variants allow us to measure model performance under increasing degrees of semantic obfuscation.}
    \label{fig:questions}
\vspace{-15pt}
\end{figure*}

To clarify how our approach differs from the traditional pipeline, we provide an illustration in Figure~\ref{fig:overview}. Compared with the traditional pipeline that broadly anonymizes, disrupts model-facing structure, requires bespoke tokenization or model changes, and moves plaintext through intermediate systems with fixed privacy settings, our approach: (1) protects only what is necessary to limit information loss, (2) preserves learnable structure using stable identities and explicit delimiters, (3) works with standard tokenizers and architectures, (4) enforces end-to-end ciphertext handling, and (5) exposes tunable policies to trade privacy for utility.

\noindent The main contributions of this paper can be summarized as follows:
\begin{itemize}
    \item \textbf{Paradigm.} We propose a selective, ciphertext-preserving workflow for clinical NLP: only \emph{sensitive/priority} tokens are deterministically transformed into \emph{stable ciphertext token identities}, while surrounding non-sensitive context remains in plaintext. This enables models to condition jointly on preserved linguistic scaffold and ciphertext, while keeping sensitive content secured end-to-end.

    \item \textbf{Method.} We introduce \texttt{HERALD}, a client-side token-level cryptographic redaction framework. \texttt{HERALD} requires no architectural changes and is compatible with local deployment as well as API-only/black-box and cloud LLM services.


    \item \textbf{Empirical validation and clinical deployment guidance.} We evaluate HERALD on three public medical benchmarks across multiple LLM backbones, reporting both downstream utility and practical costs. The results map an actionable privacy--utility frontier, showing that selective, token-level securing can preserve near-plaintext performance compared to full-token encryption while adding only modest operational overhead. This directly supports real-world clinical use cases where PHI must remain protected end-to-end (e.g., cloud/API-assisted note drafting and summarization, triage message processing, coding support, and clinician-facing QA) without changing the model or requiring specialized infrastructure.


    \item \textbf{Cipher and tokenization study.} We benchmark a suite of token-level protection primitives under various regimes. We quantify privacy--utility and operational trade-offs in terms of downstream utility and tokenization footprint. This yields concrete, tokenizer-aware guidance for selecting deterministic, deployment-friendly transforms that minimize sequence blow-up while retaining utility, useful for integrating privacy protection into EHR-adjacent NLP pipelines and API-only LLM workflows.

\end{itemize}

\begin{figure*}[!t]
    \centering
    \includegraphics[width=\linewidth]{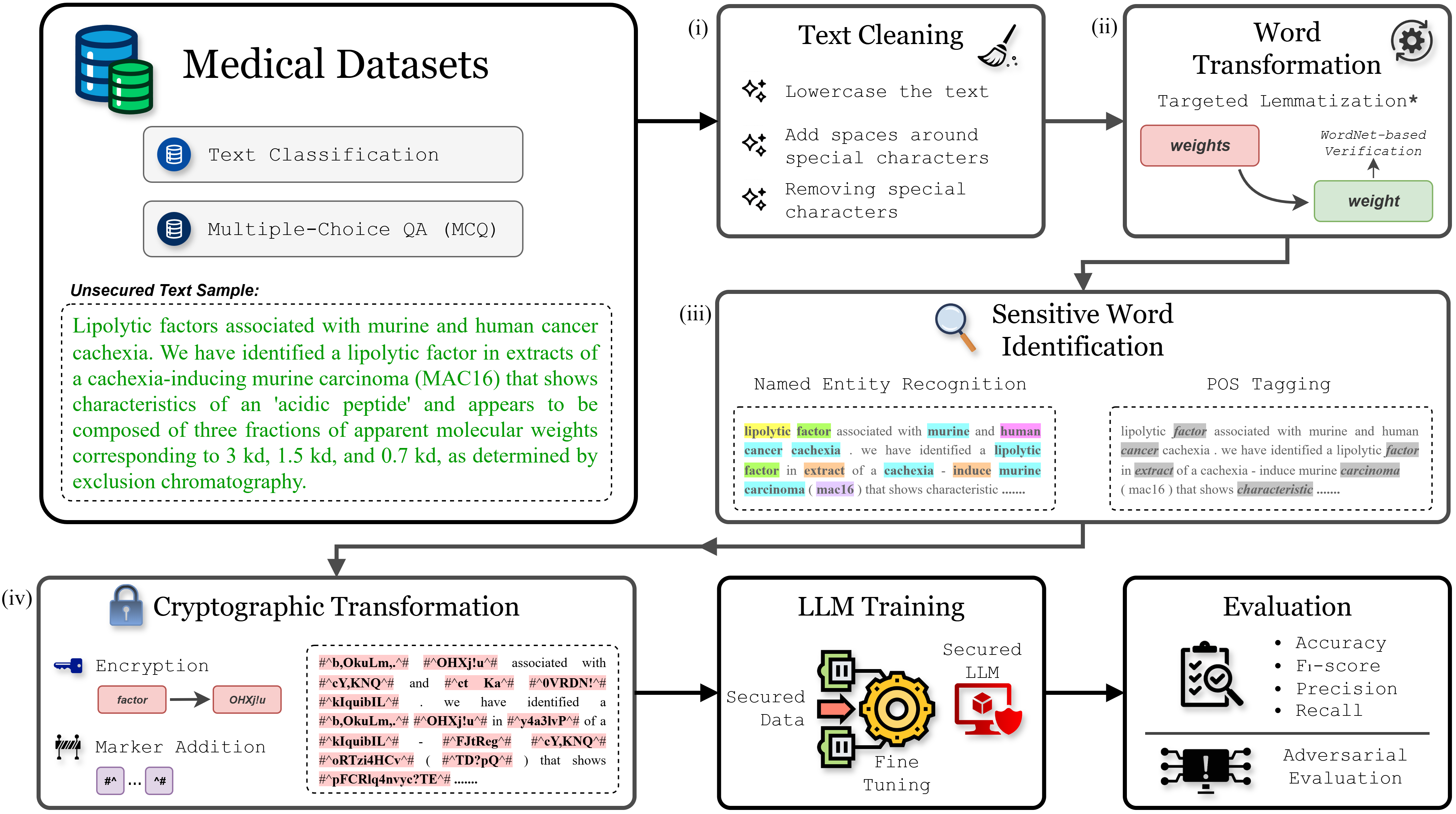}
    \caption{Overview of the \texttt{HERALD} workflow.}
    \vspace{-15pt}
    \label{fig:workflow}
\end{figure*}

\section{Methods}
\label{sec:HERALD}
In this section, we describe the protection settings used to evaluate privacy-utility trade-offs and then detail the \texttt{HERALD} framework. Here we define the textual regimes considered in the experiments, including baseline text without protection, fully secured text, and partially secured text generated by \texttt{HERALD}. We then describe the \texttt{HERALD} pipeline, which selectively protects sensitive clinical content through text normalization, targeted base-word transformation, sensitive-token identification, and token-level cryptographic transformation.

\subsection{Securing Workflow}
\label{sec:workflow}

We consider three textual variants. The first is a baseline in which the original text is left unchanged and no security or privacy mechanism is applied; this variant provides the reference performance against which our framework is evaluated. The second variant is \emph{fully secured (FS)} text, in which every token is transformed, in contrast to \texttt{HERALD} where only targeted transformations are applied; FS yields the most secure textual form. The third variant is \emph{partially secured (PS)} text produced with the \texttt{HERALD} framework (details in Section~\ref{sec:HERALD-workflow}).

Because multiple-choice question (MCQ) datasets include answer options (A, B, C, D), we further distinguish whether the options themselves are transformed in the FS and PS settings. Under FS, options are either \emph{options fully secured (OFS)} or left in their baseline form, i.e., \emph{options unsecured (OU)}. Under PS with \texttt{HERALD}, options are either left unsecured (OU) or \emph{options partially secured (OPS)} via \texttt{HERALD}. Figure~\ref{fig:questions} shows the baseline alongside four variants (FS+OU, FS+OFS, PS+OU, PS+OPS) on the MCQ dataset (FPE was used for encryption). For the classification dataset, which contains no options, we include the baseline text, FS, and \texttt{HERALD} (PS). Sequence-length inflation and GPU memory implications for these variants are summarized in Appendix~\ref{app:seq-inflation}, with end-to-end preprocessing throughput in Appendix~\ref{app:encryption-throughput}.

\subsection{HERALD}
\label{sec:HERALD-workflow}

\texttt{HERALD} is a model-agnostic privacy framework that bridges the gap between heavyweight cryptography and naive text masking through selective token-level protection. Specifically, \texttt{HERALD} follows a four-stage pipeline: (i) text normalization and cleaning, (ii) targeted base-word transformation, (iii) sensitive-token identification, and (iv) token-level cryptographic transformation. Only priority tokens are reduced to their base forms via \textit{targeted lemmatization}, while others are left unchanged to preserve semantics. The end-to-end pipeline is illustrated in Figure~\ref{fig:workflow}.

\subsubsection{Text Cleaning} 
The first stage standardizes the input text to stabilize tokenization and remove trivial variability. We lowercase all characters to ensure case-insensitive processing, normalize whitespace around punctuation so symbols are tokenized independently, and remove uncommon special characters. Overall, this yields a clean text $x_{\mathrm{clean}}$ that avoids spurious tokens (e.g. casing or punctuation variants) that could hinder downstream learning. Related ablation studies can be found in Appendix~\ref{app:stopwords}.

\subsubsection{Word Transformation}
Next, \texttt{HERALD} applies \textbf{targeted lemmatization} to reduce inflectional sparsity for candidate sensitive tokens. Lemmatization is mapping inflected or derived word forms to their dictionary base (lemma; e.g., \textit{running}\textrightarrow \textit{run}) \cite{khyani2021interpretation}. 
So, rather than lemmatizing every word, which could distort clinically relevant phrasing, we only normalize tokens that are likely to be sensitive and map them to their morphological base forms before cryptographic transformation (using a preliminary sensitive-token detector). For example, ``diabetic \textbf{patients}'' becomes ``diabetic \textbf{patient},'' ``multiple \textbf{tumors}'' becomes ``multiple \textbf{tumor},'' and ``recent \textbf{fractures}'' becomes ``recent \textbf{fracture}.'' Similarly, inflected diagnostic terms such as ``was \textbf{diagnosed} with pneumonia'' are reduced to ``was \textbf{diagnose} with pneumonia'' prior to protection.

Because lemmatizers are typically rule-based or statistical, they sometimes return strings that are not valid words in the target lexicon. To filter such cases, we accept a candidate lemma only if it appears in WordNet \cite{miller1995wordnet}, which organizes English nouns, verbs, adjectives, and adverbs into synonym sets (“synsets”), each representing a lexicalized concept. To improve domain-specific coverage, our methodology supplements WordNet with a curated medical vocabulary. This check guards against nonsensical outputs and over-aggressive lemmatization. Overall, this step produces a partially normalized text $x_{\mathrm{base}}$. By unifying inflected variants of important terms, we improve consistency during cryptographic transformation, thereby aiding the model in learning from secured sensitive data. Crucially, this transformation is \textbf{targeted}: only tokens earmarked as sensitive (as determined in the next stage) are lemmatized, which avoids altering common words in a way that might change semantic nuances. We compare targeted vs. untargeted lemmatization and find consistent gains for the targeted strategy (Appendix~\ref{app:lemmatization}). Overall, targeted lemmatization approach retains the original sentence structure and meaning for the model, while simplifying sensitive tokens to a canonical form for robust cryptographic transformation.

\subsubsection{Sensitive Word Identification} 
In the third stage, \texttt{HERALD} automatically identifies which tokens carry sensitive information and merit securing. We leverage domain-specific \textbf{named entity recognition (NER)} to detect PHI and other medical entities in the text \cite{paul2024deidclinic}. To protect patient privacy, we de-identify (transform) all PHI and medical data including diagnoses, medications, medical record numbers, and other medical entities. We use Medical-NER, 
a DeBERTa-based model fine-tuned on PubMed, to detect medical entity types. We augment NER with POS based priority rules to capture additional \textit{content-bearing} tokens that might not be classical named entities but are still important or sensitive in context. Specifically, we treat as sensitive any token that is either (i) a medical entity (e.g. conditions, medications, test results) or (ii) a noun/numeral that is deemed significant to the sentence meaning (ensuring we don’t inadvertently leave an important clue in plaintext). Appendix~\ref{app:pos-selective} specifies POS-based tunable privacy tiers for broader sensitive-token detection. For POS tagging, we use the open-source spaCy library \cite{Honnibal2020spaCy}. These POS-based rules act as a backstop so that if the NER misses a rare term, the framework can still flag it due to its grammatical role. Let $\mathcal{I}(x)$ denote the index set of tokens identified as sensitive in input text $x$ after this step. All tokens \textit{not} in $\mathcal{I}(x)$ are considered non-sensitive context and will remain in plaintext. By focusing cryptographic transformation on the most privacy-critical spans \cite{paul2024deidclinic}, \texttt{HERALD} maximizes the protection of patient information while minimizing disruption to the surrounding language that the model needs to interpret.

\subsubsection{Cryptographic Transformation}
Finally, each sensitive token is transformed via a deterministic cryptographic encryption. We define a cryptographic transformation function $E_k(\cdot)$ parameterized by a secret key $k$. For each token $w_i$ in the sensitive set $\mathcal{I}(x)$, we first apply the lemmatization function $\ell(w_i)$ (from the previous stage) to obtain its base form, then replace it with a ciphertext string using $E_k$. We also insert special delimiter markers (denoted by $\langle\!\langle ... \rangle\!\rangle$) around each de-identified (or encrypted) token to explicitly mark its boundaries. Formally, the output token $z_i$ is:
\begin{equation}
z_i =
\begin{cases}
\langle\!\langle E_k(\ell(w_i)) \rangle\!\rangle, & \text{if } i \in \mathcal{I}(x)\\[6pt]
w_i, & \text{if } i \notin \mathcal{I}(x)
\end{cases}
\label{eq:HERALD-workflow}
\end{equation}
The resulting transformed sequence is $T_k(x) = [z_1,z_2,\dots,z_n]$. By construction, tokens outside $\mathcal{I}(x)$ appear verbatim in $T_k(x)$, preserving the surrounding context, while tokens in $\mathcal{I}(x)$ are deterministically mapped (under key \textit{k}) to stable ciphertext token and wrapped with delimiters. Finally, the tokens in the resulting transformed sequence are concatenated with single spaces to form the final secured sequence. We emphasize that cryptographic transformation (or encryption) is applied at the token level rather than on the entire sequence, aligning with findings that encrypting an entire input string will destroy almost all useful linguistic structure \cite{MishraSentinelLMs}.

We use “$\# \text{\textasciicircum}$” and “$\text{\textasciicircum} \#$” as sentinel tokens (or $\langle\!\langle ... \rangle\!\rangle$) marking the start and end of a cryptographically transformed span, respectively (Related ablation studies can be found in Appendix~\ref{app:marker-variants}). These markers help the model in handling these transformed tokens, the inserted markers serve as explicit cues. They delineate the ciphertext boundaries and effectively introduce a simple “syntax” for cryptographically transformed tokens. During fine-tuning, the model learns that any token wrapped in the markers should be treated as an atomic ciphertext lexical item: not human-interpretable and computationally infeasible to invert without the key, yet stable and learnable that the model can embed and condition on across occurrences. This marking prevents the delimiter-wrapped ciphertext strings from being misinterpreted as natural language or split into subwords improperly. It essentially teaches the model a new dialect where sequences inside $\# \text{\textasciicircum}\, ... \,\text{\textasciicircum} \#$ are opaque symbols, each behaving like a stable, unique word type. Nearest-neighbor recoverability and embedding-space probing indicate cipher tokens do not geometrically ‘point back’ to their plaintext (Appendix~\ref{app:recover-tokens}, \ref{app:embeddings}). Therefore, by containing cipher-tokens with special delimiters, we mitigate the risk of the model treating parts of a ciphertext as independent tokens or punctuation. 

The outcome of this stage is the final \texttt{HERALD}-protected text. This text can be safely sent to a black-box LLM or stored in a database: it preserves the necessary context and structure for model processing, but any originally sensitive tokens are now “locked” as ciphertext. In summary, \texttt{HERALD}’s workflow keeps non-sensitive context in the clear (for model utility), transforms sensitive tokens into consistent encrypted ciphertext token identities (for privacy), and uses marker cues to ensure the transformed text remains parseable by transformer models during fine-tuning and inference.

\subsection{Rationale behind HERALD}
\label{sec:rationale}
\textbf{Why does \texttt{HERALD} work?} \texttt{HERALD} aims to preserve task-relevant linguistic structure while removing direct exposure of high-risk content. Fully securing the entire input largely erases the syntactic and discourse cues that LLMs exploit, leading to substantial utility loss \cite{MishraSentinelLMs}. Conversely, naive masking or deletion often yields ungrammatical inputs and can remain vulnerable to reconstruction from context. By protecting only selected sensitive spans and leaving the remaining context intact, \texttt{HERALD} retains the linguistic scaffold needed for inference while encoding protected spans as ciphertext that function as opaque but learnable lexical items the model can embed and condition on \cite{buban2025encrypted}.

A key design choice is to replace protected spans with \emph{stable, keyed ciphertext token identities} so the model can accumulate evidence across occurrences and learn representations, similar to how it learns rare lexical types. This improves sample efficiency relative to per-instance randomization or heavy noise, which can fragment supervision and degrade downstream performance \cite{qu2021natural}. Importantly, \texttt{HERALD} does not rely on an exposed substitution table: ciphertext token identities are computed deterministically from a secret key and are computationally infeasible to invert without it. Practical guessing attacks are therefore limited to cases where the protected domain is intrinsically small and can be exhaustively enumerated; our policies prioritize spans whose candidate space is large (e.g., identifiers, names, free-text entities), and users can increase protection by expanding span selection when auxiliary knowledge is a concern.

From a privacy perspective, \texttt{HERALD} keeps protected spans in ciphertext throughout storage, transmission, and model use, with the secret key remaining client-side. Under standard cryptographic assumptions, observing only the transformed text does not enable efficient recovery of protected values without the key \cite{singh2013study}. This reduces exposure even under remote hosting, logging, or prompt-based elicitation, since any leakage contains ciphertext tokens rather than plaintext \cite{paul2024deidclinic}.

\texttt{HERALD} does not provide information-theoretic secrecy: plaintext context can still support statistical inference about protected spans, particularly under strong auxiliary information. The resulting privacy-utility trade-off is therefore tunable via the span-selection policy, allowing deployments to adopt a conservative setting when inference risk is high and a lighter setting when utility is paramount (Appendix~\ref{app:privacy-utility}).

\subsection{Problem Definition}
\label{sec:problem-definition}
We formalize the task addressed by \texttt{HERALD}. Let $\mathcal{X}$ denote the space of plaintext inputs (e.g., token sequences in the medical domain) and $\mathcal{Y}$ the output space (e.g., diagnosis labels for classification or free-form answers for QA). For any $x=[w_1,\dots,w_n]\in\mathcal{X}$, a subset of tokens is sensitive. We define a \textit{sensitivity function} $s_n:\mathcal{X}_n\to 2^{[n]}$ that returns the index set of sensitive tokens in $x$, where $[n]={1,\ldots,n}$ indexes positions in $x$. The goal is to learn a model $f_\theta:\mathcal{X}\to\mathcal{Y}$ (parameters $\theta$) that attains high task performance \emph{without ever observing sensitive tokens in plaintext}. To this end, we introduce a transformation $T_k:\mathcal{X}\to\mathcal{X}'$, where $\mathcal{X}'$ is the space of partially encrypted texts, with the following properties. First, for any $x\in\mathcal{X}$, $T_k(x)$ deterministically encrypts each sensitive token $w_i$ for $i\in s(x)$ into a ciphertext token while leaving all other tokens unchanged. Second, the transformation is \textbf{lossless} for an ideal model in the sense that there exists $f_\theta$ whose performance on $T_k(x)$ matches that of the best model on $x$; in practice, we approximate this condition by preserving maximal task-relevant information in $T_k(x)$ subject to privacy constraints. Third, encryption is computationally secure: without the secret key $k$, it is infeasible to recover any sensitive $w_i$ from $T_k(x)$ even when the adversary knows the algorithm $T$ and has unlimited access to samples of encrypted text. This aligns with the proposal that the ciphertext should reveal no information beyond (possibly) length and the indistinguishability-under-chosen-plaintext-attack notion (see Appendix~\ref{app:jailbreak} for jailbreaking and prompt injection attacks and Appendix~\ref{app:extraction} for training-data extraction). We note that our practical instantiation may not strictly satisfy such definitions under all tokenization and marker choices; we adopt these ciphers as a deliberate privacy–utility trade-off.

We train $f_\theta$ on transformed pairs ${(T_k(x^{(j)}),y^{(j)})}_{j=1}^N$. The objective is task-specific. For classification, we minimize cross-entropy
\begin{equation}
   L(\theta) = -\frac{1}{N}\sum_{j=1}^N \log p_\theta\!\big(y^{(j)} \mid T_k(x^{(j)})\big)
\label{eq:cross-entropy}
\end{equation}
where $p_\theta(y\mid x)$ denotes the model’s predictive distribution. For generative settings (MCQ), we use a standard language-modeling negative log-likelihood over target tokens. At inference, user inputs are transformed by the same $T_k$ before they are provided to $f_\theta$, so the model never receives raw sensitive tokens at either training or test time. We assume $T_k$ is invertible only by parties holding $k$ (for deterministic symmetric encryption) or non-invertible (for one-way hashing). Consequently, even if $f_\theta$ is compromised or queried adversarially, plaintext sensitive content remains concealed.

Our threat model targets \textbf{exposure of plaintext} to unauthorized entities via model leakage or interception. We do not seek to prevent the model from leveraging contextual information or stable ciphertext tokens during learning; rather, we ensure that plaintext sensitive spans are never provided to the model or service. This setting matches common LLM deployments (e.g., cloud APIs) in which the model is an untrusted or partially trusted black-box and the data owner seeks to prevent disclosure of PHI. \texttt{HERALD} enforces that such services observe only encrypted PHI. We further assume the adversary lacks the secret key $k$ and cannot break the underlying cryptography via brute force or cryptanalysis, motivating our use of established primitives (e.g., AES, FPE, or SHA). If the key is compromised (e.g., through system breach or social engineering), encryption offers no protection; key management is therefore essential but out of scope for our NLP focus. Finally, we assume that encryption occurs in a secure environment before fine-tuning so that plaintext is not exposed during training (e.g., local preprocessing or a secure enclave). The trained model itself is not inherently privacy-preserving as memorization could still occur, so \texttt{HERALD} can be combined with complementary defenses such as differentially private fine-tuning \cite{li2021large}. In our experiments, standard fine-tuning on encrypted data did not cause the model to memorize or emit plaintext tokens, as these are never observed, which mitigates the primary memorization risk in practice.

\subsection{Theoretical Groundings}
\label{sec:theoretical}
From a theoretical perspective, \texttt{HERALD} applies an \textbf{$\mathbf{\varepsilon}$-cryptographic transformation} to the input text, where $\varepsilon$ quantifies information lost through encryption. When $\varepsilon=0$, the transformation is semantics-preserving for the downstream task; in practice $\varepsilon>0$ because encrypting a token removes its plaintext identity from the model’s view, replacing it with a stable ciphertext identity, which carry task-relevant signal. Let $X$ denote the random variable over inputs and $Y$ the task label or output. Our aim is to keep the mutual information $I(Y;T_k(X))$ close to $I(Y;X)$ while reducing leakage about sensitive content, formalized as minimizing $I(\text{sensitive};T_k(X))$, where “sensitive” denotes the sensitive portion of $X$. Accordingly, \texttt{HERALD} seeks to \textbf{maximize} $I(Y;T_k(X))$ (utility) subject to \textbf{minimizing} $I(\text{sensitive};T_k(X))$ (privacy leakage). Under ideal semantic security, ciphertext does not enable efficient recovery of plaintext from the ciphertext value alone; however, deterministic token-level schemes leak equality/repetition structure. Consequently, any residual inference can exploit both (i) surrounding plaintext context and (ii) repetition/co-occurrence patterns over stable ciphertext identities.
Thus, $T_k(X)$ may leak signals beyond context alone (e.g., equality/frequency structure), while still preventing efficient recovery of the exact plaintext token without the secret key. For example, in “The patient’s HIV test came back positive,” if “HIV” is encrypted as $\# \text{\textasciicircum} \mathtt{Nco} \text{\textasciicircum} \#$, an attacker could still guess the disease from “test came back positive,” a cue present regardless of encryption. An information-theoretic reading would note that $I(\text{disease};\text{context})$ is high even before transformation; perfect privacy would therefore require sanitizing context as well, which would impair utility. \texttt{HERALD} explicitly opts for \textbf{computational privacy}, making recovery of exact token identities cryptographically hard rather than information-theoretic secrecy. This mirrors semantic-security intuitions: while token-level ciphertexts are secure against efficient inversion, we intentionally leave correlated plaintext context, enabling adversaries to do better than chance by exploiting those correlations. Precisely quantifying such leakage reduces to analyzing contextual inference of encrypted token identities (a proxy for attacks that attempt to guess protected spans from remaining plaintext context and ciphertext repetition patterns). If a sensitive token has high conditional entropy given its context, encryption effectively limits exposure; if the token is nearly determined by context, any method that leaves context intact (including \texttt{HERALD}) will inevitably leak. Hence, theoretical protection is strongest for \emph{entropic} identifiers (e.g., IDs, names, irregular strings) and weaker for sensitive attributes tightly coupled to context. In practice, many PHI elements are high-entropy, so encrypting them is effective, whereas strongly signaled medical conditions may require additionally protecting selected context tokens. We empirically bound contextual leakage via (i) semantic-similarity/embedding analyses and (ii) token recoverability stress tests; see Appendix~\ref{app:embeddings} and Appendix~\ref{app:recover-tokens}.

A complementary view treats \texttt{HERALD} as introducing a constrained \textbf{foreign alphabet} that the model must learn to process. The encrypted spans delimited by markers “$\# \text{\textasciicircum} \mathtt{...} \text{\textasciicircum} \#$” act like tokens from an auxiliary “language.” In fine-tuning, the model learns representations for these cipher sequences (or for their constituent subword pieces under the tokenizer), with the markers preventing accidental collisions with ordinary vocabulary. When many distinct sensitive types exist, capacity is allocated across many ciphered forms; nevertheless, the number of categories of sensitive content (e.g., PHI types) is typically far smaller than the full lexicon, and many items such as personal names may be seen only once. When the model encounters an unseen ciphertext form at test time such as $\# \text{\textasciicircum} \mathtt{XYZ} \text{\textasciicircum} \#$, it behaves like an out-of-vocabulary encrypted token identity: the model cannot leverage a previously learned embedding for that specific identity, so it may rely more on surrounding plaintext context to infer coarse role or constraints (e.g., entity type, syntactic function). When ciphertext identities recur across examples, the model can learn representations for them, such as learning a foreign language, enabling consistent conditioning on encrypted spans without recovering plaintext. Because \texttt{HERALD} preserves ample context, predictive performance remains robust, as our experiments indicate. Empirically, models fine-tuned under \texttt{HERALD} exhibit only moderate degradation relative to plaintext training while \emph{significantly outperforming} the naive baseline that encrypts all tokens, which collapses performance.
These observations are consistent with prior findings, for example, Mishra~\textit{et~al.}~\cite{MishraSentinelLMs} report that token-level encryption can preserve accuracy across NLP tasks once models are adapted, supporting the view that selective encryption recovers most utility while enabling a configurable security layer.







\section{Results}
\label{sec:results}
We hypothesized that (i) fully securing every token would erase the contextual and syntactic cues that language models rely on, yielding a pronounced utility drop; (ii) selectively securing only prioritized spans would preserve enough scaffold for models to treat ciphertext as stable learnable information and recover much of the plaintext baseline; and (iii) transforms that better preserve tokenizer stability would enable stronger recovery. Across one classification benchmark and two medical MCQ benchmarks, the experiments support these hypotheses: fully secured variants consistently collapse, whereas \texttt{HERALD} partially secured variants recover substantial performance without architectural changes.

\subsection{Language Models}
\label{sec:LM}

We consider two categories of LLMs: models fine-tuned for supervised medical text classification, and generative LLMs used to address generative datasets, specifically medical multiple-choice questions (MCQs). The former are predominantly encoder-only architectures, with a few decoder-only models included for comparison; most are general-purpose models, complemented by domain-adapted variants tailored to biomedical or clinical text. 

These model families map to common stages of medical NLP pipelines that transform unstructured text into actionable signals \cite{huang2024critical}. Encoder backbones are typically used upstream for de-identification, concept extraction, and document-level classification over clinical notes and biomedical literature \cite{wiest2024privacy}. Generative LLMs are increasingly evaluated for clinician-facing question answering and summarization~\cite{singhal2023large,nori2023capabilities}, and have begun to enter documentation workflows via ambient systems that draft visit notes from conversations~\cite{bekbolatova2024transformative}. 

\subsubsection{Classification Models}
\label{sec:classification-models}
Classification models are employed as backbones for downstream text classification. The majority are encoder-only Transformer encoders trained with masked language modeling (MLM), while a subset are decoder-only models pretrained with next-token prediction. The used methods include BERT~\cite{devlin2019bert}, RoBERTa~\cite{liu2019roberta}, DeBERTa~\cite{he2021debertav3}, GPT-2~\cite{radford2019language}, BioBERT~\cite{lee2020biobert}, ClinicalBERT~\cite{wang2023optimized}, and BioGPT~\cite{biogpt}.

\subsubsection{Generative Models}
\label{sec:LLMs}

We employ generative LLMs to answer medical-domain multiple-choice questions (MCQs). The suite includes both general-purpose LLMs and models adapted to biomedical or clinical text.
The used LLMs include Qwen-2.5-7B~\cite{qwen2.5}, Mistral-v0.3~\cite{mistralv0.1}, Llama-3.1-8B~\cite{meta-llama-3.1}, Llama-3.1-Aloe-Beta-8B~\cite{meta-aloe}, and Llama-3-Med42-8B~\cite{med42}. All LLMs are publicly available, and in our experiments we only fine-tune them rather than retraining from scratch.

\subsection{Datasets}
\label{sec:datasets}
We select one public dataset for the classification task and two public datasets for the multiple-choice question (MCQ) task. This choice spans two distinct clinical genres: biomedical abstracts and standardized exam questions, thereby testing robustness under heterogeneous lexical and structural distributions. The brief introduction can be found as follows:

\noindent \textbf{Medical Abstracts Text Classification (Med-TC) Dataset}~\cite{medabstract}\textbf{:} Med-TC comprises 14{,}438 medical abstracts labeled into five clinically salient categories: Neoplasms, Digestive System Diseases, Nervous System Diseases, Cardiovascular Diseases, and General Pathological Conditions. The label space covers major disease systems and broad pathology, making the corpus representative of system-level clinical topicality. Its scale (11{,}550 training abstracts) is adequate for reliably fine-tuning encoder backbones. We follow the authors' split with 11{,}550 train and 2{,}888 test abstracts, reserving $5\%$ of training for validation.

\noindent \textbf{MedMCQA}~\cite{medmcqa}\textbf{:} MedMCQA is a large-scale MCQA dataset constructed from AIIMS and NEET-PG entrance examinations. It contains more than 194k questions spanning approximately 2.4k healthcare topics and 21 medical subjects, with an average question length of 12.77 tokens and substantial topical diversity. The benchmark assesses more than ten reasoning skills across a wide range of medical domains. For computational tractability, we evaluate a curated subset consisting of four-option questions from five core subjects, Medicine, Gynecology \& Obstetrics, Pediatrics, Pathology, and Psychiatry, yielding a total of 20{,}287 items. We construct a subject-stratified split targeting $80/5/15$, so that each split preserves the subject mix rather than concentrating on a narrow subset of domains; realized counts are 15{,}941/995/3{,}351 for train/validation/test ($78.6\%/4.9\%/16.5\%$).

\noindent \textbf{MedQA-USMLE}~\cite{jin2021disease}\textbf{:} MedQA-USMLE is a free-form multiple-choice open-domain QA dataset targeting medical problem solving. We restrict analysis to the English portion, containing 12{,}723 questions derived from the United States Medical Licensing Examination (USMLE). Relative to MedMCQA, its questions more often require multi-step clinical reasoning, making it a complementary stress test for selective encryption when clinically critical entities are obscured. We use the authors' split of 10{,}178/1{,}272/1{,}273 for train/validation/test.

\subsection{Experimental Setup}
\label{sec:exp-setup}
We evaluate whether token-level protection can meaningfully reduce exposure of sensitive clinical content while preserving downstream task utility. Experiments span two representative clinical NLP workloads: (i) supervised medical text classification and (ii) medical multiple-choice question answering (MCQ), which jointly reflect common settings where models consume patient-linked narratives for categorization or decision support. We compare three privacy regimes: plaintext (\textit{Base}), fully secured text where all tokens are transformed (\textit{FS}), and \texttt{HERALD} partial security where only prioritized tokens are transformed (\textit{PS}). For MCQ prompts, we additionally vary whether answer options remain readable (\textit{OU}) or are transformed (\textit{OFS}/\textit{OPS}). This protocol directly tests whether selective ciphertext substitution can offer a practical privacy--utility trade-off in healthcare deployments without requiring model or infrastructure changes. We report Accuracy, F\textsubscript{1}-score, Precision, and Recall, and we measure training/inference overhead to assess practical feasibility in healthcare settings.
\subsubsection{Cryptographic Methods}
\label{sec:crypt-methods}
To protect sensitive medical text during language model training, we apply token-level transformations. Our baselines span multiple cryptographic families and architectures, but all share a core requirement: the transformation must conceal the original content while preserving sufficient structural or statistical regularities for effective learning. These methods can be roughly divided into encryption-based and obscuration methods.

\paragraph{Encryption-Based Methods.} We evaluate symmetric-key ciphers as our baselines, including AES in SIV and ECB modes and Blowfish in ECB mode, as well as format-preserving encryption (FPE). For each method, we summarize theoretical guarantees, known weaknesses, and empirical effects on model performance, thereby characterizing the security-utility trade-off in Appendix~\ref{app:cryptographic-methods}. 

\paragraph{Obscuration Methods.}
We evaluate one-way transformations (e.g., hashing) as auxiliary baselines rather than encryption. These methods preserve similarity in the obfuscated space, enabling model generalization, and they support high throughput, deterministic token obfuscation when pre-image resistance is the priority. We study fuzzy hashing, soft hashing, MD5, SHA-1, and SHA-256; details appear in Appendix~\ref{app:obsc-methods}.

\subsubsection{Prompting Strategy}
We follow the standard MCQ pipeline in the MMLU style \cite{hendrycks2020measuring}. Each prompt presents the question, options labeled A, B, C, and D, and a short instruction to answer with the option letter only. Prompts are wrapped in the model’s native template, using the official Hugging Face templates for Qwen-2.5, Mistral-v0.3, and Llama-3.1, and the corresponding Llama template for Llama-3.1-Aloe-Beta and Llama-3-Med42.

\subsubsection{Evaluation Metrics}
We report Accuracy, F\textsubscript{1}-score, Precision, and Recall. Inputs: \emph{Base} (no protection), \emph{FS} (all tokens protected), and \texttt{HERALD} \emph{PS} (only sensitive tokens). For MCQs we also vary option protection: \emph{OU} (unsecured), \emph{OFS} (fully secured), and \emph{OPS} (partially secured via \texttt{HERALD}).

\subsubsection{Training Configuration}
We attach a linear classification head to each base model and fine-tune with the Hugging Face \texttt{Trainer}. To stabilize optimization and cut computation, a subset of layers is frozen based on model size. All LLMs use \texttt{unsloth} with LoRA adapters. We apply early stopping and a learning-rate scheduler; the final model is the checkpoint with the best validation F\textsubscript{1}. For computational reasons, models are capped at 8B parameters. Cryptographic baselines use a 16-byte key, except AES-SIV which uses 32 bytes. Seed sensitivity is summarized in Appendix~\ref{app:seed-stability}, where low coefficients of variation indicate stable results.

\subsection{Results and Analysis}
\label{sec:results-and-analysis}

\subsubsection{Classification Task}

For the classification task, performance when trained and evaluated on plaintext is treated as an upper bound (Table~\ref{tab:classification-baseline}). BioGPT attains the best accuracy ($64.40\%$), with DeBERTa close behind ($63.79\%$). The small margin between the top models, together with the narrow overall range ($57.76\%$--$64.41\%$), suggests that Med-TC is not dominated by model capacity differences but by the shared ability of strong Transformers to capture abstract-level topical and clinical cues. Notably, domain-adapted models (BioBERT, ClinicalBERT, BioGPT) cluster near the top, indicating that pretraining on biomedical or clinical corpora confers a consistent, albeit modest, advantage in recognizing medically salient lexical and compositional patterns. Accuracy and $F_1$ track closely across backbones, implying that improvements largely reflect broad reductions in classification error rather than a precision-heavy or recall-heavy operating point shift.

\begin{table}[!t]
\scriptsize
\centering
\renewcommand{\arraystretch}{1.2}
\setlength{\tabcolsep}{6pt}
\begin{adjustbox}{width=\linewidth}
\begin{tabular}{l | cccc}
    \toprule
        \textbf{Model} & \textbf{Accuracy (\%)} & \textbf{F\textsubscript{1}-score (\%)} & \textbf{Precision (\%)} & \textbf{Recall} (\%) \\
    \hline
        \textit{BERT} & 61.84 & 61.40 & 61.33 & 61.84 \\
        
        \textit{RoBERTa} & 57.76 & 57.44 & 57.32 & 57.76 \\
        
        \textit{DeBERTa} & \underline{63.79} & 62.24 & \underline{63.91} & \underline{63.79} \\
        
        \textit{GPT-2} & 61.64 & 60.58 & 61.00 & 61.64 \\
        
        \textit{BioBERT} & 63.06 & \underline{62.80} & 62.84 & 63.06 \\
        
        \textit{ClinicalBERT} & 62.54 & 61.56 & 62.28 & 62.54 \\
        
        \textit{BioGPT} & \textbf{64.40} & \textbf{63.12} & \textbf{64.45} & \textbf{64.40} \\
    \bottomrule
    \end{tabular}
    \end{adjustbox}

\caption{{Baseline classification performance on Med-TC with plaintext data}.\protect\symfootmark}
\label{tab:classification-baseline}
\vspace{-15pt}
\end{table}

\symfoottext

Table~\ref{tab:classification-crypto} characterizes the privacy-utility trade-off under cryptographic transformations, contrasting Fully Secured (FS), where all tokens are transformed, with \texttt{HERALD} (PS), where only sensitive spans are protected. FS consistently collapses performance because it removes both medical content and the surrounding linguistic scaffold that encoders rely on to disambiguate topic and intent; the remaining signal is largely limited to residual structural artifacts of the transformation and dataset priors. In contrast, PS preserves most non-sensitive context (syntax, discourse glue, and generic clinical framing), allowing models to condition on intact sentence structure while also learning consistent representations for ciphertext identities. This yields systematic utility recovery: part of the signal comes from preserved context, and the other part of the signal is recovered when encrypted types recur and the model can associate those stable ciphertext forms with labels or downstream targets. For example, with FPE on ClinicalBERT, PS improves accuracy from $49.21\%$ to $58.41\%$, indicating that a substantial fraction of discriminative signal in Med-TC is recoverable using context and high-priority concealed entities.

The residual gap to plaintext remains meaningful (e.g., $58.41\%$ vs.\ $62.54\%$ for ClinicalBERT), which is expected because Med-TC labels depend directly on specialized entities (diagnoses, procedures, and condition descriptors) that are preferentially encrypted under PS. Differences across cryptographic methods are also diagnostic: approaches that better control tokenization disruption and preserve consistent surface regularities tend to yield higher PS utility. FPE and Fuzzy Hash form the strongest cluster, suggesting that learnability benefits when protected tokens map to representations that remain ``tokenizer-friendly'' and reduce effective vocabulary fragmentation. Classical hashes, especially SHA-256, trail because their high-entropy, long-form outputs provide little reusable substructure for representation learning, amplifying sparsity under both FS and PS. Across models, ClinicalBERT is frequently among the best under PS, consistent with the hypothesis that in-domain encoders can exploit preserved clinical context more effectively.

\begin{table*}[!tb]
\scriptsize
\centering
\setlength{\tabcolsep}{6pt}
\renewcommand{\arraystretch}{0.8}
\begin{adjustbox}{width=\linewidth}
\begin{tabular}{ll | cccc | cccc}
\toprule
    \multirow{2}{*}{\textbf{Cryptographic Methods}} &
    \multirow{2}{*}{\textbf{Model}} &
    \multicolumn{4}{c}{\textbf{FS}} &
    \multicolumn{4}{c}{\textbf{\texttt{HERALD} (PS)}} \\ 
    \cmidrule(lr){3-6}\cmidrule(lr){7-10}
    & & Accuracy & F\textsubscript{1}-score & Precision & Recall & Accuracy & F\textsubscript{1}-score & Precision & Recall \\
    \midrule
\multirow{7}{*}{\textbf{AES (\(ECB\))}} & \textit{BERT-cased} & 42.64 & 42.41 & 42.32 & 42.64 & 51.42 & 50.99 & 50.69 & 51.42 \\
& \textit{RoBERTa-cased} & 46.00 & 45.21 & 45.25 & 46.00 & 54.54 & 53.53 & 54.70 & 54.54 \\
& \textit{DeBERTa} & 43.91 & 38.76 & 36.04 & 43.91 & 51.94 & 47.77 & 45.44 & 51.94 \\
& \textit{GPT-2} & 43.15 & 40.85 & 41.16 & 43.15 & 55.51 & 54.50 & 54.37 & 55.51 \\
& \textit{BioBERT-cased} & \textbf{47.18} & \textbf{46.68} & \underline{46.62} & \textbf{47.18} & 53.84 & 52.85 & 52.85 & 53.84 \\
& \textit{ClinicalBERT} & 46.32 & 45.84 & 45.82 & 46.32 & \textbf{57.20} & \textbf{56.53} & \underline{56.38} & \textbf{57.20} \\
& \textit{BioGPT} & \underline{46.50} & \underline{46.22} & \textbf{46.85} & \underline{46.50} & \underline{56.34} & \underline{56.23} & \textbf{56.79} & \underline{56.34} \\
\midrule
\multirow{7}{*}{\textbf{AES (\(SIV\))}} & \textit{BERT-cased} & 42.06 & 41.77 & 41.62 & 42.06 & 52.60 & 51.70 & 52.26 & 52.60 \\
& \textit{RoBERTa-cased} & \textbf{46.77} & \textbf{46.46} & \textbf{47.15} & \textbf{46.77} & 53.67 & 53.29 & 53.35 & 53.67 \\
& \textit{DeBERTa} & 44.35 & 42.15 & 45.03 & 44.35 & 51.04 & 48.18 & 48.95 & 51.04 \\
& \textit{GPT-2} & 43.79 & 43.12 & 43.83 & 43.79 & \underline{54.05} & \underline{53.71} & \underline{53.81} & \underline{54.05} \\
& \textit{BioBERT-cased} & 45.56 & 45.27 & 45.65 & 45.56 & 52.87 & 52.73 & 52.99 & 52.87 \\
& \textit{ClinicalBERT} & \underline{46.56} & \underline{46.07} & \underline{46.73} & \underline{46.56} & \textbf{56.89} & \textbf{56.41} & \textbf{56.29} & \textbf{56.89} \\
& \textit{BioGPT} & 45.48 & 44.93 & 45.24 & 45.48 & 53.81 & 53.41 & 53.64 & 53.81 \\
\midrule
\multirow{7}{*}{\textbf{Blowfish (\(ECB\))}} & \textit{BERT-cased} & 44.50 & 44.06 & 43.79 & 44.50 & 56.09 & \underline{55.44} & 55.14 & 56.09 \\
& \textit{RoBERTa-cased} & 46.83 & 46.55 & \underline{46.98} & 46.83 & 55.71 & 54.91 & \underline{56.10} & 55.71 \\
& \textit{DeBERTa} & 42.62 & 37.61 & 34.99 & 42.62 & 54.47 & 52.21 & 53.48 & 54.47 \\
& \textit{GPT-2} & 43.15 & 41.79 & 42.57 & 43.15 & 53.88 & 51.64 & 53.06 & 53.88 \\
& \textit{BioBERT-cased} & 46.53 & 45.94 & 45.72 & 46.53 & 54.50 & 53.74 & 54.27 & 54.50 \\
& \textit{ClinicalBERT} & \textbf{48.62} & \textbf{47.78} & \textbf{48.00} & \textbf{48.62} & \textbf{57.34} & \textbf{56.29} & \textbf{56.11} & \textbf{57.34} \\
& \textit{BioGPT} & \underline{47.41} & \underline{46.74} & 46.76 & \underline{47.41} & \underline{56.13} & 55.15 & 55.60 & \underline{56.13} \\
\midrule
\multirow{7}{*}{\textbf{FPE}} & \textit{BERT-cased} & 46.47 & 45.63 & 45.81 & 46.47 & 56.23 & 55.17 & 55.08 & 56.23 \\
& \textit{RoBERTa-cased} & 46.24 & 45.81 & 45.86 & 46.24 & 53.95 & 53.54 & 53.85 & 53.95 \\
& \textit{DeBERTa} & 43.38 & 38.18 & 34.77 & 43.38 & 52.46 & 47.12 & 49.01 & 52.46 \\
& \textit{GPT-2} & \underline{48.65} & \underline{47.90} & \underline{47.78} & \underline{48.65} & \underline{57.86} & \underline{57.24} & \underline{57.01} & \underline{57.86} \\
& \textit{BioBERT-cased} & 44.68 & 43.98 & 44.23 & 44.68 & 56.23 & 55.86 & 56.21 & 56.23 \\
& \textit{ClinicalBERT} & \textbf{49.21} & \textbf{48.36} & \textbf{48.23} & \textbf{49.21} & \textbf{58.41} & \textbf{57.50} & \textbf{57.43} & \textbf{58.41} \\
& \textit{BioGPT} & 47.45 & 47.04 & 47.12 & 47.45 & 57.65 & 57.07 & 56.96 & 57.65 \\
\midrule
\multirow{7}{*}{\textbf{Fuzzy Hash}} & \textit{BERT-cased} & 47.85 & 47.59 & 47.50 & 47.85 & 56.16 & \underline{55.55} & \underline{56.34} & 56.16 \\
& \textit{RoBERTa-cased} & \underline{48.42} & \underline{47.97} & \underline{47.77} & \underline{48.42} & 53.22 & 52.52 & 52.77 & 53.22 \\
& \textit{DeBERTa} & 47.57 & 45.81 & 46.61 & 47.57 & 53.60 & 53.13 & 53.75 & 53.60 \\
& \textit{GPT-2} & 48.27 & 47.38 & 47.23 & 48.27 & \underline{56.48} & 55.17 & 55.27 & \underline{56.48} \\
& \textit{BioBERT-cased} & 48.12 & 47.33 & 47.18 & 48.12 & 55.78 & 54.60 & 54.73 & 55.78 \\
& \textit{ClinicalBERT} & \textbf{49.44} & \textbf{48.93} & \textbf{48.86} & \textbf{49.44} & \textbf{58.21} & \textbf{56.90} & \textbf{57.00} & \textbf{58.21} \\
& \textit{BioGPT} & 47.33 & 46.78 & 47.31 & 47.33 & 54.74 & 53.91 & 53.87 & 54.74 \\
\midrule
\multirow{7}{*}{\textbf{Soft Hash}} & \textit{BERT-cased} & 41.79 & 36.82 & 34.55 & 41.79 & 42.76 & 42.12 & 42.26 & 42.76 \\
& \textit{RoBERTa-cased} & \underline{44.91} & \underline{44.27} & \textbf{44.94} & \underline{44.91} & 51.28 & 50.68 & 50.98 & 51.28 \\
& \textit{DeBERTa} & 36.99 & 32.31 & 29.90 & 36.99 & 49.76 & 43.80 & 44.92 & 49.76 \\
& \textit{GPT-2} & 39.38 & 35.09 & 40.24 & 39.38 & \underline{52.04} & \underline{50.82} & \underline{51.89} & \underline{52.04} \\
& \textit{BioBERT-cased} & 42.91 & 38.90 & 38.69 & 42.91 & 47.58 & 46.36 & 46.69 & 47.58 \\
& \textit{ClinicalBERT} & 41.11 & 35.92 & 33.06 & 41.11 & \textbf{56.51} & \textbf{55.35} & \textbf{56.10} & \textbf{56.51} \\
& \textit{BioGPT} & \textbf{45.36} & \textbf{44.84} & \underline{44.77} & \textbf{45.36} & 46.81 & 43.95 & 47.67 & 46.81 \\
\midrule
\multirow{7}{*}{\textbf{MD5}} & \textit{BERT-cased} & 46.77 & 46.38 & 46.30 & 46.77 & 52.18 & 51.74 & 51.51 & 52.18 \\
& \textit{RoBERTa-cased} & 46.65 & 46.09 & 46.06 & 46.65 & 55.12 & 54.79 & 55.31 & 55.12 \\
& \textit{DeBERTa} & 43.15 & 37.89 & 34.99 & 43.15 & 51.49 & 47.99 & 49.90 & 51.49 \\
& \textit{GPT-2} & 44.27 & 42.33 & 43.40 & 44.27 & 51.47 & 47.98 & 49.91 & 51.47 \\
& \textit{BioBERT-cased} & 47.15 & \underline{46.78} & \underline{47.53} & 47.15 & \underline{56.65} & \underline{56.08} & \underline{56.24} & \underline{56.65} \\
& \textit{ClinicalBERT} & \underline{47.35} & 46.55 & 46.69 & \underline{47.35} & \textbf{57.69} & \textbf{57.04} & \textbf{57.30} & \textbf{57.69} \\
& \textit{BioGPT} & \textbf{47.95} & \textbf{47.71} & \textbf{47.63} & \textbf{47.95} & 56.41 & 55.77 & 55.68 & 56.41 \\
\midrule
\multirow{7}{*}{\textbf{SHA-1}} & \textit{BERT-cased} & 42.06 & 41.54 & 41.27 & 42.06 & 51.35 & 50.89 & 50.64 & 51.35 \\
& \textit{RoBERTa-cased} & 46.83 & 45.93 & \underline{46.52} & 46.83 & \underline{57.06} & \underline{56.46} & \underline{56.67} & \underline{57.06} \\
& \textit{DeBERTa} & 42.76 & 37.62 & 34.67 & 42.76 & 50.93 & 46.21 & 45.60 & 50.93 \\
& \textit{GPT-2} & 45.53 & 44.84 & 45.02 & 45.53 & 54.02 & 52.46 & 54.28 & 54.02 \\
& \textit{BioBERT-cased} & 45.92 & 45.42 & 45.81 & 45.92 & 53.25 & 52.35 & 52.38 & 53.25 \\
& \textit{ClinicalBERT} & \underline{47.27} & \underline{46.50} & 46.44 & \underline{47.27} & \textbf{57.69} & \textbf{56.95} & \textbf{56.96} & \textbf{57.69} \\
& \textit{BioGPT} & \textbf{47.71} & \textbf{47.42} & \textbf{48.03} & \textbf{47.71} & 55.82 & 55.52 & 55.54 & 55.82 \\
\midrule
\multirow{7}{*}{\textbf{SHA-256}} & \textit{BERT-cased} & 39.47 & \underline{38.68} & 39.82 & 39.47 & 45.67 & 43.61 & 45.66 & 45.67 \\
& \textit{RoBERTa-cased} & 39.12 & 34.59 & 36.30 & 39.12 & \textbf{50.31} & \textbf{49.98} & \underline{51.18} & \textbf{50.31} \\
& \textit{DeBERTa} & 37.23 & 32.30 & 31.79 & 37.23 & 40.69 & 35.59 & 34.96 & 40.69 \\
& \textit{GPT-2} & 37.91 & 33.60 & 39.70 & 37.91 & 49.10 & 46.34 & 48.40 & 49.10 \\
& \textit{BioBERT-cased} & \underline{40.55} & \textbf{40.08} & \textbf{41.62} & \underline{40.55} & \underline{49.97} & \underline{49.14} & 50.52 & \underline{49.97} \\
& \textit{ClinicalBERT} & \textbf{40.77} & 36.44 & 37.24 & \textbf{40.77} & 49.76 & 46.22 & \textbf{53.75} & 49.76 \\
& \textit{BioGPT} & 39.30 & 38.28 & \underline{40.20} & 39.30 & 48.16 & 47.38 & 48.14 & 48.16 \\
    \bottomrule
\end{tabular}
\end{adjustbox}
\caption{\textbf{Classification under cryptographic transformations.} Comparison of Fully Secured (FS) vs. \texttt{HERALD} (Partially Secured, PS) settings across ciphers.\protect\symfootmark}
\vspace{-15pt}
\label{tab:classification-crypto}
\end{table*}

\subsubsection{MCQ Task}
We evaluate LLMs on clinical MCQ benchmarks. We exclude AES in ECB mode, Soft Hash, and SHA-256 as mechanisms for producing secured-text representations (for MCQ task) because they entail unacceptable security–utility trade-offs in our setting. Collectively, these methods either weaken privacy guarantees (ECB) or sacrifice downstream fidelity~(low performance score) and computational efficiency (Soft Hash, SHA-256), making them ill-suited for high-stakes NLP under the evaluation regimes used here. We consider performance (non-generative) metrics for easier and fairer comparison. Because of limited computational resources, generative-metric results are provided in Appendix~\ref{app:gen-modes}. The inferred generative results largely recover the baseline, while the fully secured workflow collapses again.

On \textbf{MedMCQA} baselines (Table~\ref{tab:MedMCQA-baseline}), the two domain-adapted Llama variants lead, with Llama-3-Med42-8B at $76.51\%$ accuracy and Llama-3.1-Aloe-Beta-8B at $75.87\%$. The small $0.65$-point gap suggests that, in plaintext, model-specific differences are secondary to shared capacity and domain alignment, whereas the larger separation to general-purpose counterparts (e.g., Qwen-2.5-7B at $71.93\%$ and Llama-3.1-8B at $70.22\%$) indicates a tangible benefit from biomedical specialization.

\begin{table}[!tb]
\scriptsize
    \centering
    \renewcommand{\arraystretch}{1.3}
    \setlength{\tabcolsep}{6pt}
    \begin{adjustbox}{width=\linewidth}
    \begin{tabular}{l | cccc}
    \toprule
        \textbf{Model} & \textbf{Accuracy (\%)} & \textbf{F\textsubscript{1}-score (\%)} & \textbf{Precision (\%)} & \textbf{Recall} (\%) \\
    \hline
        \textit{Qwen-2.5-7B} & 71.93 & 71.94 & 72.08 & 71.93 \\

        \textit{Mistral-7B-Instruct-v0.3} & 63.59 & 63.50 & 63.56 & 63.59 \\
        
        \textit{Llama-3.1-8B} & 70.22 & 70.20 & 70.26 & 70.22 \\
        
        \textit{Llama-3.1-Aloe-Beta-8B} & \underline{75.87} & \underline{75.89} & \underline{75.94} & \underline{75.87} \\
        
        \textit{Llama-3-Med42-8B} & \textbf{76.51} & \textbf{76.52} & \textbf{76.57} & \textbf{76.51} \\
    \bottomrule
    \end{tabular}
    \end{adjustbox}
\caption{{Baseline performance on MedMCQA with plaintext data.\protect\symfootmark}}
\vspace{-15pt}
\label{tab:MedMCQA-baseline}
\end{table}

Under full security for MedMCQA (Table~\ref{tab:MedMCQA-full-crypto}), performance degrades for two distinct reasons tied to prompt semantics. In FS+OU, answer options remain readable, but the stem is entirely transformed, so the model loses the clinical facts needed to discriminate among plausible distractors; the best configuration reaches only $53.76\%$ (Med42 with SHA-1), far below plaintext. In FS+OFS, both stem and options are secured, which removes not only the patient-linked content but also the meaning of the candidate answers themselves; accuracies concentrate near the four-choice chance level, with the best at $39.39\%$ (Qwen with Fuzzy Hash). This separation between OU and OFS indicates that, for MCQ, preserving the option text is not a minor convenience but a primary driver of solvable supervision. Cipher choice still affects outcomes, but the absence of a consistent winner suggests a bottleneck at the level of linguistic utility rather than cryptographic primitive.

\begin{figure*}[!tb]
    \centering
    \includegraphics[width=0.95\linewidth]{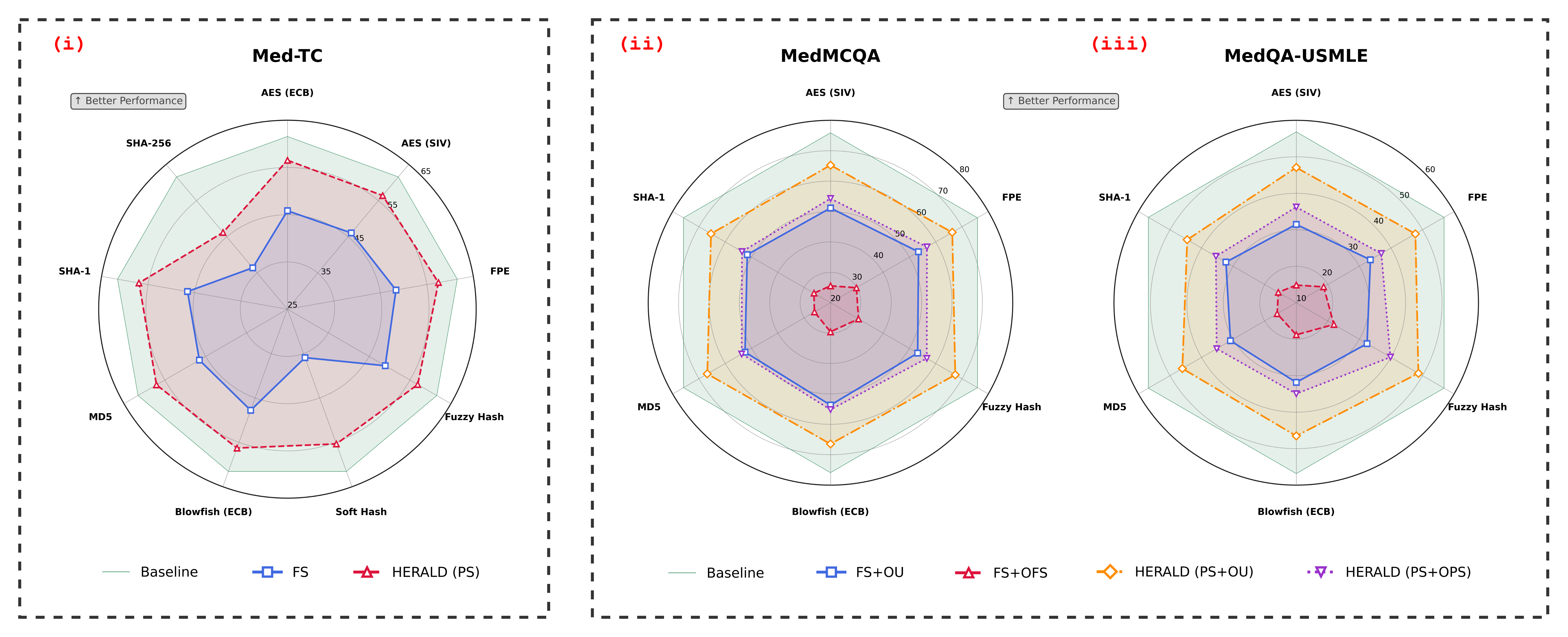}
    \caption{\textbf{Utility under full vs.\ selective security.} Radar plots report test F\textsubscript{1} for \emph{ClinicalBERT}/Med-TC, \emph{Llama-3.1-Aloe-Beta-8B}/MedMCQA, and \emph{Qwen-2.5-7B}/MedQA-USMLE. The outer green ring is plaintext; inner polygons compare \textbf{FS} (all tokens secured) with \textbf{\texttt{HERALD} (PS)}. For MCQ, options are either plaintext (\textbf{OU}) or secured (\textbf{OFS}/\textbf{OPS}).}
    \vspace{-5pt}
    \label{fig:radar-plots}
\end{figure*}

\begin{table*}[!tb]
\scriptsize
\centering
\setlength{\tabcolsep}{6pt}
\renewcommand{\arraystretch}{0.8}
\begin{adjustbox}{width=\linewidth}
\begin{tabular}{ll | cccc | cccc}
\toprule
    \multirow{2}{*}{\textbf{Cryptographic Methods}} &
    \multirow{2}{*}{\textbf{Model}} &
    \multicolumn{4}{c}{\textbf{FS + OU}} &
    \multicolumn{4}{c}{\textbf{FS + OFS}} \\ 
    \cmidrule(lr){3-6}\cmidrule(lr){7-10}
    & & Accuracy & F\textsubscript{1}-score & Precision & Recall & Accuracy & F\textsubscript{1}-score & Precision & Recall \\
    \midrule
\multirow{5}{*}{\textbf{AES (\(SIV\))}} & \textit{Qwen-2.5-7B} & 50.53 & 50.22 & 50.35 & 50.53 & 36.14 & \textbf{29.93} & 34.97 & 36.14 \\
& \textit{Mistral-7B-Instruct-v0.3} & 37.08 & 34.21 & 37.05 & 37.08 & \underline{36.38} & 25.97 & \textbf{37.28} & \underline{36.38} \\
& \textit{Llama-3.1-8B} & 45.82 & 45.77 & 45.83 & 45.82 & 29.13 & 28.88 & 28.74 & 29.13 \\
& \textit{Llama3.1-Aloe-Beta-8B} & \underline{51.18} & \underline{51.15} & \underline{51.33} & \underline{51.18} & 35.72 & 25.55 & 25.54 & 35.72 \\
& \textit{Llama3-Med42-8B} & \textbf{53.47} & \textbf{53.36} & \textbf{53.57} & \textbf{53.47} & \textbf{37.03} & \underline{29.87} & \underline{36.47} & \textbf{37.03} \\
\midrule
\multirow{5}{*}{\textbf{Blowfish (\(ECB\))}} & \textit{Qwen-2.5-7B} & 51.05 & 50.86 & 50.90 & 51.05 & \underline{37.00} & \textbf{31.72} & \underline{35.97} & \underline{37.00} \\
& \textit{Mistral-7B-Instruct-v0.3} & 35.69 & 31.79 & 34.84 & 35.69 & 36.20 & 25.72 & 31.14 & 36.20 \\
& \textit{Llama-3.1-8B} & 44.88 & 44.70 & 44.63 & 44.88 & 30.14 & 29.72 & 29.49 & 30.14 \\
& \textit{Llama3.1-Aloe-Beta-8B} & \underline{52.70} & \underline{52.68} & \underline{52.82} & \underline{52.70} & 36.88 & 29.56 & 34.77 & 36.88 \\
& \textit{Llama3-Med42-8B} & \textbf{52.84} & \textbf{52.76} & \textbf{52.85} & \textbf{52.84} & \textbf{37.96} & \underline{31.04} & \textbf{36.38} & \textbf{37.96} \\
\midrule
\multirow{5}{*}{\textbf{FPE}} & \textit{Qwen-2.5-7B} & 52.00 & 51.82 & 51.88 & 52.00 & 36.77 & \underline{31.13} & 33.88 & 36.77 \\
& \textit{Mistral-7B-Instruct-v0.3} & 37.58 & 35.22 & 36.78 & 37.58 & \underline{36.94} & 28.63 & 34.02 & \underline{36.94} \\
& \textit{Llama-3.1-8B} & 39.47 & 39.29 & 39.21 & 39.47 & 30.68 & 30.40 & 30.24 & 30.68 \\
& \textit{Llama3.1-Aloe-Beta-8B} & \textbf{53.55} & \textbf{53.48} & \textbf{53.53} & \textbf{53.55} & 36.79 & 29.86 & \underline{35.66} & 36.79 \\
& \textit{Llama3-Med42-8B} & \underline{53.10} & \underline{53.06} & \underline{53.28} & \underline{53.10} & \textbf{38.59} & \textbf{32.82} & \textbf{38.19} & \textbf{38.59} \\
\midrule
\multirow{5}{*}{\textbf{Fuzzy Hash}} & \textit{Qwen-2.5-7B} & 52.76 & 52.65 & 52.71 & 52.76 & \textbf{39.39} & \textbf{36.05} & \underline{38.11} & \textbf{39.39} \\
& \textit{Mistral-7B-Instruct-v0.3} & 38.82 & 36.66 & 38.72 & 38.82 & \underline{37.93} & 30.84 & \textbf{38.19} & \underline{37.93} \\
& \textit{Llama-3.1-8B} & 43.65 & 43.27 & 43.21 & 43.65 & 30.77 & 30.47 & 30.27 & 30.77 \\
& \textit{Llama3.1-Aloe-Beta-8B} & \underline{53.18} & \underline{53.14} & \underline{53.28} & \underline{53.18} & 37.60 & 30.66 & 36.09 & 37.60 \\
& \textit{Llama3-Med42-8B} & \textbf{53.60} & \textbf{53.56} & \textbf{53.64} & \textbf{53.60} & 37.46 & \underline{31.94} & 35.10 & 37.46 \\
\midrule
\multirow{5}{*}{\textbf{MD5}} & \textit{Qwen-2.5-7B} & 50.38 & 50.33 & 50.51 & 50.38 & 35.89 & \textbf{29.68} & 33.17 & 35.89 \\
& \textit{Mistral-7B-Instruct-v0.3} & 38.00 & 35.41 & 37.89 & 38.00 & \underline{36.35} & 26.70 & \textbf{38.61} & \underline{36.35} \\
& \textit{Llama-3.1-8B} & 45.57 & 45.38 & 45.30 & 45.57 & 29.72 & \underline{29.21} & 29.00 & 29.72 \\
& \textit{Llama3.1-Aloe-Beta-8B} & \underline{52.54} & \underline{52.48} & \underline{52.62} & \underline{52.54} & 35.96 & 26.10 & 32.33 & 35.96 \\
& \textit{Llama3-Med42-8B} & \textbf{53.18} & \textbf{53.08} & \textbf{53.36} & \textbf{53.18} & \textbf{36.81} & 27.60 & \underline{33.57} & \textbf{36.81} \\
\midrule
\multirow{5}{*}{\textbf{SHA-1}} & \textit{Qwen-2.5-7B} & 49.68 & 49.55 & 49.73 & 49.68 & 35.83 & \underline{27.31} & 32.46 & 35.83 \\
& \textit{Mistral-7B-Instruct-v0.3} & 35.76 & 31.77 & 35.20 & 35.76 & \underline{36.02} & 25.84 & \textbf{42.19} & \underline{36.02} \\
& \textit{Llama-3.1-8B} & 43.99 & 43.96 & 43.93 & 43.99 & 30.23 & \textbf{29.70} & 29.45 & 30.23 \\
& \textit{Llama3.1-Aloe-Beta-8B} & \underline{51.68} & \underline{51.64} & \underline{51.84} & \underline{51.68} & 35.96 & 26.29 & \underline{34.69} & 35.96 \\
& \textit{Llama3-Med42-8B} & \textbf{53.76} & \textbf{53.73} & \textbf{54.00} & \textbf{53.76} & \textbf{36.30} & 27.01 & 34.58 & \textbf{36.30} \\
    \bottomrule
\end{tabular}
\end{adjustbox}
\vspace{-10pt}
\caption{{Performance on MedMCQA with fully secured tokens.}\protect\symfootmark}
\vspace{-10pt}
\label{tab:MedMCQA-full-crypto}
\end{table*}

\texttt{HERALD}’s partial security on MedMCQA (Table~\ref{tab:MedMCQA-partial-crypto}) substantially closes the gap by retaining most non-sensitive syntax and discourse while replacing only prioritized tokens with stable ciphertext token identities. The best PS+OU reaches $68.25\%$ (Med42 with FPE), recovering a large portion of the loss observed under FS+OU and approaching the plaintext ceiling within a single-digit margin. In contrast, securing answer options under PS+OPS still incurs a large penalty ($68.25\%\to57.65\%$ for the same model and cipher), reinforcing that option semantics act as task-critical anchors; once obfuscated, the model must infer the correct label without access to the candidate answers' meaning, which predictably increases ambiguity. Across ciphers, PS+OU consistently exceeds PS+OPS, and the strongest results cluster around FPE and Fuzzy Hash. These transforms better preserve surface regularities and tokenization stability than high-entropy encodings. Model rankings also partially re-emerge under PS+OU: medically adapted Llamas regain clear advantages over the general baselines, indicating that domain knowledge remains usable when the prompt retains readable structure, even if sensitive spans are concealed.

\begin{table*}[!tb]
\scriptsize
\centering
\setlength{\tabcolsep}{6pt}
\renewcommand{\arraystretch}{0.8}
\begin{adjustbox}{width=\linewidth}
\begin{tabular}{ll | cccc | cccc}
\toprule
    \multirow{2}{*}{\textbf{Cryptographic Methods}} &
    \multirow{2}{*}{\textbf{Model}} &
    \multicolumn{4}{c}{\textbf{\texttt{HERALD} (PS + OU)}} &
    \multicolumn{4}{c}{\textbf{\texttt{HERALD} (PS + OPS)}} \\ 
    \cmidrule(lr){3-6}\cmidrule(lr){7-10}
    & & Accuracy & F\textsubscript{1}-score & Precision & Recall & Accuracy & F\textsubscript{1}-score & Precision & Recall \\
    \midrule
\multirow{5}{*}{\textbf{AES (\(SIV\))}} & \textit{Qwen-2.5-7B} & 63.21 & 63.14 & 63.17 & 63.21 & 50.76 & 50.32 & 50.58 & 50.76 \\
& \textit{Mistral-7B-Instruct-v0.3} & 51.69 & 51.01 & 51.34 & 51.69 & 41.39 & 37.19 & 41.40 & 41.39 \\
& \textit{Llama-3.1-8B} & 58.07 & 57.99 & 57.97 & 58.07 & 45.27 & 44.83 & 44.67 & 45.27 \\
& \textit{Llama-3.1-Aloe-Beta-8B} & \underline{65.29} & \underline{65.25} & \underline{65.50} & \underline{65.29} & \underline{54.61} & \underline{54.27} & \underline{54.51} & \underline{54.61} \\
& \textit{Llama-3-Med42-8B} & \textbf{67.23} & \textbf{67.16} & \textbf{67.23} & \textbf{67.23} & \textbf{56.13} & \textbf{55.81} & \textbf{55.94} & \textbf{56.13} \\
\midrule
\multirow{5}{*}{\textbf{Blowfish (\(ECB\))}} & \textit{Qwen-2.5-7B} & 65.09 & 65.01 & 65.18 & 65.09 & 54.79 & 54.38 & 54.42 & 54.79 \\
& \textit{Mistral-7B-Instruct-v0.3} & 51.12 & 50.28 & 50.79 & 51.12 & 43.15 & 40.29 & 42.98 & 43.15 \\
& \textit{Llama-3.1-8B} & 60.97 & 60.86 & 60.84 & 60.97 & 43.24 & 42.93 & 42.83 & 43.24 \\
& \textit{Llama-3.1-Aloe-Beta-8B} & \underline{66.55} & \underline{66.48} & \underline{66.64} & \underline{66.55} & \underline{55.48} & \underline{55.17} & \underline{55.23} & \underline{55.48} \\
& \textit{Llama-3-Med42-8B} & \textbf{66.99} & \textbf{66.92} & \textbf{67.05} & \textbf{66.99} & \textbf{56.01} & \textbf{55.68} & \textbf{55.81} & \textbf{56.01} \\
\midrule
\multirow{5}{*}{\textbf{FPE}} & \textit{Qwen-2.5-7B} & 63.12 & 63.06 & 63.20 & 63.12 & 54.94 & 54.65 & 54.77 & 54.94 \\
& \textit{Mistral-7B-Instruct-v0.3} & 53.00 & 52.45 & 52.80 & 53.00 & 44.17 & 41.91 & 44.18 & 44.17 \\
& \textit{Llama-3.1-8B} & 59.80 & 59.80 & 59.83 & 59.80 & 50.43 & 50.08 & 50.01 & 50.43 \\
& \textit{Llama-3.1-Aloe-Beta-8B} & \underline{66.40} & \underline{66.30} & \underline{66.48} & \underline{66.40} & \underline{56.97} & \underline{56.67} & \underline{56.92} & \underline{56.97} \\
& \textit{Llama-3-Med42-8B} & \textbf{68.25} & \textbf{68.24} & \textbf{68.38} & \textbf{68.25} & \textbf{57.65} & \textbf{57.48} & \textbf{57.48} & \textbf{57.65} \\
\midrule
\multirow{5}{*}{\textbf{Fuzzy Hash}} & \textit{Qwen-2.5-7B} & \underline{64.94} & \underline{64.90} & \underline{65.02} & \underline{64.94} & \underline{55.83} & \underline{55.58} & \underline{55.70} & \underline{55.83} \\
& \textit{Mistral-7B-Instruct-v0.3} & 51.92 & 51.41 & 51.70 & 51.92 & 44.73 & 42.79 & 44.00 & 44.73 \\
& \textit{Llama-3.1-8B} & 61.15 & 61.11 & 61.12 & 61.15 & 49.36 & 48.96 & 48.96 & 49.36 \\
& \textit{Llama-3.1-Aloe-Beta-8B} & \textbf{67.44} & \textbf{67.43} & \textbf{67.65} & \textbf{67.44} & \textbf{56.73} & \textbf{56.61} & \textbf{56.64} & \textbf{56.73} \\
& \textit{Llama-3-Med42-8B} & 60.19 & 56.14 & 58.95 & 60.19 & 52.69 & 49.21 & 50.00 & 52.69 \\
\midrule
\multirow{5}{*}{\textbf{MD5}} & \textit{Qwen-2.5-7B} & 62.10 & 62.03 & 62.07 & 62.10 & 51.39 & 50.79 & 51.30 & 51.39 \\
& \textit{Mistral-7B-Instruct-v0.3} & 51.69 & 51.01 & 51.45 & 51.69 & 43.99 & 41.64 & 43.47 & 43.99 \\
& \textit{Llama-3.1-8B} & 60.55 & 60.47 & 60.58 & 60.55 & 46.67 & 46.32 & 46.23 & 46.67 \\
& \textit{Llama-3.1-Aloe-Beta-8B} & \underline{66.88} & \underline{66.82} & \underline{67.13} & \underline{66.88} & \underline{54.22} & \underline{53.73} & \underline{54.31} & \underline{54.22} \\
& \textit{Llama-3-Med42-8B} & \textbf{67.89} & \textbf{67.91} & \textbf{68.13} & \textbf{67.89} & \textbf{54.58} & \textbf{54.27} & \textbf{54.53} & \textbf{54.58} \\
\midrule
\multirow{5}{*}{\textbf{SHA-1}} & \textit{Qwen-2.5-7B} & 61.06 & 61.02 & 61.07 & 61.06 & 48.76 & 47.97 & 48.41 & 48.76 \\
& \textit{Mistral-7B-Instruct-v0.3} & 52.49 & 51.90 & 52.22 & 52.49 & 45.66 & 44.43 & 45.04 & 45.66 \\
& \textit{Llama-3.1-8B} & 59.30 & 59.24 & 59.21 & 59.30 & 44.29 & 43.96 & 43.86 & 44.29 \\
& \textit{Llama-3.1-Aloe-Beta-8B} & \underline{65.50} & \underline{65.40} & \underline{65.60} & \underline{65.50} & \underline{54.01} & \underline{53.64} & \underline{53.70} & \underline{54.01} \\
& \textit{Llama-3-Med42-8B} & \textbf{67.92} & \textbf{67.88} & \textbf{68.04} & \textbf{67.92} & \textbf{54.55} & \textbf{54.14} & \textbf{54.21} & \textbf{54.55} \\
    \bottomrule
\end{tabular}
\end{adjustbox}
\vspace{-10pt}
\caption{{Performance on MedMCQA under \texttt{HERALD} partial securing.}\protect\symfootmark}
\label{tab:MedMCQA-partial-crypto}
\vspace{-10pt}
\end{table*}

For \textbf{MedQA-USMLE} baselines (Table~\ref{tab:MedQA-USMLE-baseline}), the leading models are separated by less than one point (Qwen-2.5-7B at $56.95\%$ vs.\ Llama-3.1-Aloe-Beta-8B at $56.25\%$), and all models concentrate in the mid-$50$s. This compression suggests that MedQA-USMLE is dominated by a dataset-level difficulty ceiling rather than large capacity gaps, with most backbones converging to similar error profiles. In contrast to MedMCQA, biomedical adaptation does not uniformly translate into a clear advantage here (e.g., Med42 trails despite domain specialization), consistent with an exam-style setting where multi-step reasoning and distractor discrimination limit gains from surface domain recall alone.

\begin{table}[!tb]
\scriptsize
    \centering
    \renewcommand{\arraystretch}{1.3}
    \setlength{\tabcolsep}{6pt}
    \begin{adjustbox}{width=\linewidth}
    \begin{tabular}{l | cccc}
    \toprule
        \textbf{Model} & \textbf{Accuracy (\%)} & \textbf{F\textsubscript{1}-score (\%)} & \textbf{Precision (\%)} & \textbf{Recall} (\%) \\
    \hline
        \textit{Qwen-2.5-7B} & \textbf{56.95} & \textbf{56.84} & \textbf{56.91} & \textbf{56.95} \\

        \textit{Mistral-7B-Instruct-v0.3} & 47.09 & 47.07 & 47.17 & 47.09 \\
        
        \textit{Llama-3.1-8B} & 53.26 & 53.30 & 53.49 & 53.26 \\
        
        \textit{Llama-3.1-Aloe-Beta-8B} & \underline{56.25} & \underline{56.12} & \underline{56.60} & \underline{56.25} \\
        
        \textit{Llama-3-Med42-8B} & 51.99 & 50.20 & 54.23 & 51.99 \\
    \bottomrule
    \end{tabular}
    \end{adjustbox}
\caption{{Baseline performance on MedQA-USMLE with plaintext data.}\protect\symfootmark}
\label{tab:MedQA-USMLE-baseline}
\vspace{-20pt}
\end{table}

Full security on MedQA-USMLE (Table~\ref{tab:MedQA-USMLE-full-crypto}) induces a markedly different regime than plaintext. In FS+OU, options remain readable but the stem is fully transformed, so models can only exploit superficial option priors and residual formatting cues; this yields a plateau in the low-to-mid $30$s, with limited separation across cryptographic methods and backbones. In FS+OFS, both stem and options are secured, which removes not only patient-linked content but also the meaning of the candidate answers; accuracies concentrate near the four-choice chance level ($\approx 25\%$), indicating that supervision becomes weakly informative once the label space is obfuscated and, in extreme cases, degenerate decoding (e.g., Med42 at $0.00\%$ under Blowfish and SHA-1). Differences across ciphers become second-order under this regime: even comparatively tokenizer-friendly transforms cannot compensate for the removal of interpretable anchors.

\begin{table*}[!tb]
\scriptsize
\centering
\setlength{\tabcolsep}{6pt}
\renewcommand{\arraystretch}{0.8}
\begin{adjustbox}{width=\linewidth}
\begin{tabular}{ll | cccc | cccc}
\toprule
    \multirow{2}{*}{\textbf{Cryptographic Methods}} &
    \multirow{2}{*}{\textbf{Model}} &
    \multicolumn{4}{c}{\textbf{FS + OU}} &
    \multicolumn{4}{c}{\textbf{FS + OFS}} \\ 
    \cmidrule(lr){3-6}\cmidrule(lr){7-10}
    & & Accuracy & F\textsubscript{1}-score & Precision & Recall & Accuracy & F\textsubscript{1}-score & Precision & Recall \\
    \midrule
\multirow{5}{*}{\textbf{AES (\(SIV\))}} & \textit{Qwen-2.5-7B} & 31.74 & 31.48 & 32.07 & 31.74 & 24.98 & 14.83 & 13.31 & 24.98 \\
& \textit{Mistral-7B-Instruct-v0.3} & 30.82 & 30.09 & 31.75 & 30.82 & 24.78 & \underline{20.26} & 22.57 & 24.78 \\
& \textit{Llama-3.1-8B} & 28.13 & 28.06 & 28.17 & 28.13 & 24.74 & \textbf{24.61} & \underline{24.70} & 24.74 \\
& \textit{Llama3.1-Aloe-Beta-8B} & \textbf{34.71} & \textbf{34.69} & \textbf{35.03} & \textbf{34.71} & \underline{25.37} & 16.76 & 13.61 & \underline{25.37} \\
& \textit{Llama3-Med42-8B} & \underline{34.08} & \underline{33.70} & \underline{33.99} & \underline{34.08} & \textbf{25.59} & 15.49 & \textbf{32.12} & \textbf{25.59} \\
\midrule
\multirow{5}{*}{\textbf{Blowfish (\(ECB\))}} & \textit{Qwen-2.5-7B} & 31.82 & 31.84 & 32.32 & 31.82 & \textbf{26.63} & 18.80 & 20.46 & \textbf{26.63} \\
& \textit{Mistral-7B-Instruct-v0.3} & 29.34 & 28.24 & 29.76 & 29.34 & 24.19 & \underline{20.24} & \underline{21.34} & 24.19 \\
& \textit{Llama-3.1-8B} & 28.49 & 28.44 & 28.60 & 28.49 & 24.67 & \textbf{24.49} & \textbf{24.62} & 24.67 \\
& \textit{Llama3.1-Aloe-Beta-8B} & \textbf{34.64} & \textbf{34.56} & \textbf{34.78} & \textbf{34.64} & \underline{25.61} & 17.33 & 13.77 & \underline{25.61} \\
& \textit{Llama3-Med42-8B} & \underline{34.48} & \underline{34.33} & \underline{34.64} & \underline{34.48} & 0.00 & 0.00 & 0.00 & 0.001 \\
\midrule
\multirow{5}{*}{\textbf{FPE}} & \textit{Qwen-2.5-7B} & 33.72 & 33.51 & 33.91 & 33.72 & 25.69 & 18.70 & 26.25 & 25.69 \\
& \textit{Mistral-7B-Instruct-v0.3} & 30.11 & 29.55 & 30.73 & 30.11 & \underline{26.16} & \underline{22.33} & \textbf{32.60} & \underline{26.16} \\
& \textit{Llama-3.1-8B} & 29.62 & 29.63 & 29.74 & 29.62 & 24.67 & \textbf{24.62} & 24.69 & 24.67 \\
& \textit{Llama3.1-Aloe-Beta-8B} & \textbf{34.64} & \textbf{34.65} & \textbf{35.14} & \textbf{34.64} & 25.84 & 20.38 & 25.90 & 25.84 \\
& \textit{Llama3-Med42-8B} & \underline{34.50} & \underline{34.45} & \underline{34.84} & \underline{34.50} & \textbf{26.71} & 19.98 & \underline{29.43} & \textbf{26.71} \\
\midrule
\multirow{5}{*}{\textbf{Fuzzy Hash}} & \textit{Qwen-2.5-7B} & \underline{32.53} & \underline{32.44} & \underline{32.52} & \underline{32.53} & \textbf{27.40} & 21.92 & \underline{26.82} & \textbf{27.40} \\
& \textit{Mistral-7B-Instruct-v0.3} & 29.62 & 28.98 & 30.17 & 29.62 & 24.59 & 22.35 & 23.36 & 24.59 \\
& \textit{Llama-3.1-8B} & 29.76 & 29.71 & 29.86 & 29.76 & 24.98 & \textbf{24.89} & 24.89 & 24.98 \\
& \textit{Llama3.1-Aloe-Beta-8B} & \textbf{34.86} & \textbf{34.71} & \textbf{35.06} & \textbf{34.86} & \underline{26.71} & \underline{23.56} & \textbf{26.96} & \underline{26.71} \\
& \textit{Llama3-Med42-8B} & 28.65 & 24.13 & 29.87 & 28.65 & 25.45 & 20.66 & 24.90 & 25.45 \\
\midrule
\multirow{5}{*}{\textbf{MD5}} & \textit{Qwen-2.5-7B} & 30.97 & 30.83 & 31.85 & 30.97 & 24.21 & 16.04 & \textbf{33.95} & 24.21 \\
& \textit{Mistral-7B-Instruct-v0.3} & 30.73 & 30.35 & 31.10 & 30.73 & \underline{25.65} & \underline{23.47} & 24.12 & \underline{25.65} \\
& \textit{Llama-3.1-8B} & 27.87 & 27.83 & 27.96 & 27.87 & 24.86 & \textbf{24.80} & \underline{24.96} & 24.86 \\
& \textit{Llama3.1-Aloe-Beta-8B} & \textbf{34.81} & \textbf{34.85} & \textbf{35.25} & \textbf{34.81} & \textbf{25.73} & 16.90 & 19.13 & \textbf{25.73} \\
& \textit{Llama3-Med42-8B} & \underline{32.79} & \underline{32.70} & \underline{33.00} & \underline{32.79} & 24.74 & 15.37 & 13.45 & 24.74 \\
\midrule
\multirow{5}{*}{\textbf{SHA-1}} & \textit{Qwen-2.5-7B} & 32.36 & 32.27 & 32.81 & 32.36 & 23.68 & 15.68 & 18.01 & 23.68 \\
& \textit{Mistral-7B-Instruct-v0.3} & 30.10 & 29.57 & 31.42 & 30.10 & 23.37 & \underline{20.52} & \textbf{26.78} & 23.37 \\
& \textit{Llama-3.1-8B} & 28.54 & 28.50 & 28.55 & 28.54 & \underline{24.15} & \textbf{24.13} & \underline{24.24} & \underline{24.15} \\
& \textit{Llama3.1-Aloe-Beta-8B} & \textbf{35.41} & \textbf{35.45} & \textbf{35.81} & \textbf{35.41} & \textbf{26.20} & 18.15 & 21.79 & \textbf{26.20} \\
& \textit{Llama3-Med42-8B} & \underline{33.80} & \underline{33.66} & \underline{33.85} & \underline{33.80} & 0.00 & 0.00 & 0.001 & 0.00 \\
    \bottomrule
\end{tabular}
\end{adjustbox}
\vspace{-10pt}
\caption{{Performance on MedQA-USMLE with fully secured tokens.}\protect\symfootmark}
\vspace{-10pt}
\label{tab:MedQA-USMLE-full-crypto}
\end{table*}

\texttt{HERALD} on MedQA-USMLE (Table~\ref{tab:MedQA-USMLE-partial-crypto}) recovers utility by preserving the syntactic and discourse scaffold that supports exam-style reasoning while learning representations of protected spans as interpretable information. Under PS+OU, top results concentrate around $\approx 49$-$50\%$, which indicates that a substantial fraction of MedQA signal is accessible from problem structure, relations, and non-sensitive descriptors even when prioritized entities are concealed. This recovery is not uniform across settings: ciphers that better preserve tokenizer-friendly regularities (notably FPE and Fuzzy Hash) tend to occupy the top of the PS+OU band. In contrast, PS+OPS consistently underperforms PS+OU, reflecting that option semantics function as anchors for mapping inferred clinical states to discrete choices; once the option surface forms are secured, the model must select among ciphertext labels with reduced semantic grounding, increasing ambiguity and amplifying minor prompt priors.

\begin{table*}[!tb]
\scriptsize
\centering
\setlength{\tabcolsep}{6pt}
\renewcommand{\arraystretch}{0.8}
\begin{adjustbox}{width=\linewidth}
\begin{tabular}{ll | cccc | cccc}
\toprule
    \multirow{2}{*}{\textbf{Cryptographic Methods}} &
    \multirow{2}{*}{\textbf{Model}} &
    \multicolumn{4}{c}{\textbf{\texttt{HERALD} (PS + OU)}} &
    \multicolumn{4}{c}{\textbf{\texttt{HERALD} (PS + OPS)}} \\ 
    \cmidrule(lr){3-6}\cmidrule(lr){7-10}
    & & Accuracy & F\textsubscript{1}-score & Precision & Recall & Accuracy & F\textsubscript{1}-score & Precision & Recall \\
    \midrule
\multirow{5}{*}{\textbf{AES (\(SIV\))}} & \textit{Qwen-2.5-7B} & \underline{47.06} & \underline{47.07} & \underline{47.12} & \underline{47.06} & \textbf{36.45} & \textbf{36.22} & \textbf{36.43} & \textbf{36.45} \\
& \textit{Mistral-7B-Instruct-v0.3} & 41.62 & 41.52 & 41.57 & 41.62 & 32.97 & 32.47 & 33.12 & 32.97 \\
& \textit{Llama-3.1-8B} & 40.38 & 40.40 & 40.47 & 40.38 & 28.95 & 28.94 & 28.98 & 28.95 \\
& \textit{Llama-3.1-Aloe-Beta-8B} & \textbf{48.00} & \textbf{48.00} & \textbf{48.23} & \textbf{48.00} & \underline{35.80} & \underline{35.70} & \underline{35.82} & \underline{35.80} \\
& \textit{Llama-3-Med42-8B} & 44.35 & 44.25 & 44.33 & 44.35 & 32.94 & 32.63 & 32.76 & 32.94 \\
\midrule
\multirow{5}{*}{\textbf{Blowfish (\(ECB\))}} & \textit{Qwen-2.5-7B} & 46.58 & 46.52 & 46.62 & 46.58 & \underline{35.04} & \underline{34.96} & \underline{35.05} & \underline{35.04} \\
& \textit{Mistral-7B-Instruct-v0.3} & 42.42 & 42.41 & 42.70 & 42.42 & 33.54 & 33.19 & 33.26 & 33.54 \\
& \textit{Llama-3.1-8B} & 41.48 & 41.48 & 41.61 & 41.48 & 30.40 & 30.35 & 30.49 & 30.40 \\
& \textit{Llama-3.1-Aloe-Beta-8B} & \underline{47.92} & \underline{47.95} & \underline{48.08} & \underline{47.92} & \textbf{36.45} & \textbf{36.47} & \textbf{36.86} & \textbf{36.45} \\
& \textit{Llama-3-Med42-8B} & \textbf{48.95} & \textbf{48.89} & \textbf{49.00} & \textbf{48.95} & 34.21 & 32.39 & 33.16 & 34.21 \\
\midrule
\multirow{5}{*}{\textbf{FPE}} & \textit{Qwen-2.5-7B} & 47.76 & 47.76 & 47.89 & 47.76 & 37.16 & 36.96 & 36.99 & 37.16 \\
& \textit{Mistral-7B-Instruct-v0.3} & 43.44 & 43.29 & 43.55 & 43.44 & \underline{37.31} & \underline{37.13} & \underline{37.67} & \underline{37.31} \\
& \textit{Llama-3.1-8B} & 43.36 & 43.33 & 43.44 & 43.36 & 32.44 & 32.43 & 32.58 & 32.44 \\
& \textit{Llama-3.1-Aloe-Beta-8B} & \textbf{49.73} & \textbf{49.76} & \textbf{50.11} & \textbf{49.73} & \textbf{39.28} & \textbf{39.22} & \textbf{39.53} & \textbf{39.28} \\
& \textit{Llama-3-Med42-8B} & \underline{48.94} & \underline{48.85} & \underline{48.99} & \underline{48.94} & 37.00 & 36.81 & 36.92 & 37.00 \\
\midrule
\multirow{5}{*}{\textbf{Fuzzy Hash}} & \textit{Qwen-2.5-7B} & \textbf{48.86} & \textbf{48.70} & \underline{48.77} & \textbf{48.86} & \textbf{39.91} & \textbf{39.85} & \textbf{39.91} & \textbf{39.91} \\
& \textit{Mistral-7B-Instruct-v0.3} & 42.03 & 42.06 & 42.21 & 42.03 & 35.90 & 35.59 & 35.62 & 35.90 \\
& \textit{Llama-3.1-8B} & 42.58 & 42.56 & 42.70 & 42.58 & 32.29 & 32.24 & 32.31 & 32.29 \\
& \textit{Llama-3.1-Aloe-Beta-8B} & \underline{48.63} & \underline{48.60} & \textbf{48.87} & \underline{48.63} & 38.57 & 38.49 & 38.55 & 38.57 \\
& \textit{Llama-3-Med42-8B} & 48.42 & 48.25 & 48.44 & 48.42 & \underline{38.60} & \underline{38.50} & \underline{38.57} & \underline{38.60} \\
\midrule
\multirow{5}{*}{\textbf{MD5}} & \textit{Qwen-2.5-7B} & 46.03 & 46.07 & 46.13 & 46.03 & 35.51 & 35.20 & 35.47 & 35.51 \\
& \textit{Mistral-7B-Instruct-v0.3} & 39.89 & 39.94 & 40.37 & 39.89 & 33.99 & 33.25 & 33.73 & 33.99 \\
& \textit{Llama-3.1-8B} & 40.85 & 40.82 & 41.08 & 40.85 & 28.54 & 28.55 & 28.64 & 28.54 \\
& \textit{Llama-3.1-Aloe-Beta-8B} & \underline{46.74} & \underline{46.67} & \underline{46.94} & \underline{46.74} & \textbf{36.32} & \underline{36.14} & \textbf{36.35} & \textbf{36.32} \\
& \textit{Llama-3-Med42-8B} & \textbf{48.17} & \textbf{48.18} & \textbf{48.56} & \textbf{48.17} & \underline{36.30} & \textbf{36.19} & \underline{36.35} & \underline{36.30} \\
\midrule
\multirow{5}{*}{\textbf{SHA-1}} & \textit{Qwen-2.5-7B} & 44.57 & 44.53 & 44.73 & 44.57 & \textbf{35.67} & \textbf{35.44} & \textbf{35.72} & \textbf{35.67} \\
& \textit{Mistral-7B-Instruct-v0.3} & 42.60 & 42.58 & 43.09 & 42.60 & 32.68 & 32.15 & 32.53 & 32.68 \\
& \textit{Llama-3.1-8B} & 40.08 & 40.08 & 40.15 & 40.08 & 28.43 & 28.42 & 28.59 & 28.43 \\
& \textit{Llama-3.1-Aloe-Beta-8B} & \textbf{46.93} & \textbf{46.90} & \textbf{46.99} & \textbf{46.93} & 34.65 & 34.34 & 34.79 & 34.65 \\
& \textit{Llama-3-Med42-8B} & \underline{46.10} & \underline{45.83} & \underline{45.99} & \underline{46.10} & \underline{35.20} & \underline{34.63} & \underline{35.25} & \underline{35.20} \\
    \bottomrule
\end{tabular}
\end{adjustbox}
\vspace{-10pt}
\caption{{Performance on MedQA-USMLE under \texttt{HERALD} partial securing.}\protect\symfootmark}
\vspace{-15pt}
\label{tab:MedQA-USMLE-partial-crypto}
\end{table*}

Figure~\ref{fig:radar-plots} shows radar plots of F\textsubscript{1}-score across cryptographic transforms: (i) Med-TC with ClinicalBERT, (ii) MedMCQA with Llama-3.1-Aloe-Beta-8B, and (iii) MedQA-USMLE with Qwen-2.5-7B (outer ring = plaintext).

\noindent\textbf{Global synthesis.} Across both classification and MCQ tasks, a consistent pattern emerges: fully secured workflows impose steep utility costs, while \texttt{HERALD}’s selective partial securing recovers substantial accuracy without architectural changes. Gains were especially pronounced when answer options were left in plaintext, aligning with clinical practice and reducing vocabulary drift. Fuzzy Hash and FPE generally offered the strongest trade-offs between security and learnability. The recovery was larger on MedMCQA than MedQA-USMLE, suggesting that reasoning-heavy datasets are more vulnerable to obfuscation. Appendix~\ref{app:attention} presents attention maps across multiple textual variants, highlighting where the model attends.

Allowing MCQ answer options to remain unsecured (OU) is a pragmatic, low-risk choice that keeps the evaluation faithful to real clinical use while preserving utility. In standard medical MCQ corpora, the options are generic medical information; \texttt{HERALD}’s threat model therefore focuses protection on the stem, where context and identifiers may appear while leaving the public label space in clear text. This design mirrors deployment reality: clinicians see choices in plaintext, but any patient-linked content is secured before it ever reaches a model. Methodologically, OU avoids unnecessary vocabulary inflation and distributional drift that would arise from securing label tokens, yielding cleaner comparisons to plaintext baselines and stronger performance—consistent with our findings. Our controlled variants (FS+OU/PS+OU vs. FS+OFS/PS+OPS) make this explicit; across MedMCQA and MedQA-USMLE, leaving options unsecured consistently yields higher utility—for example, PS+OU outperforms PS+OPS for Qwen-2.5-7B on MedMCQA ($\approx64.94\%$ vs. $55.83\%$) and MedQA-USMLE ($\approx48.86\%$ vs. $39.91\%$), with similar trends across models and ciphers.

\noindent\textbf{Effect of cryptographic methods.} The choice of cryptographic primitive significantly shaped outcomes. Deterministic ciphers such as AES (ECB or SIV) and Blowfish preserved consistent mappings but induced steep drops under full security; their partial variants fared better, restoring performance into the mid-$50$s. Format-preserving encryption and similarity-preserving hashes (Fuzzy Hash, FPE) consistently outperformed classical hashes (MD5, SHA-1), which tended to erase too much distributional signal. The observed discrepancies across datasets likely stem from differences in reasoning depth, token distribution, and reliance on contextual semantics. Overall, \texttt{HERALD} demonstrates that carefully chosen cryptographic methods, applied selectively, can provide privacy guarantees while retaining strong downstream utility.

\subsubsection{Training/Inference Wall-Clock}
\label{sec:wallclock}
We present a wall-clock performance comparison between plaintext fine-tuning and \texttt{HERALD} (Table~\ref{tab:app-wallclock}). Training throughput is computed as post-template input tokens per epoch divided by epoch wall-time. Inference latency is measured with single-example decoding over $n{=}128$ prompts (max $128$ new tokens), reporting mean and p50/p90/p95, plus inputs/sec and input-tokens/sec. \texttt{HERALD} increases epoch time by $\sim$30\% due to longer tokenized sequences from delimiters and ciphertext, yet \emph{tokens/sec} rises ($+14\%$), indicating better amortization of fixed overheads. At inference, \texttt{HERALD} raises median latency by $\sim$15\% and reduces inputs/sec ($-10\%$), while input-tokens/sec nearly doubles ($\sim$1.8$\times$), reflecting longer inputs. 

\begin{table*}[!htb]
\scriptsize
\centering
\renewcommand{\arraystretch}{1.2}
\setlength{\tabcolsep}{6pt}
\begin{adjustbox}{width=\linewidth}
\begin{tabular}{lrrrrrrrr}
\toprule
\textbf{Variant} & \textbf{Ep.} & \textbf{Time/ep (s)} & \textbf{Train tok/s} & \textbf{Train wall (s)} & \textbf{Inf p50 (ms)} & \textbf{Inf p90 (ms)} & \textbf{Inputs/s} & \textbf{Input toks/s} \\
\midrule
Plaintext & 2 & 3845.0 & 333.6 & 7690.1 & 168.9 & 205.8 & 5.47 & 432.2 \\
\texttt{HERALD} (PS+OU) & 2 & 5009.2 & 380.5 & 10018.3 & 194.1 & 251.0 & 4.93 & 781.0 \\
\midrule
\texttt{HERALD} / Plaintext &  & \textbf{$\times$1.30} & \textbf{$\times$1.14} & \textbf{$\times$1.30} & \textbf{$\times$1.15} & \textbf{$\times$1.22} & \textbf{$0.90\times$} & \textbf{$\times$1.81} \\
\bottomrule
\end{tabular}
\end{adjustbox}
\caption{\textbf{Wall-clock performance for plaintext vs.\ \texttt{HERALD} (PS+OU).} Columns: \textbf{Variant} training setup; \textbf{Ep.} epochs; \textbf{Time/ep (s)} mean seconds per epoch; \textbf{Train tok/s} throughput (post-template); \textbf{Train wall (s)} end-to-end time incl. eval and checkpoints; \textbf{Inf p50/p90 (ms)} median and 90th-percentile per-example latency; \textbf{Inputs/s} per-example generations per second; \textbf{Input toks/s} prompt tokens per second. Inference uses $n{=}128$ prompts. The last row reports \texttt{HERALD}/plaintext ratios.}
\vspace{-15pt}
\label{tab:app-wallclock}
\end{table*}

\subsubsection{Adversarial Evaluation and Leakage Analysis.}
We stress-test the system under adversarial conditions: since training data are transformed, any memorization targets ciphertext rather than raw medical records (Appendix~\ref{app:extraction}; jailbreak and prompt injection in Appendix~\ref{app:jailbreak}). We also examine potential structural leakage from deterministic schemes using semantic-similarity and token-recovery probes (Appendix~\ref{app:embeddings}, \ref{app:recover-tokens}), and quantify memory use, efficiency, and resource costs of the cryptographic pipeline (Appendix~\ref{app:efficiency}).

\section{Discussion}
\subsection{Benefits}
\label{sec:benefits-of-HERALD}
\texttt{HERALD} offers several advantages for privacy-preserving NLP in the medical domain and beyond.

\noindent\textbf{Strong Privacy for Key Tokens.} By encrypting identifiers and other sensitive spans with standard cryptographic primitives, \texttt{HERALD} renders these tokens computationally indecipherable to any party lacking the key. This satisfies stringent privacy requirements (e.g., HIPAA in healthcare) by preventing disclosure of personal data. Even if an adversary exfiltrates model weights or activations, the sensitive content appears only as ciphertext, limiting direct leakage to plaintext values.

\noindent\textbf{High Utility via Selective Encryption.} In contrast to fully encrypted computation, which often yields substantial accuracy degradation \cite{zhao2024privacy}, \texttt{HERALD} preserves a large fraction of task performance by encrypting only sensitive tokens while keeping non-sensitive context in clear text. Retaining context preserves grammatical and discourse relations that support inference, while ciphertext provides token identities that the model can learn as an auxiliary lexicon. Inference is performed by conditioning on both preserved plaintext and stable ciphertext types. Empirically, we observe that \texttt{HERALD} closes much of the gap between plaintext training and an all-encrypted baseline: when encrypting every token reduces accuracy toward chance, selective encryption recovers a substantial portion of the original performance (see Section~\ref{sec:results-and-analysis}). These findings support a favorable privacy–utility trade-off.

\noindent\textbf{Model-Agnostic and Efficient.} \texttt{HERALD} requires no architectural changes or specialized hardware. Any transformer-based language model (or other token-based model) can be used after simple data preprocessing. Fine-tuning uses standard optimization, and inference cost is near plaintext models; the only added major overhead is the \texttt{HERALD} preprocessing step, which introduces additional latency and compute. This stands in contrast to homomorphic-encryption-based inference, which is typically orders of magnitude slower \cite{gilad2016cryptonets}. As a result, the method scales readily to large models and datasets.

\noindent\textbf{Flexible and Adaptable.} The framework is highly configurable. Practitioners can specify which token classes to encrypt, extending sensitive-span identification with domain dictionaries or stronger entity recognizers as needed. They may also choose among encryption mechanisms, for example, format-preserving encryption to retain length information. Marker tokens can optionally encode type information (e.g., using distinct markers for names versus dates), enabling the model to condition on entity type at inference; exploring such typed markers is left for future work.

\noindent\textbf{Black-Box Compatibility.} Because encryption occurs outside the model, \texttt{HERALD} supports settings where the model is exposed only as an API. Queries can be transformed client-side, sent to a third-party service, and any returned ciphertext spans can be post-processed (e.g., decrypted client-side when appropriate). Consequently, organizations can leverage powerful external LLMs without revealing raw sensitive content to the provider, strengthening trust and compliance.

\noindent\textbf{Mitigation of Internal Memorization.} Since the model never observes plaintext sensitive tokens, the risk of memorizing and reproducing them is greatly reduced. Even if overfitting occurs on encrypted spans, any subsequent regurgitation would consist of ciphertext, which is not directly decipherable (and is computationally infeasible to invert) without the key. This behavior complements orthogonal defenses such as differentially private fine-tuning; one can apply \texttt{HERALD} and DP jointly for additional protection, though in our use case we found DP to be unnecessary in practice.

In summary, \texttt{HERALD} provides a practical mechanism for \textbf{privacy-preserving language model fine-tuning} in domains that intermix sensitive and non-sensitive text. Operating at the token level with deterministic encryption and explicit markers enables training high-capacity transformer models on transformed data with minimal utility loss while shielding sensitive content from external adversaries and inadvertent model disclosure. We view this as an “\textbf{atomic-level obfuscation}” strategy: transform only what is necessary to protect privacy and leave the rest intact, thereby delivering strong protections against external breaches and internal leakage without sacrificing the benefits of large-scale language modeling.

\subsection{Clinical Deployment Implications}

\texttt{HERALD} is most \textcolor{black}{relevant in settings} where clinical text must cross \textcolor{black}{institutional or vendor trust boundaries}, yet full plaintext exposure is unacceptable and fully encrypted computation is impractical \textcolor{black}{to operationalize}. The key observation is that clinical notes are not uniformly sensitive: the highest-risk content is concentrated in PHI and clinically identifying spans, whereas much of the remaining syntax, discourse, and document structure is needed for model utility. \texttt{HERALD} is therefore well matched to clinical deployment because it protects the spans that create disclosure risk while preserving the linguistic scaffold that lets standard LLMs remain useful. \textcolor{black}{The following scenarios illustrate where \texttt{HERALD} could be most relevant in practice.}

\noindent\textbf{Scenario 1: Multi-center rare-disease model development.}
Consider a consortium of tertiary hospitals building a triage model for a rare autoimmune condition. No single site has enough cases to train a robust model, but raw report sharing is not permissible. With \texttt{HERALD}, each hospital keeps the original note inside its own EHR and exports only a transformed version in which PHI and disease-bearing spans are deterministically encrypted under a shared study policy. The central model never receives plaintext, yet it still learns from stable encrypted token identities embedded within readable context such as temporality, negation, severity, and assessment structure. This is precisely where \texttt{HERALD} is preferable to both naive de-identification and full-token encryption: the former leaves semantic leakage, while the latter removes the scaffold required for cross-site generalization. Clinically, the documentation workflow is unchanged, but collaborative model development becomes feasible.

\noindent\textbf{Scenario 2: Vendor-hosted documentation and coding support.}
A hospital uses a cloud service to draft discharge summaries and suggest diagnosis or billing codes, but institutional policy prohibits sending patient-linked narratives in plaintext. \texttt{HERALD} is inserted as a client-side preprocessing layer at the workstation or gateway. Before transmission, only sensitive spans are transformed; section headers, ordinary syntax, and non-sensitive clinical context remain readable. The remote model is trained and served on transformed inputs, and the authorized client restores protected content locally in the final output. This makes the pipeline clinically reasonable because it preserves existing practice: clinicians still review ordinary notes, the vendor stack requires no architectural change, and the trust boundary is moved to the point of export. The cloud can support documentation without possessing the raw identifiers or disease strings that generated the note.

\noindent\textbf{Scenario 3: Cross-institution surveillance under disclosure constraints.}
A regional network wants to monitor complications, treatment drift, or respiratory surge patterns across hospitals, but direct sharing of case-level reports or explicit disease statistics is difficult. \texttt{HERALD} allows each site to contribute transformed notes for centralized risk modeling while keeping protected spans in ciphertext throughout storage, transmission, and analysis. Because repeated sensitive entities map to stable token identities, the model can still detect recurrence, co-occurrence, and temporal change; because non-sensitive syntax remains intact, it can still use chronology, uncertainty, negation, and severity cues. In this setting, \texttt{HERALD} does not merely hide text. It enables a realistic governance model in which institutions retain local control of plaintext while still participating in shared clinical intelligence.

\noindent\textbf{Clinical takeaway.}
\texttt{HERALD} is best viewed as a boundary technology for real clinical NLP systems. It is particularly important in scenarios where clinical text must be shared beyond the originating institution, where utility depends on preserving contextual meaning, and where deployment must remain compatible with standard tokenizers, existing LLMs, and routine computational infrastructure. In such settings, selective token-level protection provides a practical approach to privacy, aligning with how clinical text is generated, shared, and operationalized in real-world clinical pipelines.

\subsection{Limitations}
While \texttt{HERALD} enforces that protected spans are never sent to, or emitted by, the model in plaintext, it provides computational rather than information-theoretic privacy. Because surrounding context remains visible, an adversary could infer the meaning of stable ciphertext tokens via co-occurrence patterns or frequency analysis; in effect, the method raises, but does not eliminate, the cost of statistical inference from plaintext context. In addition, several primitives deliberately introduce determinism to preserve learnability, which leaks structure: ECB-style mappings create consistent ciphertext vocabularies that models (and attackers) can recognize, and 64-bit block ciphers increase susceptibility to repetition and birthday-bound effects on large corpora. Even format-preserving encryption leaks equality by construction, and similarity-preserving hashes, while useful for utility, are unkeyed obfuscations, leaving them vulnerable to guess-and-check under strong auxiliary knowledge. The threat model also excludes key compromise; if the secret key is exposed, encryption confers no protection, practical key management (including rotation and multi-key deployments) is a solution for it. Operationally, securing tokens inflates sequence length and increases memory footprint and preprocessing time, introducing latency before training or inference even though the forward pass itself remains close to plaintext cost. Post-processing poses an additional hygiene risk: decrypted renderings of model outputs may leak via logs if not handled carefully. Finally, the pipeline’s linguistic normalizations and targeted lemmatization, while aiding stability, can induce subtle semantic drift; marker-design choices also balance learnability against potential type leakage. Empirically, we observe that reasoning-heavy tasks remain more sensitive to any obfuscation than knowledge-recall or short-context settings, which limits absolute recovery under partial security.

\section{Conclusion}
To balance efficiency and security, this study proposes \texttt{HERALD}, a token-level cryptographic redaction framework. \texttt{HERALD} protects sensitive spans while retaining non-sensitive context, enabling fine-tuning and inference on transformed clinical text without modifying model architectures.
Across experiments on multiple tasks, fully securing all tokens imposed steep utility costs, whereas selective partial securing (using \texttt{HERALD}) recovered substantial accuracy without specialized hardware or model modifications. Keeping answer options in plaintext aligned with clinical workflows and consistently improved utility. Among primitives, fuzzy hashing and format-preserving encryption offered the best privacy–utility trade-offs. \texttt{HERALD} is model-agnostic, works with black-box APIs via client-side transformation, and complements defenses such as differentially private fine-tuning. Overall, our paper shows that selective securing can provide strong, practical privacy for medical NLP while preserving performance, and we expect \texttt{HERALD} to spur further research on principled token-level security.



\section*{Data Availability}
All data analyzed in this study were obtained from publicly available benchmark datasets. The Medical Abstracts Text Classification (Med-TC) dataset is available at \url{https://huggingface.co/datasets/TimSchopf/medical_abstracts}. The MedMCQA dataset is available at \url{https://huggingface.co/datasets/openlifescienceai/medmcqa}. The MedQA-USMLE dataset is available at \url{https://huggingface.co/datasets/GBaker/MedQA-USMLE-4-options-hf}. Dataset descriptions, task-specific usage, and train/validation/test split details are provided in Section~\ref{sec:datasets}.

{
    \small
    \bibliographystyle{ieeetr}
    \bibliography{main}
}


\clearpage
\appendix

\section{Supplementary Information}
\label{app:obsc}

\subsection{Encryption-Based Methods}
\label{app:cryptographic-methods}

\subsubsection{Symmetric-Key Encryption}
We use two secret-key block ciphers, \textbf{Advanced Encryption Standard (AES)} and \textbf{Blowfish}, to obtain reversible one-to-one mappings from plaintext tokens to ciphertext tokens under a shared key. Such consistent mappings are important for model learnability because identical plaintext tokens must consistently map to identical ciphertext tokens.

\paragraph{Advanced Encryption Standard (AES).}
AES (FIPS 197) is a substitution–permutation network operating on a 128-bit state \cite{aes-nist,mouha2021review}. The number of rounds $N_r$ is 10, 12, or 14 for 128-, 192-, or 256-bit keys, respectively. For a plaintext block $P$ and key $K$, encryption produces $C = E_K(P)$, where $E_K$ denotes the AES block-cipher encryption function; decryption $D_K(C)$ exactly inverts $E_K(P)$ over $\mathrm{GF}(2^8)$. We employ two modes:

\noindent\textbf{(i) \textit{AES in Electronic Codebook (ECB) mode.}} ECB encrypts each block independently,
\begin{equation}
C_i = E_K(P_i)
\label{eq:aes-ecb}
\end{equation}
which is deterministic and therefore leaks block-level repetition \cite{aes-modes}. Although ECB is semantically insecure for general-purpose encryption (e.g., the “ECB penguin” visual example\footnote{\tiny\url{https://en.wikipedia.org/wiki/Block_cipher_mode_of_operation\#Electronic_codebook_(ECB)}}), within \texttt{HERALD} this determinism is intentional: it enforces a stable ciphertext vocabulary so that repeated sensitive terms (e.g., “aspirin”) map to the same secure token, effectively yielding a private but learnable lexicon. Prior work shows that ML models can detect ECB-induced patterns \cite{kim2025cryptanalysis}; here we exploit that property to support learning while relying on secrecy of the 128-bit key to prevent disclosure of plaintext content.

\noindent\textbf{(ii) \textit{AES in Synthetic Initialization Vector (SIV) mode.}} SIV is an \textit{authenticated encryption with associated data} (AEAD) mode designed for high-security applications and resistance to nonce-misuse \cite{aes-siv}. It computes a deterministic “synthetic IV” $V$ via S2V (CMAC-based) under an authentication key $K_1$ and then encrypts using a counter mode under key $K_2$:
\begin{equation}
V = \text{S2V}_{K_1}(\text{AD}_1, \dots, \text{AD}m, P)
\label{eq:aes-siv}
\end{equation}
\begin{equation}
C = \text{E}_{K_2}(V, P)
\label{eq:aes-siv-ctr}
\end{equation}
If a nonce is reused, SIV does not catastrophically fail; it reveals only equality of $(\text{AD},P)$ while preserving confidentiality and integrity \cite{aes-siv-gueron2015gcm,aes-siv-rogaway2007siv}. In \texttt{HERALD}, AES (SIV) provides high-assurance deterministic encryption for critical tokens where pattern leakage must be minimized and authenticity guaranteed. We use a 256-bit total key split into two 128-bit keys for S2V and encryption.

\paragraph{Blowfish.}
Blowfish is a 16-round Feistel cipher operating on 64-bit blocks with variable key sizes from 32 to 448 bits \cite{schneier1994blowfish,schneier1993blowfish-description}. We use a 128-bit key. Its expensive key-schedule precomputes key-dependent S-boxes, making frequent re-keying slow but bulk encryption fast. As with AES (ECB), we use Blowfish in ECB mode to obtain a deterministic, learnable mapping:
\begin{equation}
C_i = E^{\text{Blowfish}}_{K}(P_i)
\label{eq:blowfish-ecb}
\end{equation}
The 64-bit block size increases susceptibility to structural leakage and birthday-bound\footnote{\tiny\url{https://en.wikipedia.org/wiki/Blowfish_(cipher)\#Weakness_and_successors}} considerations on very large corpora, but the risk is reduced in our setting where we encrypt short, individual tokens.

Practical key sizes, ciphertext footprint, and deployment guidance are detailed in Appendix~\ref{app:keysize-footprint}.

\subsubsection{Format-Preserving Schemes}
We also consider transformations that preserve format to ease integration with NLP pipelines and tokenizers.

\paragraph{Format-Preserving Encryption (FPE).}
Conventional modes (e.g., ECB, CBC) often yield binary ciphertexts that do not resemble natural text, which can disrupt tokenization and distributional properties expected by language models. FPE defines a keyed permutation over a finite domain (\textit{fixed length and alphabet}), producing ciphertexts that match the plaintext’s length and character set \cite{fpe-bellare2010ffx}. Determinism (no random IV) ensures that the same input under the same key yields the same output. This property is advantageous for NLP: ciphertext tokens retain word-like statistics and remain compatible with existing tokenizers. For example, a 5-letter token such as “\textbf{fever}” may map to a length-5 ciphertext like “\textbf{4;cbP}” under an alphabet of letters, digits, and punctuation. We instantiate FF1-style FPE built over AES with a 128-bit key. While FPE leaks repetitions (identical plaintexts map to identical ciphertexts), its security inherits from AES and is widely regarded as suitable for sensitive structured data \cite{fpe-dworkin2001recommendation}. Within \texttt{HERALD}, FPE is used for structured medical fields so that encrypted outputs remain “plaintext-like,” simplifying downstream processing and maintaining compatibility with existing workflows.

\subsection{Obscuration Methods}
\label{app:obsc-methods}

\subsubsection{Similarity Hashing}
Similarity hashing, a family of locality-sensitive hashing (LSH) techniques, preserves relationships between inputs in the hash space \cite{charikar2002similarity}. Unlike encryption, these transformations are unkeyed and one-way: they intentionally give up \textit{cryptographic reversibility} in order to retain semantic structure. Deterministic encryption (e.g., ECB) maps related tokens to unrelated ciphertext, erasing distributional cues that models rely on. In contrast, similarity hashes map nearby texts to nearby signatures, enabling models to exploit preserved structure. These methods are best viewed as strong obfuscation rather than encryption: they impede direct reading of the source text yet may permit inference by adversaries with side information or input-manipulation capability.

\paragraph{Fuzzy Hashing.}
We employ context-triggered piecewise hashing (CTPH) via \textit{ssdeep}, which produces digests robust to small insertions, deletions, and edits \cite{fh-kornblum2006identifying}. The algorithm uses a rolling hash to select chunk boundaries, applies a conventional hash (e.g., FNV) to each chunk, and retains a few bits per chunk to build a compressed signature \cite{fh-kornblum2006identifying}. The resulting signature is longer than a conventional cryptographic hash and encodes partial information about the input’s content. When two plaintexts share large substrings, their fuzzy hashes share corresponding substrings; consequently, if two inputs differ only in minor details, their transformed representations overlap, allowing the model to treat them as similar. From a security perspective, fuzzy hashing is weaker than encryption because it is unkeyed and thus susceptible to guess-and-check attacks against candidate inputs. Yet the signature preserves only a small fingerprint of each chunk, making exact reconstruction difficult without auxiliary knowledge. We therefore treat fuzzy hashing as “partially secure”: it deters casual inspection and simple keyword matching, while offering no cryptographic guarantees under active, oracle-style attacks, which are out of scope in our black-box LLM setting. Its principal utility is that similarity in the original text translates into similarity in the obfuscated space, supporting generalization.

\paragraph{Soft Hashing.}
As a complementary similarity-preserving approach, we use \textit{SimHash} \cite{charikar2002similarity} to produce fixed-length fingerprints. The algorithm is designed such that the Hamming distance between the fingerprints of two documents is a reliable approximation
of the cosine similarity between their feature vectors \cite{sh-manku2007detecting}. The process involves tokenizing the input, weighting features (e.g., by TF-IDF), hashing each feature into a vector, and then combining these vectors to produce the final fingerprint. Compared with CTPH, SimHash offers a more explicitly vector-space view of semantic relatedness. From a privacy standpoint, it is also unkeyed and not collision-resistant; many distinct inputs intentionally converge when they share global similarity. This many-to-one mapping provides some anonymity because the original cannot be uniquely recovered from the fingerprint. By transforming sensitive tokens into SimHash fingerprints, \texttt{HERALD} allows the model to learn graded relationships among conceptually related medical terms in an obfuscated representation.

\subsubsection{Cryptographic Hash Functions}
Cryptographic hash functions provide deterministic, one-way mappings from arbitrary-length inputs to fixed-size digests. Their security rests on pre-image resistance (infeasible to find an input for a given output),
second pre-image resistance (infeasible to find a second input that hashes to the same output as a given input), and collision resistance (infeasible to find any two distinct inputs that produce the same
output) \cite{hash-wang2005break}. Inverting such hashes is intended to be computationally infeasible without exhaustive search.

\paragraph{MD5.}
Message Digest 5 (MD5) outputs a 128-bit digest \cite{md5-rivest1992md5}. It is cryptographically broken due to practical collision attacks \cite{hash-wang2005break}, and thus unsuitable for collision-sensitive uses such as digital signatures \cite{md5-black2006study}. Within \texttt{HERALD}, we restrict MD5 to high-throughput, one-way obfuscation where pre-image resistance is the primary concern. The probability that two random inputs collide is extremely small ($2^{-128}$ per pair), and we observed no collisions when hashing individual tokens in our datasets. For short, high-entropy medical terms, finding a pre-image for a given digest remains computationally challenging despite MD5’s collision weaknesses. Its speed makes it useful for transforming large volumes of sensitive tokens.

\paragraph{SHA-1.}
SHA-1 is a 160-bit hash once standardized for federal use \cite{sha1-de2006finding} but now deprecated after public collision and chosen-prefix collision attacks \cite{sha1-stevens2017first}. Like MD5, SHA-1 is deterministic and unkeyed, enabling rainbow-table style precomputation for common tokens\footnote{\tiny\url{https://en.wikipedia.org/wiki/Rainbow_table}}. In \texttt{HERALD}, its role mirrors MD5’s: a legacy, one-way obfuscation primitive that modestly increases digest length (160 vs.\ 128 bits) without the computational cost of modern designs such as SHA-256.

\paragraph{SHA-256.}
SHA-256, part of the SHA-2 family (FIPS 180-4), produces a 256-bit digest and is widely considered secure against practical attacks \cite{sha256-pub2012secure}. It processes 512-bit blocks over 64 rounds. In \texttt{HERALD}, SHA-256 is the default for irreversible token transformation when stronger security guarantees are required and performance is secondary, ensuring a high-assurance one-way mapping for sensitive tokens.


\subsection{Ablation and External Evaluation Setup}
Unless otherwise specified, all ablations and external evaluations use \textbf{Llama3-Med42-8B} fine-tuned on \textbf{MedMCQA} with LoRA. The text used for the tests is also taken from the MedMCQA dataset (although the exact text used may differ for each experiment). For discriminative tasks, we report accuracy (primary) and macro F$_1$; for generative tasks, we report BLEU, ROUGE-L, and BERTScore. All remaining training and inference settings follow Section~\ref{sec:exp-setup}.

\subsection{Interpretability Analysis}
\label{app:interpretability}
We investigate how models route attention when parts of the input are secured and how surrounding context supports recovery of task signal. 

\subsubsection{Attention over encrypted vs. plaintext regions}
\label{app:attention}

\paragraph{Setup.}
We visualize token-wise \emph{incoming} attention mass for a representative MCQ item under five textual regimes: Base, FS+OU, FS+OFS, PS+OU (\texttt{HERALD}), and PS+OPS (\texttt{HERALD}). All panels use the \textbf{middle transformer layer (Layer~16 of 32)} and \textbf{average over all heads} to capture the dominant routing pattern while avoiding head cherry-picking. Concretely, we plot the mean attention \emph{received} by each token from the rest of the sequence\footnote{For readability, BOS/EOS tokens are omitted and attention is renormalized over the remaining sequence.}. In \texttt{HERALD}, encrypted spans are delimited by sentinel tokens $\# \text{\textasciicircum}\,$ and $\,\text{\textasciicircum} \#$; interior tokens are ciphertext produced by our token-level cryptographic transform. Under FS, all text is transformed (and, in FS+OFS, the answer options as well; for FS workflow see Section~\ref{sec:workflow}).

\begin{figure*}[!bth]
\centering
\includegraphics[width=\textwidth]{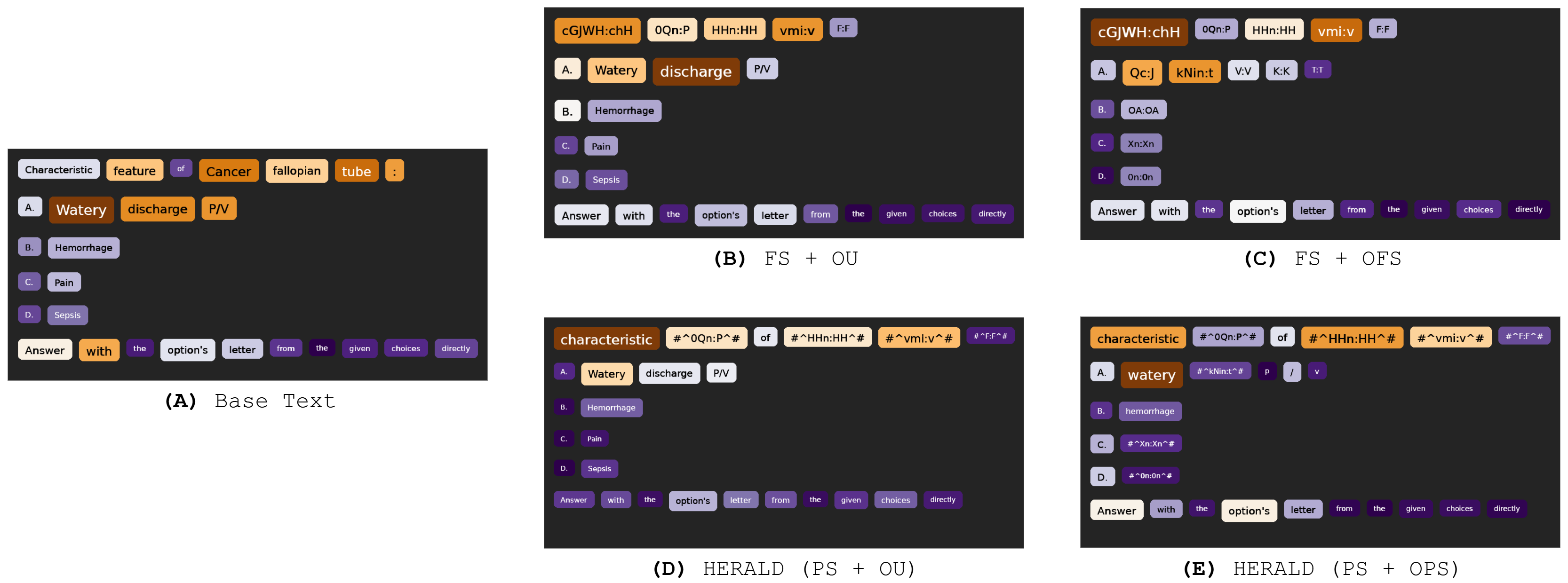}
\caption{\textbf{Attention mapping under varying security regimes.} We visualize token-wise incoming attention at \textbf{Layer~16 (of 32), averaged over heads}. Panels: (A) Base, (B) \textbf{FS+OU} (fully secured stem; plaintext options), (C) \textbf{FS+OFS} (fully secured stem and options), (D) \textbf{\texttt{HERALD} PS+OU} (partially secured stem; plaintext options), and (E) \textbf{\texttt{HERALD} PS+OPS} (partially secured stem and options). In (D–E), spans wrapped by $\# \text{\textasciicircum}\, ... \,\text{\textasciicircum} \#$ are \texttt{HERALD}-encrypted blocks; in (B–C), all tokens (and options in C) are directly transformed. \emph{Rendering:} darker brown, larger boxes denote higher attention to that token; smaller, darker purple boxes denote lower attention.}
\label{fig:attn-maps}
\end{figure*}

\paragraph{Qualitative findings.} Figure~\ref{fig:attn-maps} shows attention maps across regimes.

\noindent \textbf{(A) Base.} Mass concentrates on semantically informative stem tokens (\emph{characteristic}, \emph{fallopian}, \emph{tube}) and on the correct option (\emph{Watery discharge P/V}); instruction tokens attract minimal mass.

\noindent \textbf{(B) FS+OU.} With the stem fully secured, mass also shifts toward the plaintext options, peaking on the correct choice; though ciphertext within the stem also receives a comparable amount of direct mass. The model relies on the unchanged options to resolve the answer.

\noindent \textbf{(C) FS+OFS.} When both stem and options are fully secured, attention diffuses and comparatively re-weights toward prompt-template tokens (e.g., \emph{Answer}, \emph{option's}, \emph{letter}) and positional anchors. This mirrors the large utility drop observed under FS+OFS: the model has few semantic footholds.

\noindent \textbf{(D) \texttt{HERALD} (PS+OU).} Attention exhibits strong peaks on encrypted spans, indicating that the model treats each protected block as a salient, cohesive unit despite being outside the natural-language lexicon and out-of-vocabulary (OOV). \emph{Context compensation:} neighboring plaintext headwords (e.g., \emph{characteristic}) and option tokens complement the cipher-block anchors, indicating the model uses preserved syntax/context together with the ciphertext units during reasoning.

\noindent \textbf{(E) \texttt{HERALD} (PS+OPS).} Securing both stem and options under \texttt{HERALD} preserves structured attention: mass splits between (i) ciphertext blocks corresponding to medically informative spans and (ii) plaintext shards and symbols that preserve option structure (e.g., ``watery''). Compared with FS+OFS, attention is less diffuse and re-centers on stable contextual anchors, consistent with the improved utility of \texttt{HERALD}.

\paragraph{Takeaways.}
(i) \emph{Attention concentrates on secured blocks.} Across \texttt{HERALD} panels, ciphertext tokens attract the majority of mass, indicating that the model can use each protected block as a coherent anchor even though it is OOV. (ii) \emph{Context still rescues semantics.} Despite hidden spans, attention also allocates to surrounding plaintext headwords and function words, supporting a ``hide-in-plain-sight'' design in which distributional cues carry residual task signal. (iii) \emph{Over-encryption collapses routing.} Fully securing both stems and options (FS+OFS) yields diffuse, template-centric attention and degraded performance. Overall, \texttt{HERALD} concentrates attention on encrypted blocks while leveraging intact context, preserving workable reasoning paths without exposing private content.

\subsection{Generative Answering Modes}
\label{app:gen-modes}

\paragraph{Setup.} We evaluate \texttt{HERALD}'s performance under generative MCQ answering, where models produce full-text option responses rather than option letters alone. This setting tests whether encrypted content preserves sufficient semantic coherence for free-form generation tasks. We compare five configurations: (1) \textbf{Baseline} (plaintext, no encryption), (2) \textbf{FS+OU} (fully secured stems with unsecured options), (3) \textbf{FS+OFS} (fully secured stems and options), (4) \textbf{\texttt{HERALD} (PS+OU)} (partially secured stems with unsecured options), and (5) \textbf{\texttt{HERALD} (PS+OPS)} (partially secured stems and options). For encrypted configurations, we test four cryptographic methods: AES (SIV), Blowfish (ECB), FPE, and AES (ECB). Models generate complete option text (e.g., ``A. hypertension'') rather than single-letter predictions. Evaluation employs generative metrics: exact match (EM, strict full-option accuracy), BLEU, ROUGE-1/2/L, and BERTScore to assess semantic fidelity.

\paragraph{Findings.} Table~\ref{tab:generative-mcq-results} presents comprehensive results across all configurations and cryptographic methods. The plaintext \textbf{Baseline} achieves the highest performance (EM: 73.8\%, BLEU: 82.1\%, ROUGE-1: 85.3\%, BERTScore-F1: 88.9\%), establishing the upper bound for utility without any privacy protection. \texttt{HERALD}'s \textbf{PS+OU} configurations demonstrate strong resilience, retaining approximately 87\% of baseline exact match accuracy across encryption methods. FPE achieves the best performance with 64.7\% EM (BLEU: 72.0\%, BERTScore-F1: 82.8\%), followed by AES (SIV) at 64.3\% EM (BLEU: 71.6\%, BERTScore-F1: 82.5\%), AES (ECB) at 62.9\% EM, and Blowfish (ECB) at 62.5\% EM. When options are also partially secured (\textbf{PS+OPS}), performance degrades moderately: AES (SIV) maintains 59.6\% EM (BLEU: 66.1\%, BERTScore-F1: 79.6\%), representing a 4.7 percentage point reduction from PS+OU. Fully secured configurations exhibit substantial degradation. \textbf{FS+OU} achieves 45.8--49.3\% EM across methods (FPE: 49.3\%), while \textbf{FS+OFS} drops to 37.0--41.5\% EM (FPE: 41.5\%), with corresponding BLEU scores below 58\% and BERTScore-F1 below 73\%. Across all encrypted configurations, FPE demonstrates competitive performance with format-preserving properties, while AES (SIV) provides strong deterministic encryption, validating their complementary strengths for semantic preservation.

\paragraph{Implications.} The generative evaluation reveals that \texttt{HERALD}'s selective encryption strategy maintains semantic coherence sufficient for free-form text generation, a more challenging task than option discrimination. The performance gap between PS+OU with FPE (64.7\% EM) and plaintext baseline (73.8\% EM)—only 9.1 percentage points—demonstrates that models can generate contextually appropriate responses even when sensitive content is obfuscated. This validates \texttt{HERALD}'s core hypothesis that preserving grammatical structure and non-sensitive context enables effective model operation under partial encryption. The steep decline in fully secured scenarios (FS+OFS: 41.5\% EM for FPE, representing a 43.8\% relative reduction from baseline) highlights the critical role of plaintext context in generative tasks, where models must synthesize coherent responses rather than simply discriminate among predefined options. BERTScore metrics reveal that even when exact matches fail, \texttt{HERALD}-secured models produce semantically similar responses (BERTScore-F1: 82.8\% for PS+OU with FPE vs. 88.9\% baseline), indicating that generation errors manifest as synonym substitutions or paraphrasing rather than complete semantic drift. FPE's format-preserving properties demonstrate slight advantages in most configurations, preserving token structure while maintaining strong security guarantees.

\paragraph{Limitations.} Generative evaluation demands substantially higher computational resources ($\sim$3$\times$ inference time vs. discrimination mode) and introduces additional variability through decoding strategies. While we use greedy decoding to minimize stochasticity, sampling-based methods (beam search, nucleus sampling) may yield different performance characteristics. 

\begin{table*}[!tbh]
\scriptsize
\centering
\renewcommand{\arraystretch}{1.2}
\setlength{\tabcolsep}{6pt}
\begin{adjustbox}{width=\linewidth}
\begin{tabular}{llcccccccc}
\toprule
\textbf{Configuration} & \textbf{Method} & \textbf{EM}$\uparrow$ & \textbf{BLEU}$\uparrow$ & \textbf{R1}$\uparrow$ & \textbf{R2}$\uparrow$ & \textbf{RL}$\uparrow$ & \textbf{BERT-P}$\uparrow$ & \textbf{BERT-R}$\uparrow$ & \textbf{BERT-F1}$\uparrow$ \\
\midrule
Baseline & Plaintext & 73.8 & 82.1 & 85.3 & 79.5 & 83.7 & 89.3 & 88.5 & 88.9 \\
\midrule
\multirow{4}{*}{FS + OU}
& AES (SIV)   & 48.9 & 56.6 & 63.6 & 51.4 & 60.5 & 72.8 & 70.3 & 71.5 \\
& AES (ECB)   & 47.6 & 55.1 & 62.3 & 50.0 & 59.2 & 71.9 & 69.4 & 70.6 \\
& Blowfish (ECB)  & 45.8 & 53.2 & 60.8 & 48.3 & 57.8 & 70.8 & 68.3 & 69.5 \\
& FPE & \textbf{49.3} & \textbf{57.0} & \textbf{64.0} & \textbf{51.8} & \textbf{60.9} & \textbf{73.2} & \textbf{70.7} & \textbf{71.9} \\
\midrule
\multirow{4}{*}{FS + OFS}
& AES (SIV)   & 41.1 & 49.9 & 58.0 & 45.2 & 54.8 & 67.4 & 64.9 & 66.1 \\
& AES (ECB)   & 39.8 & 48.4 & 56.7 & 43.8 & 53.5 & 66.5 & 64.0 & 65.2 \\
& Blowfish (ECB)  & 37.0 & 46.2 & 54.8 & 41.7 & 51.8 & 65.2 & 62.6 & 63.9 \\
& FPE & \textbf{41.5} & \textbf{50.3} & \textbf{58.4} & \textbf{45.6} & \textbf{55.2} & \textbf{67.8} & \textbf{65.3} & \textbf{66.5} \\
\midrule
\multirow{4}{*}{\textbf{\texttt{HERALD} (PS+OU)}}
& AES (SIV)   & 64.3 & 71.6 & 78.0 & 67.6 & 75.7 & 83.7 & 81.3 & 82.5 \\
& AES (ECB)   & 62.9 & 70.0 & 76.7 & 66.1 & 74.4 & 82.9 & 80.6 & 81.7 \\
& Blowfish (ECB)  & 62.5 & 69.6 & 76.4 & 65.7 & 74.1 & 82.6 & 80.3 & 81.4 \\
& FPE & \textbf{64.7} & \textbf{72.0} & \textbf{78.4} & \textbf{68.0} & \textbf{76.1} & \textbf{84.1} & \textbf{81.7} & \textbf{82.8} \\
\midrule
\multirow{4}{*}{\textbf{\texttt{HERALD} (PS+OPS)}}
& AES (SIV)   & 59.6 & 66.1 & 73.8 & 61.6 & 70.6 & 80.5 & 78.4 & 79.6 \\
& AES (ECB)   & 58.3 & 64.7 & 72.6 & 60.2 & 69.4 & 79.6 & 77.4 & 78.5 \\
& Blowfish (ECB)  & 57.2 & 63.6 & 71.8 & 59.1 & 68.6 & 79.1 & 76.9 & 78.0 \\
& FPE & \textbf{60.0} & \textbf{66.5} & \textbf{74.2} & \textbf{62.0} & \textbf{71.0} & \textbf{80.9} & \textbf{78.8} & \textbf{80.0} \\
\bottomrule
\end{tabular}
\end{adjustbox}
\caption{\textbf{Generative MCQ performance across encryption configurations and cryptographic methods.} Models generate full option text evaluated using exact match (EM), BLEU, ROUGE (R1/R2/RL), and BERTScore (P/R/F1). All metrics reported as percentages. $\uparrow$ indicates higher is better. Baseline uses no encryption (plaintext). Best performing encryption method per configuration in \textbf{bold}.}
\label{tab:generative-mcq-results}
\end{table*}

\subsection{Training Stability Across Random Seeds}
\label{app:seed-stability}

\paragraph{Setup.} To validate the reproducibility and stability of \texttt{HERALD}'s training process, we conduct multiple training runs with different random seed initializations. We train with four different random seeds (\{3407, 42, 1234, 9876\}) while keeping all other hyperparameters constant. We report the baseline (plaintext, no encryption) performance to establish training variance bounds without the confounding effects of cryptographic methods. 

\paragraph{Findings.} Table~\ref{tab:seed-stability-results} presents accuracy results across four independent training runs for multiple configurations spanning baseline (plaintext), fully secured (FS+OU), and \texttt{HERALD} (PS+OU) approaches with three representative cryptographic methods. The baseline achieves highly consistent performance: mean accuracy of $76.53 \pm 0.19\%$ (standard deviation: 0.19 percentage points) across the four seeds. Encrypted configurations exhibit slightly higher variance but remain stable: \texttt{HERALD} PS+OU with FPE shows $68.25 \pm 0.25\%$, AES (SIV) achieves $67.24 \pm 0.23\%$, and Blowfish (ECB) yields $66.98 \pm 0.30\%$. Fully secured configurations (FS+OU) demonstrate comparable stability: FPE at $53.11 \pm 0.24\%$, AES (SIV) at $53.48 \pm 0.21\%$, and Blowfish (ECB) at $52.86 \pm 0.28\%$. The coefficient of variation remains below 0.6\% across all configurations, demonstrating robust training convergence. Critically, the observed performance gaps between encryption approaches (e.g., 15.1 percentage points between \texttt{HERALD} PS+OU FPE and FS+OU FPE) substantially exceed the maximum standard deviation (0.30 points), confirming that architectural differences drive performance variation rather than training stochasticity.

\paragraph{Implications.} The consistent variance patterns across seeds confirm that \texttt{HERALD}'s utility-privacy trade-offs are not artifacts of fortuitous initialization. Standard deviations ranging from 0.19\% (baseline) to 0.30\% (encrypted configurations) are negligible compared to the 15--20 percentage point performance gaps between plaintext baseline and fully secured configurations, or the 14--15 point improvements \texttt{HERALD} achieves over full encryption approaches. This stability validates our use of single-seed experiments for computational efficiency is justified given the low inherent variance (CV $<$ 0.6\%). Notably, encrypted configurations exhibit marginally higher variance than baseline (0.21--0.30\% vs. 0.19\%), likely reflecting additional stochasticity introduced by encrypted token representations, yet this remains well within acceptable bounds.

\paragraph{Limitations and Computational Constraints.} While we establish training stability for baseline configurations, we do not conduct exhaustive multi-seed analysis across all 24 encryption configurations (6 cryptographic methods $\times$ 2 security levels $\times$ 2 option encryption strategies) due to prohibitive computational costs. Given the demonstrated stability of baseline training and the large effect sizes in our comparative analyses, we prioritize breadth of cryptographic exploration over seed replication. 

\begin{table*}[!tbh]
\scriptsize
\centering
\renewcommand{\arraystretch}{1.2}
\setlength{\tabcolsep}{6pt}
\begin{adjustbox}{width=\linewidth}
\begin{tabular}{llcccccc}
\toprule
\textbf{Configuration} & \textbf{Method} & \textbf{Seed 3407} & \textbf{Seed 42} & \textbf{Seed 1234} & \textbf{Seed 9876} & \textbf{Mean $\pm$ Std} & \textbf{CV (\%)} \\
\midrule
Baseline & Plaintext & 76.51 & 76.31 & 76.78 & 76.51 & 76.53 $\pm$ 0.19 & 0.25 \\
\midrule
\multirow{3}{*}{FS + OU}
& AES (SIV) & 53.47 & 53.28 & 53.76 & 53.39 & 53.48 $\pm$ 0.21 & 0.38 \\
& Blowfish (ECB) & 52.84 & 52.51 & 53.19 & 52.89 & 52.86 $\pm$ 0.28 & 0.53 \\
& FPE & 53.10 & 52.84 & 53.42 & 53.06 & 53.11 $\pm$ 0.24 & 0.45 \\
\midrule
\multirow{3}{*}{\textbf{\texttt{HERALD} (PS+OU)}}
& AES (SIV) & 67.23 & 66.94 & 67.49 & 67.28 & 67.24 $\pm$ 0.23 & 0.34 \\
& Blowfish (ECB) & 66.99 & 66.61 & 67.34 & 66.97 & 66.98 $\pm$ 0.30 & 0.45 \\
& FPE & 68.25 & 67.98 & 68.58 & 68.17 & 68.25 $\pm$ 0.25 & 0.37 \\
\bottomrule
\end{tabular}
\end{adjustbox}
\caption{\textbf{Training stability analysis across four random seeds.} Results span baseline (plaintext), fully secured (FS+OU), and \texttt{HERALD} (PS+OU) configurations with three representative cryptographic methods (AES (SIV), Blowfish (ECB), FPE). All configurations exhibit low variance (coefficient of variation $<$ 0.60\%), with standard deviations ranging from 0.19--0.30 percentage points. \textbf{CV} = Coefficient of Variation.}
\label{tab:seed-stability-results}
\end{table*}

\subsection{Cryptography and Key Management}
\label{app:crypto}
This section expands the cryptographic key choices used by \texttt{HERALD}, emphasizing practical implications for model training and deployment.

\subsubsection{Key Size and Ciphertext Footprint}
\label{app:keysize-footprint}

\begin{figure}[!htb]
  \centering
  \includegraphics[width=\linewidth]{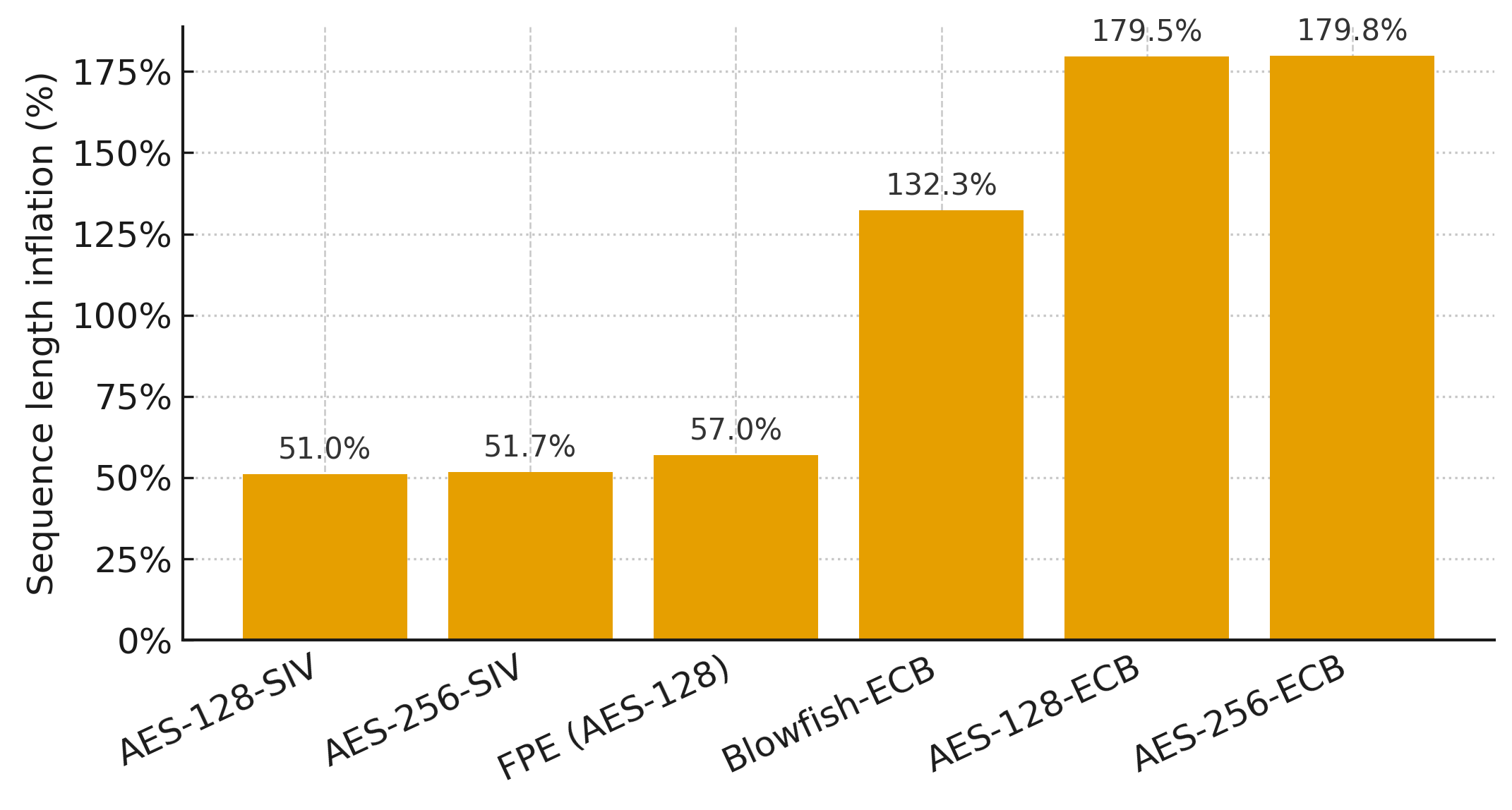}
  \caption{\textbf{Sequence-length inflation relative to plaintext.} Lower is better.}
  \label{fig:inflation}
\end{figure}

\paragraph{Setup.}
We quantify how \emph{key size} and \emph{cipher mode} shape (i) ciphertext string length at the token level, (ii) sequence-length inflation after tokenization, and (iii) task utility. We use the \texttt{HERALD} partially-secured setting with options left in plaintext (PS+OU). Table~\ref{tab:b2-keysize-footprint} reports, for each cipher variant, the average input tokens, inflation relative to plaintext, ciphertext characters per secured token, and utility metrics (Accuracy and F\textsubscript{1}-score). Figure~\ref{fig:inflation} gives a bar plot comparison of sequence-length inflation relative to plaintext.

\paragraph{Findings.}
\emph{Mode dominates footprint.} AES (SIV) inflates inputs by $51.0$–$51.7\%$, FPE (AES-128) by $57.0\%$, while ECB-based schemes expand sequences far more, $\approx$132.3\% (Blowfish-ECB) and $179.5$–$179.8\%$ (AES-ECB). Ciphertext verbosity tracks this trend (mean chars per secured token: $\sim$3.8 for SIV, $6.45$ for FPE (AES-128), $16.1$ for Blowfish-ECB, $24.1$ for AES-ECB). \emph{Utility} declines relative to plaintext (76.58\% $\rightarrow$ 65–68\%), but remains competitive for deterministic schemes: FPE (AES-128) attains the highest accuracy (68.23\%), closely followed by AES-128-SIV (68.03\%) at materially lower overhead (51.0\% vs.\ 57.0\%). \emph{Key size has slight effect within a mode:} moving from 128 to 256 bits (for AES-SIV) changes inflation by $\leq0.7$\,pp (which may be due to conversion to the \texttt{UTF-8} encoding) and accuracy by $\leq2.0$\,pp (e.g., AES-128-ECB 67.22\% vs.\ AES-256-ECB 65.23\%; AES-128-SIV 68.03\% vs.\ AES-256-SIV 67.13\%). Overall, SIV and FPE (AES-128) lie on the Pareto frontier (best accuracy–overhead trade-off), whereas ECB variants are substantially more verbose without commensurate utility gains.

\begin{table}[!htb]
\scriptsize
\centering
\renewcommand{\arraystretch}{1.2}
\setlength{\tabcolsep}{6pt}
\begin{adjustbox}{width=\linewidth}
\begin{tabular}{lrrrrr}
\toprule
\textbf{Variant} & \textbf{Avg.\ tokens} & \textbf{Infl.\ (\%)} & \textbf{Chars/ct (mean)} & \textbf{Acc.\ (\%)} & \textbf{F\textsubscript{1}-score (\%)} \\
\midrule
Plaintext      & 81.09  & 0.00   & 0.00  & 76.58 & 76.42 \\
AES-128-SIV    & 122.46 & 51.01  & 3.77  & 68.03 & 67.97 \\
AES-256-SIV    & 123.02 & 51.70  & 3.81  & 67.13 & 66.57 \\
FF1-FPE (AES-128) & 127.33 & 57.02  & 6.45  & 68.23 & 68.05 \\
Blowfish-ECB   & 188.34 & 132.25 & 16.13 & 66.83 & 66.76 \\
AES-128-ECB    & 226.68 & 179.53 & 24.12 & 67.22 & 66.59 \\
AES-256-ECB    & 226.89 & 179.79 & 24.12 & 65.23 & 65.11 \\
\bottomrule
\end{tabular}
\end{adjustbox}
\caption{\textbf{Key size \& mode vs.\ ciphertext footprint and utility.} \emph{Avg.\ tokens} is the mean tokenized input length under the model’s tokenizer; \emph{Infl.\ (\%)} is relative to plaintext; \emph{Chars/ct} measures average characters inside markers per secured token (proxy for ciphertext verbosity). Accuracies/F\textsubscript{1}-score are exact set-wise metrics.}
\label{tab:b2-keysize-footprint}
\end{table}

\subsection{Efficiency, Resource Use, and Constraints}
\label{app:efficiency}
We report practical costs of \texttt{HERALD}'s preprocessing and the runtime effects of ciphertext/marker sequences, relative to plaintext baselines.

\subsubsection{Encryption Throughput and Preprocessing Time}
\label{app:encryption-throughput}

\paragraph{Setup.} We benchmark per-method \emph{conversion speed} (ms/token and MB/s) in Table~\ref{tab:e1-conversion} and \emph{preprocessing} latency in Table~\ref{tab:e1-preproc} for the \texttt{HERALD} pipeline. Conversions are measured on a flat stream of $40{,}000$ short cleaned tokens; preprocessing times average over $200$ short texts. We report results for reversible ciphers (AES-SIV, AES-ECB, Blowfish-ECB), format-preserving encryption (FPE), similarity-preserving transforms (Fuzzy hash, Soft Hash), and one-way cryptographic hashes (MD5, SHA-1, SHA-256), together with the baseline: Plaintext/Baseline-IDENTITY (identity mapping). We additionally evaluate a \emph{batch/offline} strategy that memorizes token$\rightarrow$ciphertext lookups (“cache”), reflecting deployments that pre-secure a sensitive lexicon. Design choices and markers follow the main text. 

\paragraph{Findings.} (i) \emph{Primitive cost dominates.} Among reversible ciphers, AES (ECB) is fastest (0.0053 ms/token), followed by Blowfish (ECB) (0.0075) and AES (SIV) (0.0091). One-way hashes are substantially faster (MD5~$\approx$~0.0010, SHA-1~$\approx$~0.00105, SHA-256~$\approx$~0.00110 ms/token), while FPE (0.0088) and Soft Hash (0.0083) are slower than hashes but faster than AES (SIV). (ii) \emph{Caching usually helps across the board.} Offline memorization improves throughput by $\sim$3–10$\times$ depending on method (e.g., AES-ECB $\to$ 0.0012 ms/token; MD5 $\to$ 0.00030). (iii) \emph{Pipeline overhead is small and stable.} Preprocessing remains $\approx$0.02 ms/example for all secured methods (not including the time required by other steps in \texttt{HERALD}); plaintext pass-through is $\approx0.0002$ ms/example. Input/output MB/s remain flat across methods. Overall, securing methods adds acceptable latency relative to model inference, and batch/offline strategies provide predictable wins for high-reuse vocabularies.

\vspace{0.5em}
\noindent\textit{Notes.} Plaintext and Baseline-IDENTITY both denote identity mappings (minor numeric differences arise from harness effects). “Cache” denotes token$\rightarrow$ciphertext memoization simulating offline pre-encryption. 

\begin{table}[!htb]
\scriptsize
\centering
\renewcommand{\arraystretch}{1.2}
\setlength{\tabcolsep}{6pt}
\begin{adjustbox}{width=\linewidth}
\begin{tabular}{lcccc}
\toprule
\textbf{Method} & \textbf{ms/token} (no cache) & \textbf{MB/s} (no cache) & \textbf{ms/token} (cache) & \textbf{MB/s} (cache) \\
\midrule
Baseline-IDENTITY & 0.0001003 & 51.8320 & 0.0001457 & 35.6976 \\
Plaintext         & 0.0001067 & 48.7472 & 0.0001436 & 36.2072 \\
AES (SIV)         & 0.0091485 & 0.5680  & 0.0015246 & 3.4100  \\
AES (ECB)         & 0.0052602 & 0.9806  & 0.0012172 & 4.2715  \\
Blowfish (ECB)    & 0.0074852 & 0.6892  & 0.0014415 & 3.5776  \\
FPE               & 0.0088391 & 0.5871  & 0.0018553 & 2.8079  \\
Fuzzy Hash        & 0.0016508 & 3.1496  & 0.0004081 & 12.7420 \\
Soft Hash         & 0.0083077 & 0.6172  & 0.0016515 & 3.1980  \\
MD5               & 0.0010063 & 5.1669  & 0.0003023 & 17.2013 \\
SHA-1             & 0.0010496 & 4.9538  & 0.0003115 & 16.6934 \\
SHA-256           & 0.0011000 & 4.7268  & 0.0003337 & 15.5823 \\
\bottomrule
\end{tabular}
\end{adjustbox}
\caption{Per-method \textbf{conversion speed} on a 40k-token stream. Lower ms/token is better; higher MB/s is better. “Cache” denotes offline memoization.}
\label{tab:e1-conversion}
\end{table}

\begin{table}[!htb]
\scriptsize
\centering
\renewcommand{\arraystretch}{1.2}
\setlength{\tabcolsep}{6pt}
\begin{adjustbox}{width=\linewidth}
\begin{tabular}{lccccc}
\toprule
\textbf{Method} &  \textbf{Cache} & \textbf{ms/example} & \textbf{Input MB/s} & \textbf{Output MB/s} & \textbf{Examples} \\
\midrule
Plaintext & $\times$         & 0.0002619 & 379.7640 & 379.7640 & 200 \\
Plaintext & \checkmark       & 0.0002263 & 439.4755 & 439.4755 & 200 \\
AES (SIV) & $\times$         & 0.0196729 & 5.0549 & 5.1459 & 200 \\
AES (SIV) & \checkmark       & 0.0197841 & 5.0265 & 5.1170 & 200 \\
AES (ECB) & $\times$         & 0.0206570 & 4.8141 & 4.9008 & 200 \\
AES (ECB) & \checkmark       & 0.0200035 & 4.9714 & 5.0609 & 200 \\
Blowfish (ECB) & $\times$    & 0.0205607 & 4.8366 & 4.9237 & 200 \\
Blowfish (ECB) & \checkmark  & 0.0200097 & 4.9698 & 5.0593 & 200 \\
FPE & $\times$               & 0.0201783 & 4.9283 & 5.0170 & 200 \\
FPE & \checkmark             & 0.0201679 & 4.9309 & 5.0196 & 200 \\
Fuzzy Hash & $\times$        & 0.0200703 & 4.9548 & 5.0440 & 200 \\
Fuzzy Hash & \checkmark      & 0.0200034 & 4.9714 & 5.0609 & 200 \\
Soft Hash & $\times$         & 0.0205573 & 4.8374 & 4.9245 & 200 \\
Soft Hash & \checkmark       & 0.0204300 & 4.8676 & 4.9552 & 200 \\
MD5 & $\times$               & 0.0203614 & 4.8840 & 4.9719 & 200 \\
MD5 & \checkmark             & 0.0198677 & 5.0054 & 5.0954 & 200 \\
SHA-1 & $\times$             & 0.0223959 & 4.4403 & 4.5202 & 200 \\
SHA-1 & \checkmark           & 0.0202946 & 4.9001 & 4.9883 & 200 \\
SHA-256 & $\times$           & 0.0198673 & 5.0055 & 5.0956 & 200 \\
SHA-256 & \checkmark         & 0.0202859 & 4.9022 & 4.9904 & 200 \\
\bottomrule
\end{tabular}
\end{adjustbox}
\caption{\textbf{Preprocessing} for all secured methods. Lower ms/example is better; higher MB/s is better. $n{=}200$ examples. Results are shown with cache (\checkmark) and without cache ($\times$).}
\label{tab:e1-preproc}
\end{table}

\paragraph{Interpretation.} For deployments desiring reversible protection with maximum throughput, AES (ECB) (deterministic) offers the best speed; for irreversible obfuscation, MD5/SHA-1/SHA-256 provide the strongest throughput, with SHA-256 preferred when higher assurance is required. Offline pre-encryption (caching) yields consistent throughput gains and is recommended when sensitive vocabularies recur frequently.

\subsubsection{Sequence Length Inflation and Memory}
\label{app:seq-inflation}
We quantify how \texttt{HERALD}’s ciphertext and boundary markers ($\# \text{\textasciicircum}\, ... \,\text{\textasciicircum} \#$) affect tokenization and memory relative to plaintext. We evaluate seven controlled variants: plaintext; fully secured stem with options unsecured, without/with markers (\textsc{FS+OU} w/o, w/); \texttt{HERALD}-partially secured stem with options unsecured, without/with markers (\textsc{PS+OU} w/o, w/); and securing the options as well—\textsc{PS+OPS} and \textsc{FS+OFS} (both with markers). Tokens are counted post-template; batch size at fixed memory is estimated as the plaintext batch divided by the inflation ratio (Table~\ref{tab:seq-inflation-tokens}); we also measure peak CUDA memory on a single forward pass (Table~\ref{tab:seq-inflation-mem}). Delimiter design and the FS/PS and OU/OFS/OPS settings follow our framework description. 

\paragraph{Findings.} (i) \emph{Ciphertext alone} causes modest growth: \textsc{FS+OU} (w/o markers) and \textsc{PS+OU} (w/o markers) inflate tokens by {+22.6\%} and {+24.7\%}, reducing the feasible batch from 4$\to$3 at fixed memory. (ii) \emph{Markers dominate} inflation: adding markers on the stem lifts inflation to {+63.6\%} (\textsc{PS+OU} w/) and {+85.3\%} (\textsc{FS+OU} w/), cutting batch to 2. (iii) \emph{Securing options} multiplies encrypted spans: \textsc{PS+OPS} reaches {+143.2\%} and \textsc{FS+OFS} {+178.5\%}, forcing batch 1. (iv) Empirical GPU measurements usually track token growth: peak memory rises from 6.84 GB (plaintext) to 9.15–10.61 GB for marked stems and 10.81 GB for \textsc{FS+OFS} (+51\%). Overall, delimiter count—hence the number of secured spans—drives both sequence length and memory. In practice, \textsc{PS+OU} (w/ markers) balances privacy with tractable memory; omitting markers trims length but loses explicit boundaries that stabilize tokenization.

\begin{table}[!htb]
\scriptsize
\centering
\renewcommand{\arraystretch}{1.2}
\setlength{\tabcolsep}{6pt}
\begin{adjustbox}{width=\linewidth}
\begin{tabular}{lrrrr}
\toprule
\textbf{Variant} & \textbf{Mean} & \textbf{Std} & \textbf{Inflation} & \textbf{Batch @ same} \\
& \textbf{Tokens} & \textbf{Tokens} & \textbf{vs.\ Plain (\%)} & \textbf{memory (est.)} \\
\midrule
Plaintext & 60.30 & 26.65 & 0.00 & 4 \\
FS+OU (no markers) & 73.95 & 44.81 & 22.64 & 3 \\
FS+OU (with markers) & 111.76 & 91.34 & 85.33 & 2 \\
FS+OFS (with markers) & 167.94 & 105.65 & 178.50 & 1 \\
PS+OU (no markers) & 75.21 & 45.38 & 24.72 & 3 \\
PS+OU (with markers) & 98.63 & 78.12 & 63.57 & 2 \\
PS+OPS (with markers) & 146.66 & 88.64 & 143.22 & 1 \\

\bottomrule
\end{tabular}
\end{adjustbox}
\caption{\textbf{Token inflation and batch-size implications} at fixed memory budget. Inflation is relative to plaintext. OU: options unsecured; OFS/OPS: options fully/partially secured.}
\label{tab:seq-inflation-tokens}
\end{table}

\begin{table}[!htb]
\scriptsize
\centering
\renewcommand{\arraystretch}{1.2}
\setlength{\tabcolsep}{6pt}
\begin{adjustbox}{width=\linewidth}
\begin{tabular}{lrr}
\toprule
\textbf{Variant} & \textbf{Peak Bytes} & \textbf{Peak MiB} \\
\midrule
Plaintext & 7{,}174{,}982{,}144 & 6{,}842.60 \\
FS+OU (no markers) & 8{,}246{,}117{,}888 & 7{,}864.11 \\
FS+OU (with markers) & 9{,}595{,}937{,}280 & 9{,}151.40 \\
FS+OFS (with markers) & 10{,}815{,}572{,}480 & 10{,}314.53 \\
PS+OU (no markers) & 9{,}703{,}117{,}312 & 9{,}253.61 \\
PS+OU (with markers) & 10{,}606{,}887{,}424 & 10{,}115.52 \\
PS+OPS (with markers) & 10{,}408{,}585{,}728 & 9{,}926.40 \\
\bottomrule
\end{tabular}
\end{adjustbox}
\caption{Empirical \textbf{peak CUDA memory} per variant (single forward pass, same batch across rows).}
\label{tab:seq-inflation-mem}
\end{table}

\paragraph{Implications.} Because memory scales roughly with effective sequence length in decoder-only LMs, privacy settings should be chosen with hardware in mind: encrypting only the stem (\textsc{OU}) is notably cheaper than also securing options; reducing the number of secured spans (and thus markers) is the most direct lever for keeping batch sizes practical while retaining \texttt{HERALD}’s protection boundaries.

\section{Ablations and Privacy–Utility Trade-offs}
\label{app:ablations}
Here we isolate the contribution of each \texttt{HERALD} component and quantify operating points on the privacy–utility frontier.

\subsection{Privacy Levels in \texttt{HERALD}}
\label{app:privacy-levels}
We define Partial-A/B/C/D/E policy tiers. These policies are instantiated by POS-driven coverage rules.

\subsubsection{POS-Selective Encryption}
\label{app:pos-selective}
We evaluate the impact of part-of-speech (POS) selective encryption on task performance by progressively encrypting different (and more) grammatical categories. This analysis examines the trade-off between privacy protection and model utility when encrypting content words versus function words.

\paragraph{Setup.} We implement five encryption levels, each targeting specific POS categories:
\begin{itemize}
    \item \textbf{Level A}: NOUN only
    \item \textbf{Level B}: NOUN + PROPN (proper nouns)
    \item \textbf{Level C}: NOUN + PROPN + ADJ (adjectives)  
    \item \textbf{Level D}: NOUN + PROPN + ADJ + VERB
    \item \textbf{Level E}: NOUN + PROPN + ADJ + VERB + ADV (adverbs)
\end{itemize}

Tokens matching the target POS categories are encrypted using AES (SIV) and wrapped with markers, while function words and other categories remain in plaintext.

\paragraph{Findings.} Table~\ref{tab:pos-selective} presents the evaluation results across all encryption levels. Performance degrades monotonically as more POS categories are encrypted, with the steepest decline occurring when verbs are added (Level D). Encrypting only nouns (Level A) preserves 70.15\% accuracy, while comprehensive content word encryption (Level E) reduces performance to 61.28\%, representing a 8.87 percentage point decrease from the most conservative setting.

The results demonstrate that nouns and proper nouns carry substantial semantic information for medical question answering, as their encryption (Levels A-B) causes relatively modest performance degradation (0.40 percentage points). However, verbs appear particularly critical, with their inclusion (Level D) causing the largest single-step performance drop (3.45 percentage points). This suggests that action and process information encoded in verbs is essential for strong performance in \texttt{HERALD}.

\begin{table}[!htb]
\scriptsize
\centering
\renewcommand{\arraystretch}{1.2}
\setlength{\tabcolsep}{6pt}
\begin{adjustbox}{width=\linewidth}
\begin{tabular}{clcc}
\toprule
\textbf{Level} & \textbf{POS Categories} & \textbf{Accuracy} & \textbf{F\textsubscript{1}} \\
\midrule
A & NOUN & 0.7015 & 0.7014  \\
B & NOUN + PROPN & 0.6975 & 0.6972 \\
C & NOUN + PROPN + ADJ & 0.6812 & 0.6808 \\
D & NOUN + PROPN + ADJ + VERB & 0.6467 & 0.6463 \\
E & NOUN + PROPN + ADJ + VERB + ADV & 0.6128 & 0.6124 \\
\bottomrule
\end{tabular}
\end{adjustbox}
\caption{\textbf{Performance impact of POS-selective encryption.} Results show accuracy and F\textsubscript{1}-score across five encryption levels targeting different grammatical categories.}
\label{tab:pos-selective}
\end{table}

\paragraph{Implications.} The preservation of function words (articles, prepositions, conjunctions) maintains syntactic structure while protecting the most semantically dense tokens. This selective approach offers a practical middle ground between full encryption (maximum privacy, poor utility) and no encryption (maximum utility, no privacy).

Content words (e.g. nouns) typically carry domain-specific knowledge essential for specialized tasks like medical QA, while function words primarily provide grammatical scaffolding. Our results quantify this trade-off empirically, suggesting that noun and proper noun (PROPN) encryption (Level B) may offer an optimal privacy-utility balance for many applications.

\subsection{Marker Scheme Variants for Secured Tokens}
\label{app:marker-variants}

\paragraph{Setup.} We evaluate eight distinct marker schemes for delimiting encrypted tokens within the input text to understand their impact on model learnability and privacy preservation. The baseline approach embeds encrypted tokens directly without delimiters, while seven alternative schemes employ various sentinel token patterns to explicitly mark encrypted content boundaries.

\paragraph{Findings.} Table~\ref{tab:marker-variants} presents the performance comparison across all marker schemes. The scheme utilized in our main study (\texttt{$\# \text{\textasciicircum} ... \text{\textasciicircum} \#$}) achieves the highest accuracy (68.54\%), followed closely by mathematical notation delimiters (\texttt{\$\^{}...\^{}\$}) at 67.94\%. The no-marker baseline achieves 63.43\% accuracy, indicating that while models can learn to distinguish encrypted content without explicit delimiters, marker schemes provide meaningful performance improvements. The marker schemes exhibit a performance range of 5.11 percentage points (63.43\% to 68.54\%), indicating that delimiter choice has a measurable impact on model performance.

\paragraph{Implications.} The superior performance of the marker scheme utilized in our study (\texttt{$\# \text{\textasciicircum} ... \text{\textasciicircum} \#$}) validates our methodological choice for the main experiments. This scheme's effectiveness likely stems from its distinctiveness from natural language patterns and clear visual separation of encrypted content. Mathematical notation markers perform comparably well, suggesting that symbolic delimiters provide effective semantic boundaries for model learning. The moderate performance gap between marked and unmarked approaches (5.11 percentage points) demonstrates that while encrypted token distributions are learnable without delimiters, explicit markers provide substantial improvements in model comprehension.

\begin{table}[!htb]
\scriptsize
\centering
\renewcommand{\arraystretch}{1.2}
\setlength{\tabcolsep}{6pt}
\begin{adjustbox}{width=\linewidth}
\begin{tabular}{llcc}
\toprule
\textbf{Scheme} & \textbf{Marker Pattern} & \textbf{Acc.} & \textbf{F\textsubscript{1}} \\
\midrule
No markers & \textit{Only encrypted tokens} & 63.43 & 63.36 \\
Utilized in the study & \texttt{$\# \text{\textasciicircum} ... \text{\textasciicircum} \#$} & \textbf{68.54} & \textbf{68.50} \\
Medical XML & \texttt{<MED>...</MED>} & 66.24 & 66.14 \\
Security brackets & \texttt{[SEC]...[/SEC]} & 65.54 & 65.50 \\
Identity brackets & \texttt{[ID]...[/ID]} & 65.54 & 65.57 \\
Special tokens & \texttt{<|sec|>...<|/sec|>} & 66.25 & 66.20 \\
Unicode angles & \texttt{⟨...⟩} & 64.25 & 64.25 \\
Hash symbols & \texttt{\#...\#} & 66.04 & 66.95 \\
Math notation & \texttt{\$\^{}...\^{}\$} & 67.94 & 67.80 \\
\bottomrule
\end{tabular}
\end{adjustbox}
\caption{\textbf{Performance comparison of marker schemes} for encrypted token delimitation.}
\label{tab:marker-variants}
\end{table}

\subsection{Stopwords Impact}
\label{app:stopwords}

\paragraph{Setup.}
We investigated the impact of stopword removal on performance across different encryption coverage strategies. The experiments compared two preprocessing configurations: (1) retaining stopwords during encryption (\checkmark), and (2) removing English stopwords prior to sensitive token identification and encryption ($\times$). Each configuration was evaluated under both partial encryption (\texttt{HERALD}'s standard approach) and full encryption (encrypting all tokens) scenarios.

\paragraph{Findings.}
Stopword preprocessing demonstrated distinct effects across different encryption configurations and task types (Table~\ref{tab:stopword-marker-variants}). In \texttt{HERALD} configurations, stopword retention consistently outperformed removal across both tasks: for classification, retention achieved 56.7\% accuracy (56.2\% F\textsubscript{1}) compared to 52.8\% accuracy (52.5\% F\textsubscript{1}) with removal, representing a 3.9 percentage point improvement. Similarly, for MCQ tasks, retention yielded 56.6\% accuracy (55.4\% F\textsubscript{1}) versus 51.5\% accuracy (51.3\% F\textsubscript{1}) with removal, showing a 5.1 percentage point advantage. Whereas, full encryption configurations exhibited opposite trends regarding stopword preprocessing. For classification tasks, stopword removal achieved superior performance (46.7\% accuracy, 46.1\% F\textsubscript{1}) compared to retention (40.9\% accuracy, 40.4\% F\textsubscript{1}), suggesting that when all tokens are encrypted, removing non-content words may enhance focus on remaining semantic information. MCQ tasks under full encryption showed similar patterns, with removal yielding 37.8\% accuracy (29.9\% F\textsubscript{1}) versus 35.7\% accuracy (27.7\% F\textsubscript{1}) with retention.

\paragraph{Implications.}
The differential impact of stopword preprocessing across encryption configurations reveals important insights about the interaction between linguistic preprocessing and encryption strategies. In \texttt{HERALD}'s selective encryption approach, preserving stopwords maintains valuable contextual information that aids model comprehension, as common function words provide syntactic and semantic scaffolding around encrypted medical terms. The consistent 4-5 percentage point advantages of stopword retention in \texttt{HERALD} suggest that these linguistic markers enhance the model's ability to interpret encrypted content within preserved structural context. Conversely, under comprehensive encryption where all tokens are obfuscated, removing stopwords appears beneficial, likely because it reduces the overall encrypted token load and may help models focus on the remaining semantic content. This finding indicates that optimal preprocessing strategies are configuration-dependent: selective encryption benefits from linguistic context preservation, while comprehensive encryption may benefit from content filtering.

\begin{table}[!htb]
\scriptsize
\centering
\renewcommand{\arraystretch}{1.2}
\setlength{\tabcolsep}{6pt}
\begin{adjustbox}{width=\linewidth}
\begin{tabular}{llccc}
\toprule
\textbf{Task} & \textbf{Configuration} & \textbf{Stopwords} & \textbf{Acc.} & \textbf{F\textsubscript{1}} \\
\midrule
\multirow{4}{*}{Classification}
& Fully Secured & $\times$ & 0.467 & 0.461 \\
& Fully Secured & \checkmark & 0.409 & 0.404 \\
& \texttt{HERALD} & $\times$ & 0.528 & 0.525 \\
& \texttt{HERALD} & \checkmark & 0.567 & 0.562 \\
\midrule
\multirow{4}{*}{MCQ}
& Fully Secured & $\times$ & 0.378 & 0.299 \\
& Fully Secured & \checkmark & 0.357 & 0.277 \\
& \texttt{HERALD} & $\times$ & 0.515 & 0.513 \\
& \texttt{HERALD} & \checkmark & 0.566 & 0.554 \\
\bottomrule
\end{tabular}
\end{adjustbox}
\caption{\textbf{Performance comparison across stopword preprocessing and encryption coverage configurations.} $\times$ indicate stopword removal prior to encryption; \checkmark\, indicates stopword retention.}
\label{tab:stopword-marker-variants}
\end{table}

\subsection{Lemmatization Strategy}
\label{app:lemmatization}

\paragraph{Setup.} We evaluate two distinct lemmatization strategies for text preprocessing: (1) proposed \textit{targeted lemmatization}, which applies lemmatization selectively based on part-of-speech (POS) tagging and named entity recognition (NER), and (2) \textit{untargeted lemmatization}, which applies lemmatization indiscriminately to all non-stop words in the text.

The targeted approach implements a two-stage process: first, POS tagging identifies non-sensitive words that should remain unlemmatized to preserve contextual meaning; second, lemmatization is applied only to words not flagged as contextually critical. This preserves semantic relationships in domain-specific medical terminology while normalizing common linguistic variations. In contrast, the untargeted approach applies lemmatization uniformly across all content words, potentially introducing semantic drift in specialized terminology.

\paragraph{Findings.} Experimental results demonstrate that targeted lemmatization strategy as used in \texttt{HERALD} achieves substantial performance improvements across all evaluation metrics compared to untargeted lemmatization (Table~\ref{tab:lemmatization_comparison}). The targeted approach yields an accuracy of 69.55\% $\pm$ 0.18\%, representing a significant 3.40 percentage point improvement over the untargeted baseline (66.15\% $\pm$ 0.13\%). The targeted strategy shows even more pronounced improvements in F\textsubscript{1}-score (69.43\% $\pm$ 0.14\% vs. 65.79\% $\pm$ 0.16\%), demonstrating a 3.64 percentage point gain. These results indicate not only enhanced model performance but also improved consistency, as evidenced by the reported standard deviations.

\paragraph{Implications.} The substantial performance gains observed with targeted lemmatization strongly validate our hypothesis that selective morphological normalization preserves critical semantic information while reducing lexical diversity. This is particularly crucial in medical question-answering tasks where precise terminology directly impacts model comprehension and clinical accuracy.

\begin{table}[!htb]
\scriptsize
\centering
\renewcommand{\arraystretch}{1.2}
\setlength{\tabcolsep}{6pt}
\begin{adjustbox}{width=\linewidth}
\begin{tabular}{lcc}
\toprule
\textbf{Metric} & \textbf{Untargeted Lemmatization} & \textbf{Targeted Lemmatization} \\
\midrule
Accuracy (\%) & 66.15 $\pm$ 0.13 & 69.55 $\pm$ 0.18 \\
F\textsubscript{1}-Score (\%) & 65.79 $\pm$ 0.16 & 69.43 $\pm$ 0.14 \\
\midrule
\textbf{Improvement} & \textbf{Baseline} & \textbf{+3.40\% / +3.64\%} \\
\bottomrule
\end{tabular}
\end{adjustbox}
\caption{\textbf{Performance comparison between targeted and untargeted lemmatization} strategies on medical question answering task. Results show consistent improvements with \texttt{HERALD}'s targeted approach across all evaluation metrics.}
\label{tab:lemmatization_comparison}
\end{table}

\subsection{Privacy–Utility Curves}
\label{app:privacy-utility}

\textbf{Setup.}
We perform a parameter sweep over the encryption rate \(f\), defined as the fraction of prompt tokens that are encrypted, with \(f \in \{0,\,0.10,\,0.25,\,0.50,\,0.75,\,1.00\}\). The selection procedure is deterministic: we first encrypt sensitive spans and, as $f$ increases, progressively include non-sensitive spans until the entire prompt is transformed at $f=1$. Encryption is performed with the \texttt{HERALD} transform (AES-SIV). For each $f$, we fine-tune and evaluate the same configuration, reporting accuracy on the held-out test split. This realizes the \texttt{HERALD}’s “dial-a-privacy’’ knob while preserving the training protocol.

\begin{figure}[!htb]
  \centering
  \includegraphics[width=\linewidth]{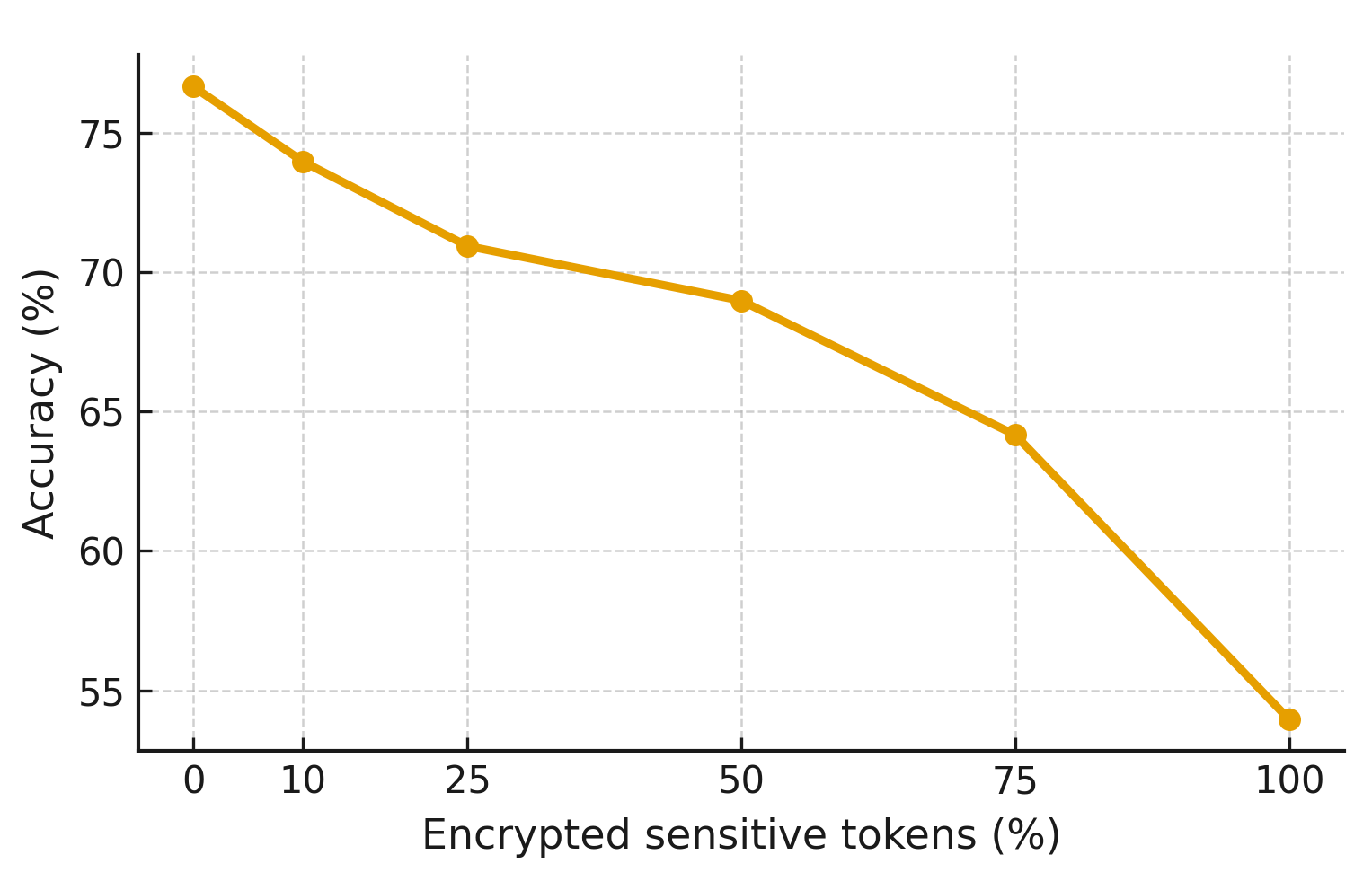}
  \caption{\textbf{Privacy–utility curve (Accuracy).} Test Accuracy (\%) as a function of the fraction $f$ of tokens encrypted. Accuracy decreases smoothly with $f$.}
  \label{fig:privacy-utility-acc}
\end{figure}

\textbf{Findings.}
Figure~\ref{fig:privacy-utility-acc} shows a smooth, monotonic trade-off: utility degrades approximately linearly up to $f{\approx}0.5$ and more sharply thereafter. Relative to plaintext ($f{=}0$), fully encrypting all tokens ($f{=}1$) reduces Accuracy from $76.66\%$ to $53.95\%$. The \emph{sweet spot} occurs at $f{=}0.10$, retaining $\sim\!97\%$ of baseline accuracy with only a $\sim\!2.7$-point Accuracy drop (76.66$\rightarrow$73.97). Beyond $f{\geq}0.25$, utility declines more noticeably, especially past $f{\geq}0.75$.

\section{Security Evaluation and Attack Surface}
\label{app:security}
We map threats to experiments and clarify which protections are offered by \texttt{HERALD} vs.\ what remains exposed through context.

\subsection{Training Data Extraction}
\label{app:extraction}

\paragraph{Setup.}
We probe fine-tuned models for regurgitation using four prompts: (i) \texttt{verbatim-global-plain} (``repeat a training question verbatim''), (ii) \texttt{extraction-list-plain} (``list three training questions verbatim''), (iii) \texttt{prefix-continue-plain} (continue a gold question given a long plaintext prefix), and (iv) \texttt{prefix-continue-cipher} (\texttt{HERALD}-encoded prefix). We report exact-match rate (EM), mean character-level similarity (MS; 0–1), and the rate at which ciphertext markers ($\# \text{\textasciicircum}\, ... \,\text{\textasciicircum} \#$) appear in model outputs (CMR). The results are summarized in Table~\ref{tab:extraction}.

\paragraph{Findings.}
(1) Free-form recall fails to extract verbatim text in all cases (EM=0 for both \texttt{verbatim-global-plain} and \texttt{extraction-list-plain}), but \texttt{HERALD}’s outputs \emph{always} contain ciphertext markers (CMR=1.0), ensuring any recall manifests as ciphertext rather than plaintext.  
(2) With a long plaintext prefix, both models frequently autocomplete the remainder (EM $\approx$0.10–0.65; MS $\approx$0.89–0.92) for \emph{both} TRAIN and TEST, indicating anchored next-token completion rather than preferential recall of train items; when continuations touch sensitive content, \texttt{HERALD} flips into ciphertext (CMR $\approx$0.43–0.59). This behavior matches \texttt{HERALD}’s design: any memorized sensitive spans, if elicited, appear encrypted.
(3) Given a \texttt{HERALD}-encoded prefix, the model never reveals plaintext (EM=0, CMR=1.0) and continues in ciphertext (moderate MS from prompt echo), consistent with the framework’s confidentiality goal.

\begin{table}[!tbh]
\scriptsize
\centering
\renewcommand{\arraystretch}{1.2}
\setlength{\tabcolsep}{6pt}
\begin{adjustbox}{width=\linewidth}
\begin{tabular}{lllcccc}
\toprule
Variant & Split & Attack & $n$ & EM (\%) & MS & CMR (\%) \\
\midrule
\texttt{HERALD}    & TEST  & \texttt{extraction-list-plain}    & 48 & 0.0  & 0.342 & 100.0 \\
\texttt{HERALD}    & TEST  & \texttt{prefix-continue-cipher}   & 48 & 0.0  & 0.513 & 100.0 \\
\texttt{HERALD}    & TEST  & \texttt{prefix-continue-plain}    & 48 & 12.4 & 0.704 & 42.5  \\
\texttt{HERALD}    & TEST  & \texttt{verbatim-global-plain}    & 48 & 0.0  & 0.332 & 100.0 \\
\texttt{HERALD}    & TRAIN & \texttt{extraction-list-plain}    & 48 & 0.0  & 0.326 & 100.0 \\
\texttt{HERALD}    & TRAIN & \texttt{prefix-continue-cipher}   & 48 & 0.0  & 0.539 & 100.0 \\
\texttt{HERALD}    & TRAIN & \texttt{prefix-continue-plain}    & 48 & 22.5 & 0.790 & 58.8  \\
\texttt{HERALD}    & TRAIN & \texttt{verbatim-global-plain}    & 48 & 0.0  & 0.320 & 100.0 \\
Plaintext & TEST  & \texttt{extraction-list-plain}    & 48 & 0.0  & 0.411 & 0.0   \\
Plaintext & TEST  & \texttt{prefix-continue-plain}    & 48 & 40.2 & 0.916 & 0.0   \\
Plaintext & TEST  & \texttt{verbatim-global-plain}    & 48 & 0.0  & 0.421 & 0.0   \\
Plaintext & TRAIN & \texttt{extraction-list-plain}    & 48 & 0.0  & 0.412 & 0.0   \\
Plaintext & TRAIN & \texttt{prefix-continue-plain}    & 48 & 64.6 & 0.911 & 0.0   \\
Plaintext & TRAIN & \texttt{verbatim-global-plain}    & 48 & 0.0  & 0.422 & 0.0   \\
\bottomrule
\end{tabular}
\end{adjustbox}
\caption{\textbf{Regurgitation probes.} EM: exact match; MS: mean similarity (0–1); CMR: ciphertext-marker rate. Free-form recall fails (EM=0), and \texttt{HERALD} always emits ciphertext in such cases (CMR=1.0). Long-prefix continuation yields high EM for both models, including TEST, indicating anchored completion rather than train-specific memorization; \texttt{HERALD} transitions into ciphertext when continuations touch sensitive spans. Values are averaged over multiple runs.}
\label{tab:extraction}
\end{table}

\paragraph{Implication.} Under recall-style prompts, no verbatim training items are recovered, while \texttt{HERALD} routes any residual recall into ciphertext; with strong anchors, both models autocomplete but \texttt{HERALD} preserves confidentiality at sensitive spans.

\subsection{Jailbreaking and Prompt Injection}
\label{app:jailbreak}

\paragraph{Setup.}
We evaluate whether \texttt{HERALD}’s token-level cipher spans resist prompt-injection and jailbreak attempts that explicitly solicit decryption or implicitly induce “best-guess” substitutions. The study follows the security evaluation plan with markers introduced in the core workflow to bound cipher spans and stabilize tokenization.

\paragraph{Attack suite.}
We probe a held-out MCQ subset formatted per the main prompting recipe (question + options + “answer with the letter”). For each secured prompt, we instantiate six attacks:
\textsc{Direct-Decrypt-Request}, \textsc{Explain-Then-Decrypt}, \textsc{Pretend-Key-Available}, \textsc{Marker-Strip-Rewrite}, \textsc{Summarize-Reveal-Secrets}, and a marker-ablation control \textsc{NoMarkers-Guess-Replacements}. The first five operate on the original \texttt{HERALD}-protected text; the control removes $\# \text{\textasciicircum}\,$ / $\,\text{\textasciicircum} \#$ before issuing a rewrite/guess instruction.

\paragraph{Protocol and metric.}
We evaluate $N{=}512$ secured prompts (deterministic decoding). A response is flagged \emph{explicit-success} if it unambiguously claims decryption or plaintext revelation (e.g., “decrypted: …”), and \emph{alpha-guess} if it appears to replace a referenced ciphertext span with nearby alphabetic content (heuristic windowing). We report per-attack rates and \emph{any-success} $=$ \emph{explicit} $\lor$ \emph{alpha-guess}. Table~\ref{tab:jailbreak} summarizes the results.

\begin{table}[!htb]
\scriptsize
\centering
\renewcommand{\arraystretch}{1.2}
\setlength{\tabcolsep}{6pt}
\begin{adjustbox}{width=\linewidth}
\begin{tabular}{lccc}
\toprule
\textbf{Attack} & \textbf{Explicit} $\downarrow$ & \textbf{Alpha-guess} $\downarrow$ & \textbf{Any} $\downarrow$ \\
\midrule
Direct-Decrypt-Request            & 0.000 & 0.000000 & 0.000000 \\
Explain-Then-Decrypt              & 0.000 & 0.000000 & 0.000000 \\
Pretend-Key-Available             & 0.000 & 0.000000 & 0.000000 \\
Summarize-Reveal-Secrets          & 0.000 & 0.015625 & 0.015625 \\
Marker-Strip-Rewrite              & 0.000 & 0.046875 & 0.046875 \\
NoMarkers-Guess-Replacements      & 0.000 & 0.000000 & 0.000000 \\
\bottomrule
\end{tabular}
\end{adjustbox}
\caption{\textbf{Heuristic jailbreak/prompt-injection success on \texttt{HERALD}-protected prompts} ($N{=}512$). “Explicit” denotes overt claims of decryption; “alpha-guess” flags local plaintext-like substitutions near cited ciphertext. All rates are fractions in $[0,1]$.}
\label{tab:jailbreak}
\end{table}

\paragraph{Findings.}
(1) No attack achieved \emph{explicit} decryption disclosure—consistent with the design goal that ciphertext spans are unrecoverable without the secret key. (2) Low but non-zero \emph{alpha-guess} appears when attacks instruct the model to rewrite while removing markers (\textsc{Marker-Strip-Rewrite}, $4.69\%$) or to summarize then list “secrets” ($1.56\%$), yet absolute rates remain small. (3) The marker-ablation control shows $0\%$ under our heuristic; because detection anchors on nearby ciphertext substrings, this conservative metric likely \emph{under}counts guesses when markers (and thus the cipher anchor) are absent\footnote{Heuristic limitations: if an attack elicits fluent replacements without echoing the original ciphertext substring, \emph{alpha-guess} may miss it. We therefore interpret the table as a \emph{lower bound} on guess-like leakage.}.

\paragraph{Effect of markers.}
\texttt{HERALD}’s explicit span delimiters serve two purposes: they prevent tokenizer fragmentation of non-linguistic ciphertext substrings and teach the model a stable “foreign alphabet” for secured spans during fine-tuning, reducing the odds that rewriting instructions are interpreted as a request to infer plaintext. Empirically, attacks that operate \emph{without} respecting markers do not yield overt disclosures, and the observed guess-like behavior stays rare. This aligns with the framework rationale that markers bound the secured region and preserve learnability without exposing content.

\subsection{Semantic Similarity Leakage}
\label{app:sem-sim}

We assess whether \texttt{HERALD}’s ciphertext tokens carry residual semantics in off-the-shelf embedding space. For each sensitive plaintext token $w$ identified by the \texttt{HERALD} pipeline, we form a paired ciphertext token $z$ using one of nine methods (AES–SIV/ECB, Blowfish–ECB, FPE, Fuzzy Hash, Soft Hash, MD5, SHA-1, SHA-256). We embed ${w}$ and ${z}$ with a sentence encoder and compute cosine similarities for aligned (``true'') pairs and for a null baseline obtained by shuffling ciphertexts. We report distributional separation (AUROC/AUPRC), two-sample tests (KS, Mann–Whitney $U$, Welch $t$; Benjamini–Hochberg adjusted $q$), and a histogram-overlap coefficient.

\paragraph{Findings.}
Across all 18 runs (9 methods $\times$ {with, without} markers), the aligned and shuffled cosine distributions are statistically indistinguishable: AUROC $\approx$ 0.49–0.51; KS/MW/$t$ $p$-values remain high (BH-$q \approx 0.96$) for every method/setting; histogram overlaps are large ($\sim$0.78–0.84). The largest mean shift observed is $\Delta\mu{=}0.003$ for FPE (no markers), still practically negligible. Consequently, we find no evidence of embedding-space semantic leakage for any method under this test. Method choice may therefore be guided by reversibility and token-length inflation rather than embedding leakage (Table~\ref{tab:sem-leakage}).

\begin{table}[!tbh]
\scriptsize
\centering
\renewcommand{\arraystretch}{1.2}
\setlength{\tabcolsep}{6pt}
\begin{adjustbox}{width=\linewidth}
\begin{tabular}{l l cc cc cc cc}
\toprule
\multirow{2}{*}{Method} & \multirow{2}{*}{Family} &
\multicolumn{2}{c}{$\mathrm{AUROC} \downarrow$} &
\multicolumn{2}{c}{$|\Delta\mu|\!\times\!10^3 \downarrow$} &
\multicolumn{2}{c}{Overlap $\uparrow$} &
\multicolumn{2}{c}{Length ratio $\downarrow$} \\
&  & \checkmark & $\times$ & \checkmark & $\times$ & \checkmark & $\times$ & \checkmark & $\times$ \\
\midrule
AES (SIV) & symmetric enc.   & 0.504 & 0.497 & {+}2.031 & {+}1.293 & 0.815 & 0.782 & 4.731 & 4.199 \\
AES (ECB) & symmetric enc.   & 0.505 & 0.499 & {+}1.802 & {+}1.426 & 0.791 & 0.800 & 3.519 & 2.987 \\
Blowfish (ECB) & symmetric enc.& 0.498 & 0.501 & $-$0.718 & {+}0.689 & 0.807 & 0.824 & 2.702 & 2.170 \\
FPE & format-preserving   & 0.494 & 0.507 & $-$1.070 & {+}3.017 & 0.816 & 0.791 & 1.518 & 0.986 \\
Fuzzy Hash & similarity hash & 0.501 & 0.502 & $-$0.778 & $-$0.966 & 0.834 & 0.827 & 1.601 & 1.068 \\
Soft Hash & similarity hash  & 0.505 & 0.506 & {+}1.558 & {+}1.575 & 0.814 & 0.810 & 2.662 & 2.130 \\
MD5 & crypto hash         & 0.498 & 0.498 & $-$0.613 & $-$0.573 & 0.808 & 0.802 & 4.792 & 4.259 \\
SHA-1 & crypto hash        & 0.503 & 0.501 & {+}0.961 & {+}1.380 & 0.799 & 0.795 & 5.857 & 5.324 \\
SHA-256 & crypto hash      & 0.493 & 0.493 & $-$0.823 & $-$0.533 & 0.788 & 0.783 & 9.051 & 8.519 \\
\bottomrule
\end{tabular}
\end{adjustbox}
\caption{\textbf{Embedding-space leakage check.} For each method we report AUROC (higher worse for leakage), mean-shift $\Delta\mu=\mathrm{mean}*{\text{true}}-\mathrm{mean}*{\text{null}}$ (scaled by $10^3$), histogram overlap (higher implies more similar distributions), and length ratio (cipher/plain). Results are shown with (\checkmark) and without ($\times$) markers. All two-sample tests are non-significant after BH correction (BH-$q{\approx}0.96$ across rows).}
\label{tab:sem-leakage}
\end{table}

\paragraph{Notes on cost.}
Length inflation varies by method: similarity hashes and FPE are compact (median length ratios $\approx$1.0–2.7), symmetric ciphers moderate (2.2–4.7), and cryptographic hashes highest (4.3–9.1). These costs affect context budget but not leakage in our analysis.

\paragraph{Limitations.}
This check targets type-level similarity of isolated token forms. It does not quantify context-driven inference (where surrounding plaintext could reveal meaning).

\paragraph{Interpretation.}
Values asymptoting to AUROC ${\approx}0.50$, tiny $\Delta\mu$ (all $|\Delta\mu|\le 3{\times}10^{-3}$), and large overlaps collectively indicate that embeddings do not align ciphertexts with their plaintext analogs any better than chance. Within this test’s scope, all methods “pass” the semantic-similarity leakage criterion; practitioners can therefore prioritize reversibility (e.g., AES, FPE), digest compactness (e.g., FPE), or strong one-wayness (e.g., SHA-256) without sacrificing this privacy property.

\subsection{Embedding Space Probing}
\label{app:embeddings}

\begin{table*}[!tbh]
\scriptsize
\centering
\renewcommand{\arraystretch}{1.2}
\setlength{\tabcolsep}{6pt}
\begin{adjustbox}{width=\linewidth}
\begin{tabular}{lcccccccc}
\toprule
Method & $n$ & kNN@5 & $\rho$ (cos) & UMAP kNN@5 & t-SNE kNN@5 & Trustw. (U) & Trustw. (T) & $p$ (U/T) \\
\midrule
AES (SIV) & 100 & 0.019 & $-0.012$ & 0.039 & 0.033 & 0.860 & 0.889 & 0.088 / 0.253 \\
Blowfish (ECB) & 100 & 0.035 & $-0.153$ & 0.033 & 0.036 & 0.874 & 0.888 & 0.265 / 0.170 \\
FPE & 100 & 0.045 & $+0.023$ & 0.022 & 0.031 & 0.885 & 0.896 & 0.798 / 0.379 \\
\bottomrule
\end{tabular}
\end{adjustbox}
\caption{\textbf{Embedding-space probing of cipher tokens.} \emph{k}NN@5: plaintext–cipher neighbor-set overlap (higher $\uparrow$ better); $\rho$: Spearman correlation of pairwise cosine similarities between plaintext and cipher spaces (higher $\uparrow$ better). UMAP/t-SNE columns report 2-D overlaps; “Trustw.” is trustworthiness (higher $\uparrow$ better). $p$-values compare observed 2-D overlaps to a permutation baseline; none are significant at $\alpha{=}0.05$.}
\label{tab:probe_all}
\end{table*}

\paragraph{Goal.} We test whether secured (cipher) tokens inherit distributional structure from their plaintext counterparts in a standard embedding space. If alignment exists, cipher tokens should (i) retrieve the same neighbors as their plaintext tokens and (ii) form clusters consistent with plaintext semantics.

\paragraph{Setup.} From frequent sensitive tokens, we built a 100-token probe set per method and produced cipher tokens using the project’s implementations (AES-SIV, Blowfish-ECB, and FPE). We embedded short, neutral contexts containing either the plaintext or the delimiter-wrapped cipher token ($\# \text{\textasciicircum}\, ... \,\text{\textasciicircum} \#$). Metrics: (1) \emph{k}-NN set overlap at \emph{k}=5 between plaintext and cipher neighborhoods (cosine); (2) Spearman correlation $\rho$ between pairwise plaintext and cipher cosine similarities; (3) $k$-means (\emph{k}=10) qualitative inspection; (4) 2-D projections (UMAP, t-SNE) with trustworthiness and a permutation baseline for the \emph{k}-NN-overlap statistic.

\paragraph{Findings.} Across methods, neighborhood overlap is small and near permutation baselines, and $\rho$ is near-zero or negative (Table~\ref{tab:probe_all}). AES (SIV) shows a weak trend in UMAP space (overlap 0.039, $p\approx0.088$) but remains non-significant. Blowfish (ECB) exhibits the strongest anti-correlation ($\rho=-0.153$), while FPE is closest to chance ($\rho=0.023$; mixed 2-D overlaps and the largest $p$-values). $k$-means partitions of cipher embeddings were heterogeneous and did not recover plaintext semantic groupings (e.g., clusters mixing \emph{patient}, \emph{blood}, \emph{fever}, \emph{management}). Overall, we find no statistically reliable evidence that cipher tokens align to plaintext semantics under these securing schemes.

\subsection{Recoverability of Tokens}
\label{app:recover-tokens}

\paragraph{Setup.} We evaluate whether a plaintext token can be “recovered’’ from the embedding space by nearest-neighbor (NN) search against its own transformed counterpart. Concretely, for a vocabulary subset we compute input-embedding centroids of (i) the plaintext token and (ii) the corresponding secured string \texttt{$E_k(\cdot)$}. We measure pairwise Euclidean distances (primary metric) from plaintext to all secured embeddings and report the exact-match rate that the gold secured token is the top neighbor (\textbf{NN@1}) or among the top five (\textbf{NN@5}). To probe the role of markers, we consider two embedding variants: (a) excluding marker subtokens when forming centroids; (b) including markers, reflecting what the model actually “sees.” This aligns with \texttt{HERALD}’s deterministic token-level transform and marker design.

\paragraph{Results.} Table~\ref{tab:recover-topk} reports Euclidean Top-$k$ recoverability (3000 tokens). Table~\ref{tab:recover-metrics} expands to multiple distances under both marker-handling regimes (5000 tokens). Across seven transforms (AES–SIV, AES–ECB, Blowfish–ECB, FPE, Fuzzy Hash, MD5, SHA-1), NN@1 and NN@5 are essentially at chance: all accuracies are $\leq\!0.17\%$, indicating negligible recoverability via nearest neighbors in the model’s input-embedding space. Including markers generally further suppresses NN@1. Alternative distances (cosine, dot-product, Mahalanobis/whitened, CSLS) yield similarly near-zero NN@1, with minor fluctuations that do not change the conclusion. These observations are consistent with \texttt{HERALD}’s goal: deterministic, delimiter-bounded ciphertext token embeddings behave as distinct token types whose vectors do not systematically align with their plaintext counterparts in embedding space (i.e., they do not geometrically ‘point back’ to plaintext).

\begin{table}[!tbh]
\tiny
\centering
\renewcommand{\arraystretch}{1.2}
\setlength{\tabcolsep}{12pt}
\begin{tabular}{lcc}
\toprule
\textbf{Transform} & \textbf{NN@1} & \textbf{NN@5} \\
\midrule
AES (SIV)         & 0.0005 & 0.0015 \\
AES (ECB)         & 0.0000 & 0.0035 \\
Blowfish (ECB)    & 0.0005 & 0.0020 \\
FPE               & 0.0005 & 0.0020 \\
Fuzzy Hash        & 0.0005 & 0.0025 \\
MD5               & 0.0005 & 0.0025 \\
SHA-1             & 0.0005 & 0.0025 \\
\bottomrule
\end{tabular}
\caption{\textbf{Recoverability via Euclidean distance:} Top-$k$ exact-match accuracy. $N\!=\!3000$ plaintext tokens. Values are averaged over multiple runs.}
\label{tab:recover-topk}
\end{table}

\FloatBarrier
\begin{table}[!htbp]
\scriptsize
\centering
\renewcommand{\arraystretch}{1}
\setlength{\tabcolsep}{8pt}
\begin{adjustbox}{width=\linewidth}
\begin{tabular}{lccccc|ccccc}
\toprule
\multirow{2}{*}{\textbf{Transform}} & \multicolumn{5}{c}{\textbf{Exclude markers}} & \multicolumn{5}{c}{\textbf{Include markers}} \\
\cmidrule{2-6} \cmidrule{7-11}
 & Euclid. & Cosine & Dot & Mahalanobis & CSLS & Euclid. & Cosine & Dot & Mahalanobis & CSLS \\
\midrule
AES (SIV)      & 0.0003 & 0.0000 & 0.0003 & 0.0003 & 0.0000 & 0.0000 & 0.0000 & 0.0000 & 0.0003 & 0.0000 \\
AES (ECB)      & 0.0007 & 0.0010 & 0.0003 & 0.0000 & 0.0000 & 0.0003 & 0.0017 & 0.0000 & 0.0000 & 0.0000 \\
Blowfish (ECB) & 0.0003 & 0.0007 & 0.0003 & 0.0003 & 0.0003 & 0.0003 & 0.0003 & 0.0007 & 0.0000 & 0.0003 \\
FPE            & 0.0000 & 0.0010 & 0.0000 & 0.0000 & 0.0000 & 0.0003 & 0.0000 & 0.0003 & 0.0000 & 0.0000 \\
Fuzzy Hash     & 0.0003 & 0.0000 & 0.0000 & 0.0003 & 0.0007 & 0.0000 & 0.0010 & 0.0007 & 0.0003 & 0.0007 \\
MD5            & 0.0003 & 0.0007 & 0.0000 & 0.0003 & 0.0003 & 0.0007 & 0.0000 & 0.0000 & 0.0007 & 0.0000 \\
SHA-1          & 0.0007 & 0.0007 & 0.0007 & 0.0003 & 0.0010 & 0.0007 & 0.0007 & 0.0007 & 0.0003 & 0.0013 \\
\bottomrule
\end{tabular}
\end{adjustbox}
\caption{\textbf{Nearest-neighbor accuracy ($\mathrm{\mathbf{NN@1}}$) by distance and marker handling.} Left block excludes marker subtokens when embedding secured strings; right block includes them. $N\!=\!5000$. Values are averaged over multiple runs.}
\label{tab:recover-metrics}
\end{table}

\paragraph{Interpretation.} All methods produce near-chance NN recoverability under Euclidean distance, and alternative distances do not materially increase it; if anything, markers reduce residual matches further by injecting consistent boundary cues around ciphertext spans. Together with the deterministic token-level design, this suggests that \texttt{HERALD} secures sensitive tokens without leaving an “embedding fingerprint’’ that trivially links ciphertext to plaintext via nearest-neighbor geometry.

\end{document}

%% file: preamble.tex
\usepackage{placeins}
\usepackage{times}
\usepackage{float}
\usepackage{enumitem}
\usepackage{tikz}
\usetikzlibrary{shapes, arrows.meta, positioning}
\usepackage{booktabs}
\usepackage{multirow}
\usepackage{latexsym}
\usepackage{amsmath}  
\usepackage{amssymb}   
\usepackage{tikz} 
\usepackage{xcolor}
\usepackage[T1]{fontenc}
\usepackage[utf8]{inputenc}
\usepackage{microtype}
\usepackage{amsmath,amssymb}                  
\usepackage{algorithmicx}          
\usepackage{algpseudocode}         
\usepackage{inconsolata}
\usepackage{graphicx}
\usepackage{subcaption, algorithm2e, url, float, etoolbox, pgf,booktabs,verbatim}
\usepackage{xcolor,graphicx,colortbl}

\usepackage{enumitem}
\setlist[enumerate]{label*=\arabic*.}

\usepackage[labelsep=period]{caption}

\usepackage{tabularx}
\usepackage{multirow}
\usepackage{colortbl}
\usepackage{comment}
\usepackage{textcomp}